\documentclass{article} 
\usepackage{iclr2027_conference,times}

\newif\ifISSbody     \ISSbodytrue
\newif\ifISSappendix \ISSappendixtrue
\ifx\ISSskipbody\undefined\else\ISSbodyfalse\fi
\ifx\ISSskipappendix\undefined\else\ISSappendixfalse\fi

\ifdefined\XeTeXversion\RequirePackage[OT1]{fontenc}\fi

\usepackage{amsmath,amsfonts,bm}

\def\eqref#1{equation~\ref{#1}}

\def\1{\bm{1}}

\DeclareMathAlphabet{\mathsfit}{\encodingdefault}{\sfdefault}{m}{sl}
\SetMathAlphabet{\mathsfit}{bold}{\encodingdefault}{\sfdefault}{bx}{n}

\newcommand{\mfOverIss}{0.75}
\newcommand{\mfVoid}{25\%}

\newcommand{\mfLam}{0.25}
\newcommand{\mfP}{0.10}
\newcommand{\mfRep}{3}
\newcommand{\mfErr}{2.8\%}
\newcommand{\mfDiscount}{0.75}

\newcommand{\mfLamA}{0.10}

\newcommand{\mfRatioA}{0.31}
\newcommand{\mfVoidA}{20\%}
\newcommand{\mfLowfidA}{9.1}

\newcommand{\mfLamB}{0.50}

\newcommand{\mfRatioB}{0.99}
\newcommand{\mfVoidB}{0\%}
\newcommand{\mfLowfidB}{0.0}

\newcommand{\mfLamC}{1.00}

\newcommand{\mfRatioC}{0.99}
\newcommand{\mfVoidC}{0\%}
\newcommand{\mfLowfidC}{0.0}

\newcommand{\absPROBEAcc}{100\%}

\newcommand{\absNAIVEAcc}{99\%}

\newcommand{\dasNets}{8}

\IfFileExists{macros/macros_missing.tex}{\input{macros/macros_missing.tex}}{}
\newcommand{\mnBypassAcc}{100.0\%}

\newcommand{\mnBypassNets}{10}

\newcommand{\mnBypassPointerShape}{99.0\%}
\newcommand{\mnBypassReads}{color}

\newcommand{\mnBypassVisIntendedTwo}{0.52}

\newcommand{\mnBypassVisIssCorrectTwo}{100\%}

\newcommand{\mnBypassVisIssCostTwo}{9.0}

\newcommand{\mnBypassVisOptTwo}{7.0}
\newcommand{\mnBypassVisReal}{100\%}

\newcommand{\mnBypassVisRpCostTwo}{26.4}

\newcommand{\mnBypassWbIntendedTwo}{0.36}

\newcommand{\mnBypassWbIssCorrectTwo}{0\%}

\newcommand{\mnBypassWbIssCostTwo}{9.0}

\newcommand{\mnBypassWbIssValRefTwo}{93\%}

\newcommand{\mnBypassWbReal}{0\%}

\newcommand{\mnBypassWbRpCostTwo}{19.1}

\newcommand{\mnBypassWbRpRefutedTwo}{97\%}

\newcommand{\mnCnnASixtyThouShapePct}{50\%}

\newcommand{\mnCnnCSixtyThouShapePct}{3\%}

\newcommand{\mnColorAcc}{100.0\%}

\newcommand{\mnColorNets}{5}

\newcommand{\mnColorReads}{color}

\newcommand{\mnColorVisIntendedPctTwo}{52\%}

\newcommand{\mnColorVisIntendedTwo}{0.52}

\newcommand{\mnColorVisIssCorrectTwo}{100\%}

\newcommand{\mnColorVisIssCostTwo}{9.2}

\newcommand{\mnColorVisOptTwo}{7.0}
\newcommand{\mnColorVisReal}{100\%}

\newcommand{\mnColorVisRpCostTwo}{26.2}

\newcommand{\mnInclassAccMin}{97.6\%}
\newcommand{\mnInclassMis}{0.56\%}

\newcommand{\mnInclassSkipped}{1.0\%}

\newcommand{\mnInclassVisCoverCostTwo}{15.1}

\newcommand{\mnInclassVisFloorPctTwo}{36\%}

\newcommand{\mnInclassVisIssCostTwo}{8.6}

\newcommand{\mnInclassVisIssNMaxTwo}{6}

\newcommand{\mnInclassVisOptTwo}{7.0}

\newcommand{\mnInclassVisRpCostTwo}{25.9}

\newcommand{\mnKtwoNets}{20}

\newcommand{\mnKtwoReal}{100\%}
\newcommand{\mnKtwoRealruns}{12{,}000}

\newcommand{\mnMixedAcc}{99.7\%}

\newcommand{\mnMixedNets}{5}

\newcommand{\mnMixedReads}{shape (84\%)}

\newcommand{\mnMixedVisIntendedTwo}{0.06}

\newcommand{\mnMixedVisIssCorrectTwo}{99\%}

\newcommand{\mnMixedVisIssCostTwo}{9.0}

\newcommand{\mnMixedVisIssValRefTwo}{100\%}

\newcommand{\mnMixedVisIssVoidTwo}{1\%}

\newcommand{\mnMixedVisNotReal}{8\%}

\newcommand{\mnMixedVisOptTwo}{7.1}
\newcommand{\mnMixedVisReal}{92\%}

\newcommand{\mnMixedVisRpCostTwo}{26.5}

\newcommand{\mnPlantedAcc}{99.0\%}

\newcommand{\mnPlantedMatch}{98\%}
\newcommand{\mnPlantedMatchSlots}{99.5\%}

\newcommand{\mnPlantedNets}{20}

\newcommand{\mnPlantedReads}{as planted (98\%)}

\newcommand{\mnPlantedVisIntendedTwo}{0.33}

\newcommand{\mnPlantedVisIssCorrectTwo}{100\%}

\newcommand{\mnPlantedVisIssCostTwo}{8.2}

\newcommand{\mnPlantedVisOptTwo}{6.9}
\newcommand{\mnPlantedVisReal}{100\%}

\newcommand{\mnPlantedVisRpCostTwo}{25.3}

\newcommand{\mnShapeAcc}{98.6\%}

\newcommand{\mnShapeNets}{10}

\newcommand{\mnShapeReads}{shape (98\%)}

\newcommand{\mnShapeVisIntendedTwo}{0.01}

\newcommand{\mnShapeVisIssCorrectTwo}{100\%}

\newcommand{\mnShapeVisIssCostTwo}{9.0}

\newcommand{\mnShapeVisOptTwo}{7.0}
\newcommand{\mnShapeVisReal}{100\%}

\newcommand{\mnShapeVisRpCostTwo}{26.4}

\newcommand{\mnThmCoverBetterThanIss}{15}

\newcommand{\mnThmCoverWorseN}{15{,}000}
\newcommand{\mnThmCoverWorseThanIss}{4{,}887}

\newcommand{\mnThmRealRuns}{53{,}496}

\newcommand{\mcAllHeadResid}{$10^{-4}$}
\newcommand{\mcBypassAcc}{100.0\%}

\newcommand{\mcBypassNets}{3}

\newcommand{\mcBypassVisIntendedHalf}{5\%}

\newcommand{\mcBypassVisIssCorrectHalf}{100\%}

\newcommand{\mcBypassVisIssCostHalf}{2.7}

\newcommand{\mcBypassVisIssNHalf}{7.9}

\newcommand{\mcBypassVisRpCostHalf}{35.4}

\newcommand{\mcBypassVisRsCostHalf}{3.3}

\newcommand{\mcBypassWbIntendedHalf}{4\%}

\newcommand{\mcBypassWbIssCorrectHalf}{63\%}

\newcommand{\mcBypassWbIssCostHalf}{2.7}

\newcommand{\mcBypassWbIssNHalf}{7.9}

\newcommand{\mcBypassWbIssValRefOne}{69\%}

\newcommand{\mcBypassWbIssVoidOne}{97\%}

\newcommand{\mcBypassWbRpCostHalf}{52.6}

\newcommand{\mcBypassWbRpRefutedOne}{97\%}

\newcommand{\mcBypassWbRsCostHalf}{3.4}

\newcommand{\mcCnnAShape}{26\%}

\newcommand{\mcCnnCShape}{3\%}

\newcommand{\mcColorAcc}{100.0\%}

\newcommand{\mcColorInBox}{100\%}
\newcommand{\mcColorNets}{3}

\newcommand{\mcColorVisIntendedHalf}{6\%}
\newcommand{\mcColorVisIntendedOne}{32\%}

\newcommand{\mcColorVisIssCorrectHalf}{100\%}

\newcommand{\mcColorVisIssCostHalf}{2.5}

\newcommand{\mcColorVisIssNHalf}{7.6}

\newcommand{\mcColorVisRpCostHalf}{34.4}

\newcommand{\mcColorVisRsCostHalf}{3.3}

\newcommand{\mcInclassAccMin}{98.6\%}

\newcommand{\mcInclassImages}{150}
\newcommand{\mcInclassInBox}{96\%}

\newcommand{\mcInclassVisFloorOne}{0.16}

\newcommand{\mcInclassVisIssCostHalf}{4.0}

\newcommand{\mcInclassVisIssNHalf}{13.6}

\newcommand{\mcInclassVisRpCostHalf}{33.9}

\newcommand{\mcInclassVisRsCostHalf}{5.3}

\newcommand{\mcMilpsIss}{223}

\newcommand{\mcOutVoid}{12\%}
\newcommand{\mcPlantedAcc}{99.2\%}

\newcommand{\mcPlantedInBox}{95\%}
\newcommand{\mcPlantedNets}{6}

\newcommand{\mcPlantedVisIntendedHalf}{3\%}

\newcommand{\mcPlantedVisIssCorrectHalf}{98\%}

\newcommand{\mcPlantedVisIssCostHalf}{3.7}

\newcommand{\mcPlantedVisIssNHalf}{13.0}

\newcommand{\mcPlantedVisRpCostHalf}{27.1}

\newcommand{\mcPlantedVisRsCostHalf}{4.7}

\newcommand{\mcShapeAAcc}{99.5\%}

\newcommand{\mcShapeAInBox}{93\%}
\newcommand{\mcShapeANets}{3}

\newcommand{\mcShapeAVisIntendedHalf}{2\%}

\newcommand{\mcShapeAVisIssCorrectHalf}{100\%}

\newcommand{\mcShapeAVisIssCostHalf}{4.0}

\newcommand{\mcShapeAVisIssNHalf}{13.4}

\newcommand{\mcShapeAVisRpCostHalf}{43.1}

\newcommand{\mcShapeAVisRsCostHalf}{5.3}

\newcommand{\mcShapeBAcc}{98.7\%}

\newcommand{\mcShapeBInBox}{97\%}
\newcommand{\mcShapeBNets}{3}

\newcommand{\mcShapeBVisIntendedHalf}{0\%}

\newcommand{\mcShapeBVisIssCorrectHalf}{100\%}

\newcommand{\mcShapeBVisIssCostHalf}{6.1}

\newcommand{\mcShapeBVisIssNHalf}{21.1}

\newcommand{\mcShapeBVisRpCostHalf}{38.0}

\newcommand{\mcShapeBVisRsCostHalf}{8.3}

\newcommand{\mcThmRealRuns}{1{,}728}

\newcommand{\mcThmWrong}{0}
\newcommand{\mcTimeIss}{0.3}
\newcommand{\mcTimeIssMax}{7}
\newcommand{\mcTimeRp}{5.9}
\newcommand{\mcTimeRpMax}{263}

\usepackage{hyperref}
\usepackage{url}
\usepackage{amsmath,amssymb,amsthm}
\usepackage{dsfont}
\usepackage{booktabs}
\usepackage{graphicx}
\usepackage{algorithm}
\usepackage{algorithmic}
\usepackage{enumitem}
\usepackage{multirow}
\usepackage{xcolor}
\usepackage{subcaption}

\makeatletter
\newcommand{\ISSexternaldocument}[1]{%
  \let\ISSbibcite\bibcite \let\bibcite\@gobbletwo
  \externaldocument{#1}%
  \let\bibcite\ISSbibcite}
\makeatother

\ifx\ISSskipappendix\undefined\else
  \usepackage{xr}
  \ISSexternaldocument{appendix}
  \newwrite\ISScounters
  \newcommand{\ISSwritecounter}[1]{%
    \immediate\write\ISScounters{\string\setcounter{#1}{\the\value{#1}}}}
  \AtEndDocument{%
    \immediate\openout\ISScounters=\jobname.cnt\relax
    \ISSwritecounter{figure}\ISSwritecounter{table}%
    \ISSwritecounter{equation}\ISSwritecounter{algorithm}%
    \ISSwritecounter{footnote}\ISSwritecounter{theorem}%
    \ISSwritecounter{proposition}\ISSwritecounter{lemma}%
    \ISSwritecounter{corollary}\ISSwritecounter{definition}%
    \ISSwritecounter{example}\ISSwritecounter{assumption}%
    \ISSwritecounter{remark}%
    \immediate\closeout\ISScounters}
\fi
\ifx\ISSskipbody\undefined\else
  \usepackage{xr}
  \ISSexternaldocument{paper}
\fi

\theoremstyle{plain}
\newtheorem{theorem}{Theorem}
\newtheorem{proposition}{Proposition}
\newtheorem{lemma}{Lemma}
\newtheorem{corollary}{Corollary}
\theoremstyle{definition}
\newtheorem{definition}{Definition}

\newtheorem{assumption}{Assumption}
\theoremstyle{remark}

\newcommand{\Do}{\mathrm{do}}
\newcommand{\Pa}{\mathrm{Pa}}
\newcommand{\Ans}[1]{A_{#1}}
\newcommand{\sep}{k_{\mathcal{Q},c}}
\newcommand{\cQ}{\mathcal{Q}}
\newcommand{\cU}{\mathcal{U}}
\newcommand{\cW}{\mathcal{W}}

\newcommand{\cH}{\mathcal{H}}

\newcommand{\cV}{V}
\newcommand{\Risk}{R}
\newcommand{\xobs}{x^{\mathrm{obs}}}
\newcommand{\ustar}{u^{\star}}
\newcommand{\Pstar}{P^{\star}}
\newcommand{\ind}[1]{\mathds{1}\!\left[#1\right]}
\newcommand{\cost}{c_{\mathrm{int}}}
\newcommand{\Qfull}{\cQ_{\mathrm{full}}}
\newcommand{\iss}{\textsc{iss}}

\newcommand{\ISStitle}{Certifying Interventional Agreement Among Observationally Equivalent Causal Models}
\title{\ISStitle}

\author{
Sourena Khanzadeh\textsuperscript{1,2},
Daniel Platnick\textsuperscript{2},
Marjan Alirezaie\textsuperscript{1,2},
Hossein Rahnama\textsuperscript{1,2,3}
\\
\textsuperscript{1}Toronto Metropolitan University \\
\textsuperscript{2}Flybits Labs, Creative AI Hub \\
\textsuperscript{3}MIT Media Lab, Massachusetts Institute of Technology \\
Correspondence: \texttt{sourena.khanzadeh@torontomu.ca}
}

\iclrfinalcopy 
\begin{document}

\ifISSbody

\maketitle
\lhead{Preprint}

\begin{abstract}

Observationally equivalent causal models can still disagree about what happens under intervention, because interventions create inputs that never occur in observational data.
We introduce Interventional Separation Selection (\iss{}), which repeatedly queries the true system with an admissible intervention on which the surviving candidate models disagree, discards the candidates the outcome contradicts, and stops once no intervention within a cost bound separates the survivors.
If the true system is among the candidates, this stopping condition certifies that every survivor agrees with it on every admissible intervention within the bound, a guarantee that no observational learner can give, however much data it sees.
The stopping condition depends only on the survivors, so it can be checked without knowing the truth.
For continuous variables the candidates form an infinite version space, and mixed-integer linear programs decide the stopping condition exactly over all of it, with agreement holding up to a tolerance.
On a three-digit colored MNIST causal abstraction task in which ink hue tracks digit size, plain convolutional networks trained on $60{,}000$ examples reach zero held-out error, yet disagree with shape-based labels on \mcCnnAShape{} of single-digit edits, as often as hue-based labels do.
Auditing the causal abstractions of networks observed only on such images, \iss{} certifies what each network perceives with \mcInclassVisIssNHalf{} interventions per image on average, and each certificate, checked against every admissible intervention, holds whenever the network's true abstraction is among the candidates.
When a network bypasses a unit that every candidate abstraction relies on, certificates covering interventions on that unit can be silently void, and twenty random validation interventions refute \mcBypassWbIssValRefOne{} of them.

\end{abstract}

\section{Introduction}
\label{sec:intro}

Machine learning models are typically evaluated by how well they generalize to unseen data. However, two models can produce the same prediction while arriving at it for different causal
reasons. This problem arises naturally under \emph{underspecification}~\citep{damour2022underspecification}: multiple models can fit
the same observational data---and even agree on held-out observations---while relying on different
underlying mechanisms and making different predictions under intervention. For instance, one model
may reach the correct prediction because it captures the intended causal relationship, while another
may reach the same prediction through a shortcut that happens to work on the observed data.
Collecting more data from the same observational regime does not resolve this ambiguity when the
relevant configurations are never observed (Theorem~\ref{thm:floor}).

We refer to a model together with the context explaining an observed outcome as a
\emph{causal perspective}. Different causal perspectives can therefore agree on \emph{what} happened
while encoding different explanations for \emph{why} it happened. These differences become
visible only under intervention.

This motivates a different question:
\begin{quote}
\emph{Can the interventions we run certify a model's behavior under those we do not?}
\end{quote}

We operationalize this question through intervention: rather than requiring identification of a
unique perspective, we ask whether all perspectives still consistent with the evidence agree on every
admissible intervention up to a specified cost. We introduce \emph{Interventional Separation Selection
(ISS)} to provide such a certificate. ISS maintains the set of observationally compatible perspectives
and actively queries interventions on which surviving perspectives disagree. After observing the
environment's response, inconsistent perspectives are eliminated, as displayed in Figure~\ref{fig:scmllm}. The procedure stops only when no
admissible intervention of cost at most $R$ can separate any surviving pair. If the true system is among the initial perspectives, this
stopping condition certifies that every surviving perspective agrees with the true system on all
admissible interventions within radius $R$.

\textbf{Contributions.}
Our main contribution is a framework for \emph{certifying causal reasoning through interventional
generalization under observational underspecification}. We introduce ISS, which actively separates
observationally compatible causal perspectives and provides an exact, cost-bounded certificate of
interventional agreement, and we extend the certificate from Boolean to continuous variables
(Proposition~\ref{prop:continuous}). We apply ISS to the causal abstractions of MNIST classifiers whose ink hue,
internal percepts and output are continuous (Section~\ref{sec:experiments}). Networks that read a digit's shape and
networks that read its ink hue cannot be told apart on the observed images; \iss{} certifies each network with \mcInclassVisIssNHalf{} image edits per image on average, and we check every certificate against every admissible intervention. 
In a Boolean version with no tolerance (Appendix~\ref{app:mnist}), \iss{} never needs more than \mnInclassVisIssNMaxTwo{} image edits to certify radius $2$, the fewest any learner can guarantee in the worst case (Theorem~\ref{thm:cost}).

\begin{figure}
    \centering
    \includegraphics[width=0.5\linewidth]{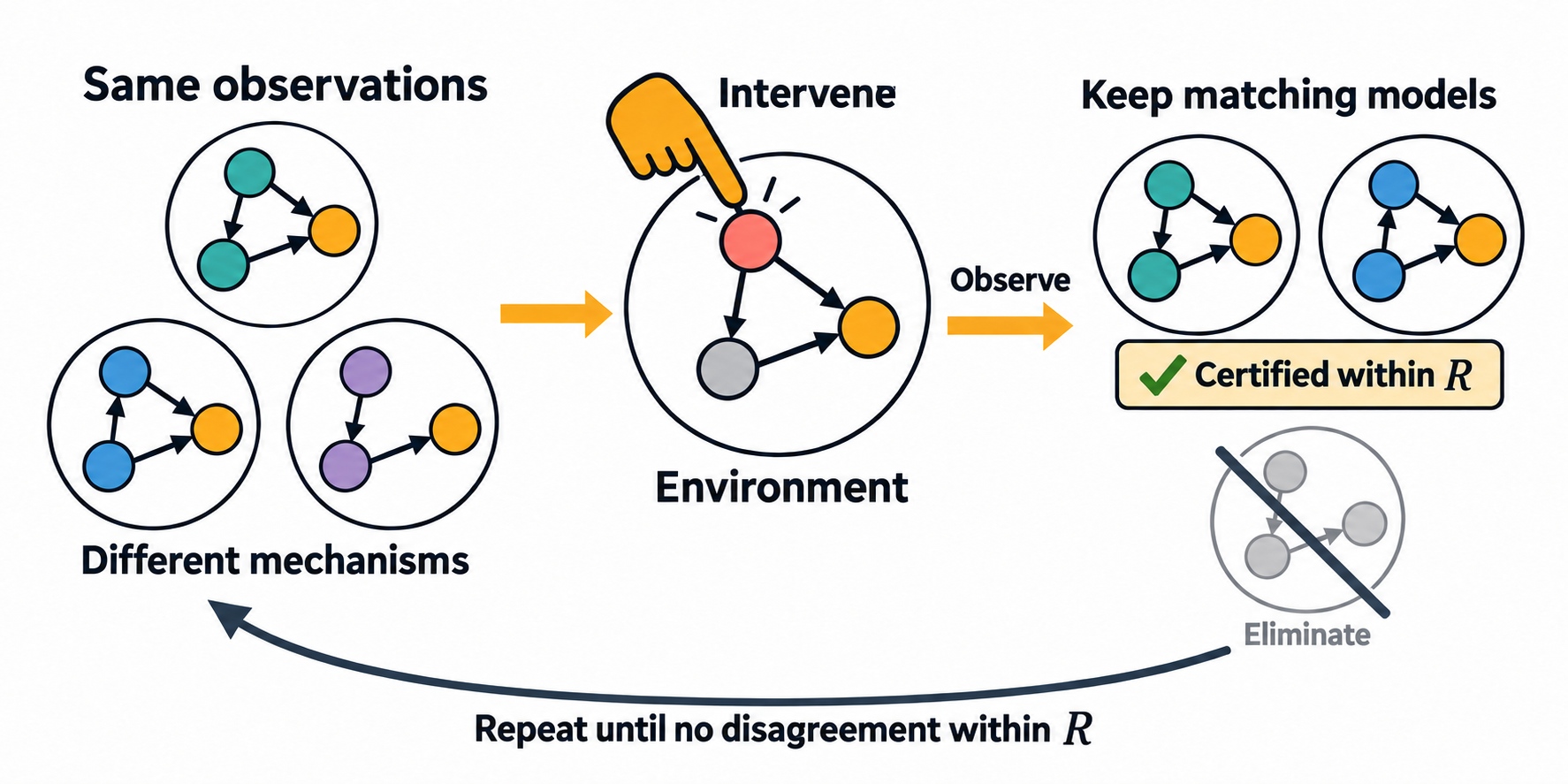}
    \caption{Observationally equivalent models can encode different mechanisms. \iss{} selects a separating intervention (minimum cost by default), observes the environment's answer, and eliminates inconsistent hypotheses. The loop certifies agreement only when no query within radius $R$ separates any surviving pair in the full version space; with exact answers and the truth in the initial space, every survivor then has zero interventional risk within $R$.}
    \label{fig:scmllm}
\end{figure}

\section{Related Work}
\label{sec:related}

\textbf{Underspecification and multiplicity.}
Predictively equivalent models can rely on different mechanisms and therefore behave differently outside the regime on which they were evaluated \citep{damour2022underspecification,marx2020predictive}. Shortcut learning provides a prominent instance of this phenomenon under distribution shift \citep{geirhos2020shortcut,sagawa2020distributionally}. Prior work has used interventions or counterfactual augmentation to expose such differences \citep{dehaan2019causal,kaushik2020learning}. Our focus is not merely to detect multiplicity, but to characterize when the surviving models are guaranteed to agree under every admissible intervention within a specified budget.

\textbf{Experimental design and active causal learning.}
Model-discrimination design chooses experiments that separate competing explanations \citep{box1967discrimination,chaloner1995bayesian}, while active causal discovery chooses interventions to resolve uncertainty about causal structure \citep{eberhardt2007interventions,hauser2014two,squires2020active}; the state of the art amortizes the design policy \citep{tigas2022interventions,annadani2024amortized}, prices the measuring oracle rather than the intervention \citep{zhang2023bayesian}, or plans under an explicit budget \citep{guo2026constrained}, and cost-aware and verification-oriented variants ask how interventions should be allocated under experimental constraints \citep{ghassami2018budgeted,choo2022verification}. \iss{} differs in its target: given a version space of fully specified causal perspectives at a fixed context, it seeks a cost-bounded certificate of interventional equivalence rather than recovery of a unique graph, and its budget indexes that guarantee rather than only constraining the design. Those methods stop when a horizon or budget is spent rather than when nothing remains to separate; an information criterion can express that test, but only if its expectation is exact over the whole version space, which an amortized or sampled posterior does not give. We therefore port their acquisition principle, expected information per unit cost, into our loop (Appendix~\ref{app:mnist}) and discuss the differences in Appendix~\ref{app:design}.

\textbf{Disagreement-based learning and testing.}
Version-space methods and disagreement-based active learning select queries that distinguish surviving hypotheses \citep{mitchell1982generalization,seung1992query,tong2001active}; differential testing similarly searches for inputs on which implementations disagree \citep{mckeeman1998differential,pei2017deepxplore}. \iss{} gives this idea a causal and cost-sensitive form: queries are interventions, disagreement is defined by counterfactual responses, and exhaustion of all separating queries within a radius yields a certificate. This objective also differs from algorithmic recourse, where the goal is to find a low-cost action that changes a particular prediction \citep{ustun2019actionable,karimi2021algorithmic}; here, cost measures how difficult it is to expose a disagreement between hypotheses.

\textbf{Causal abstraction and counterfactual reasoning.}
Causal abstraction evaluates whether interventions on a lower-level system realize the interventions of a higher-level causal model \citep{beckers2019abstracting,geiger2021causal,geiger2023causal}. We use candidate abstractions and alignments as the hypothesis class, so that observationally plausible abstractions can be actively separated and, under realizability, certified over a specified intervention family. Counterfactual inference instead typically assumes a causal model and reasons about outcomes under hypothetical interventions \citep{balke1994counterfactual,oberst2019counterfactual}; our problem is deciding among multiple such models by choosing interventions. Identification theory asks whether causal quantities are determined from available distributions \citep{pearl1995causal,shpitser2008complete}; we ask how much intervention cost is required before the remaining candidate models become indistinguishable for the interventions of interest. Further connections, including query learning, grounding, the computational complexity of causal reasoning, and additional causal-discovery work, are discussed in Appendix~\ref{app:related}.

\section{Interventional Risk and the Observational Floor}
\label{sec:framework}

\textbf{Causal perspectives.}
Let $[n]:=\{1,\dots,n\}$ and $X=(X_1,\dots,X_n)$, $X_j\in\{0,1\}$. An SCM $M=(U,X,F)$ has exogenous variables $U=(U_1,\dots,U_n)$ with finite domains $\cU_j$ and assignments $X_j=f_j(X_{\Pa_M(j)},U_j)$, where $\Pa_M(j)\subseteq[n]\setminus\{j\}$. By acyclicity every context $u\in\cU:=\prod_j\cU_j$ determines a unique solution $x(M,u)$. A \emph{causal perspective} is a pair $P=(M,u)$ with observational state $x(P):=x(M,u)$; two perspectives are \emph{history-equivalent at $\xobs$} when $x(P)=x(P')=\xobs$.

\textbf{Queries and cost.}
An intervention $I=\Do(X_S=a_S)$ replaces the assignments of $X_S$ by constants, giving $M^{I}$; a query $q=(I,T)$ has target $T\subseteq[n]$ and answer $\Ans{P}(q):=x(M^{I},u)_T$. The admissible family $\cQ$ encodes which variables may be manipulated, how many at once and which targets are observable ($\Qfull$: all). Given $\xobs$,
\begin{equation}
\cost(I;\xobs)=\sum_{s\in S}w_s\,\ind{a_s\neq\xobs_s},\qquad w_s>0,
\label{eq:cost}
\end{equation}
prices departure from the observed world; $c(q):=\cost(I_q;\xobs)$, $w_{\min}:=\min_s w_s$, and $\cQ_r:=\{q\in\cQ: 0<c(q)\le r\}$. $r$ is the maximum intervention cost allowed when deciding which queries count as admissible.

\begin{definition}[Separation cost and local equivalence]
The \emph{separating query set} is $\cW_{PP'}:=\{q\in\cQ:\Ans{P}(q)\neq\Ans{P'}(q)\}$ and the \emph{interventional separation cost} is $\sep(P,P'):=\min_{q\in\cW_{PP'}}c(q)$, with $\min\emptyset:=\infty$; $P\equiv^{r}_{\cQ,c}P'$ (\emph{$r$-local equivalence}) when $\Ans{P}(q)=\Ans{P'}(q)$ for every $q\in\cQ_r$.
\end{definition}

\textbf{The learning problem.}
An environment is a true perspective $\Pstar=(M^{\star},\ustar)$. A learner receives $N$ observational samples $S_N$ from a regime over contexts and the observed state $\xobs=x(\Pstar)$, may run queries in $\cQ$ and observe $\Ans{\Pstar}$, and outputs $\hat P$ from a class $\cH$ of perspectives history-equivalent at $\xobs$.

\begin{definition}[Interventional risk]
For $r$ with $\cQ_r\neq\emptyset$ and $\pi_r$ uniform on $\cQ_r$, $\Risk_r(P,P'):=\Pr_{q\sim\pi_r}[\Ans{P}(q)\neq\Ans{P'}(q)]$ and $\Risk_r(P):=\Risk_r(P,\Pstar)$.
\end{definition}
$\Risk_r$ is a normalized Hamming distance between answer vectors, hence a pseudo-metric.

\begin{proposition}[Separation cost is the margin]
\label{prop:margin}
For all $P,P'$ history-equivalent at $\xobs$ and all $r$ with $\cQ_r\ne\emptyset$: $\Risk_r(P,P')=0$ if $r<\sep(P,P')$ and $\Risk_r(P,P')>0$ if $r\ge\sep(P,P')$; hence $P\equiv^{r}_{\cQ,c}P'$ iff $\Risk_r(P,P')=0$ iff $\sep(P,P')>r$.
\end{proposition}

Proofs are in Appendix~\ref{app:proofs}. Separation cost is symmetric, monotone in $\cQ$ and $c$, at least $w_{\min}$ for history-equivalent perspectives (no free query is informative), and satisfies the ultrametric inequality $\sep(P,P'')\ge\min\{\sep(P,P'),\sep(P',P'')\}$, so $\equiv^{r}_{\cQ,c}$ is an equivalence relation whose classes refine as $r$ grows (Proposition~\ref{prop:basic}, Appendix~\ref{app:extra}). This makes identification well posed: \emph{identifying the truth's class is the most any learner can achieve at budget $r$---it cannot distinguish within a class---and by Proposition~\ref{prop:margin} it is enough.}

\textbf{The observational version space.}
Write each mechanism as $f_j(x_{\Pa(j)},u_j)=g_j(x_{\Pa(j)})\oplus u_j$, $u_j\in\{0,1\}$, and call $(j,z)$, $z\in\{0,1\}^{|\Pa(j)|}$, a \emph{mechanism cell}. A learner that knows the parent sets and the noise form must estimate $g_j(z)$ for every cell.
\begin{assumption}[Observational regime]
\label{ass:regime}
Contexts are independent and identically distributed (i.i.d.); roots satisfy $X_j=U_j$ and non-root noise bits are $1$ with probability $p<1/2$.
\end{assumption}
\begin{proposition}[Zero-training-error version space]
\label{prop:vspace}
Under Assumption~\ref{ass:regime} with $p=0$, let $\mathrm{sup}(S_N)$ be the set of cells that appeared in the dataset $S_N$. The set $\cV_0(S_N)$ of hypotheses with zero training error is $\{g:\ g_j(z)=g^{\star}_j(z)\ \forall (j,z)\in\mathrm{sup}(S_N)\}$, a product over the unobserved cells, and every member has the same likelihood. The $m_N$ unobserved cells split into $m_N-m_\infty$ cells the regime reaches with positive probability but have not yet occurred in the dataset, and $m_\infty$ cells it reaches with probability zero; $|\cV_0(S_N)|=2^{m_N}$, $m_N\downarrow m_\infty$ almost surely, and only the completions of the $m_\infty$ permanent cells are indistinguishable on held-out observational data. For $p>0$ the maximum-likelihood fit is the majority value on each observed cell and unobserved cells remain free.
\end{proposition}
By Proposition~\ref{prop:freedom} every member of $\cV_0$ abduces a context at $\xobs$: identical observational, different interventional behavior.

\begin{theorem}[Observational floor]
\label{thm:floor}
Let $L$ be any (possibly randomized) learner with $\hat P=L(S_N)$. For every $r$ with $\cQ_r\ne\emptyset$,
\begin{equation}
\sup_{\Pstar\in\cV_0(S_N)}\ \mathbb{E}\,\Risk_r(L(S_N);\Pstar)\ \ge\ \tfrac12\max_{P,P'\in\cV_0(S_N)}\Risk_r(P,P').
\label{eq:floor}
\end{equation}
The right side depends on $S_N$ only through $\mathrm{sup}(S_N)$, so it does not vanish as $N\to\infty$, and it is positive whenever two completions of the $m_\infty$ permanently unobserved cells are separated by some $q\in\cQ_r$.
\end{theorem}
The bound is about the problem, not an estimator: it holds for every observational learner at every $N$. Held-out data can still realize a missing finite-sample cell; for the completions of the $m_\infty$ permanent cells no held-out evaluation detects the residual risk, since they behave identically on observations.

\section{The Certificate, and Making It Exact}
\label{sec:iss}

\begin{definition}[Version-space separation cost]
For a set $\cV$ of perspectives, $\kappa(\cV):=\min_{P\ne P'\in \cV}\sep(P,P')$ ($\infty$ when all members are interventionally equivalent, in particular when $|\cV|\le1$).
\end{definition}

\begin{algorithm}[t]
\caption{Interventional separation selection (\iss)}
\label{alg:iss}
\begin{algorithmic}[1]
\REQUIRE data $S_N$, state $\xobs$, class $\cH$, family $\cQ$, cost $c$, radius $R$, budget $B$, evidence $(E,\ustar_E)$, level $\delta$
\STATE $\cV\leftarrow$ hypotheses in $\cH$ within likelihood level $\delta$ of the fit on $S_N$, history-equivalent at $\xobs$, with $u_{P,E}=\ustar_E$; spent cost $s\leftarrow0$ \hfill $\triangleright$ $\delta$: Theorem~\ref{thm:soft}; $\delta{=}0$: zero training error
\WHILE{$\kappa(\cV)\le R$ and $s+\kappa(\cV)\le B$}
\STATE $q\leftarrow$ a query in $\cQ_R$ with $c(q)\le B-s$ separating two members of $\cV$ \hfill $\triangleright$ default: $c(q)=\kappa(\cV)$
\STATE run $q$, $s\leftarrow s+c(q)$, observe $y\leftarrow\Ans{\Pstar}(q)$, set $\cV\leftarrow\{P\in\cV:\Ans{P}(q)=y\}$
\ENDWHILE
\STATE \textbf{if} $\cV=\emptyset$ \textbf{return} $s$ and ``refuted'' (no hypothesis fits, so $\Pstar\notin\cV_0$)
\RETURN $\hat P\in\cV$, $s$, and the certified radius: $R$ if $\kappa(\cV)>R$, else $\kappa(\cV)-w_{\min}$
\end{algorithmic}
\end{algorithm}

\begin{theorem}[Soundness and certified zero risk]
\label{thm:sound}
Assume $\Pstar\in\cV_0$ and let Algorithm~\ref{alg:iss} run with any rule that queries a separating query. Then (a) $\Pstar\in\cV$ at every iteration; (b) $|\cV|$ strictly decreases at every iteration, so it halts after at most $|\cV_0|-1$ interventions; (c) if it halts with $\kappa(\cV)>R$ then every $\hat P\in\cV$ satisfies $\hat P\equiv^{R}_{\cQ,c}\Pstar$ and therefore $\Risk_r(\hat P)=0$ for every $r\le R$, and if it halts on the budget the same holds with $R$ replaced by $\kappa(\cV)-w_{\min}$; (d) the halting condition is a function of $\cV$ alone, so (c) is a certificate verifiable without knowing $\Pstar$.
\end{theorem}
The certificate belongs to the stopping rule, not to the query rule: any rule that never wastes a query on a non-separating intervention inherits (a)--(d), the loop guard leaves one affordable (it costs $\kappa(\cV)\le B-s$), and by (a) an empty $\cV$ refutes $\Pstar\in\cV_0$. In our experiments (Section~\ref{sec:experiments} and Appendix~\ref{app:mnist}), every rule certifies correctly wherever the truth lies in the class, including the information-per-cost rule of Bayesian design and a rule that probes at random; the rules differ only in cost. Part (c) is deterministic, not a high-probability bound, whereas an observational learner cannot certify any positive radius (Theorem~\ref{thm:floor}). What stays uncertified is the terminal class $\{P:\Ans{P}=\Ans{\Pstar}\text{ on }\cQ_R\}$: membership follows if the induced mechanisms $\varphi^{P}_j(x):=f_j(x_{\Pa_M(j)},u_j)$ agree with $\Pstar$'s on every state some query in $\cQ_R$ realizes (Corollary~\ref{cor:survive}, Appendix~\ref{app:extra}), but not only then, since a difference on a realized cell survives when every query of cost at most $R$ masks it at the target; both kinds of survivor carry risk beyond the radius.

\textbf{How much does certification cost?}
The minimum-cost rule pays $\kappa(\cV_t)$ at step $t$, non-decreasing along the run since $\cV_{t+1}\subseteq\cV_t$, which is why it is the default. The lower bounds hold for \emph{every} learner, adaptive or not; note their quantifiers.

\begin{theorem}[Cost of certification]
\label{thm:cost}
Assume $\Pstar\in\cV_0$, let $T$ be the number of \iss{} interventions and $C:=|\cV_0/\!\equiv^{R}|$ the number of $R$-local classes. Then
\begin{enumerate}[label=(\alph*),nosep,leftmargin=*]
\item (Upper bound, minimum-cost rule) the total cost is $\sum_{t<T}\kappa(\cV_t)\le T\cdot\kappa(\cV_{T-1})\le(|\cV_0|-1)R$.
\item (Lower bounds) Any learner that halts and correctly certifies radius $R$ in environment $\Pstar$ spends at least $k^{\star}(\Pstar):=\max\{\sep(\Pstar,P'):P'\in\cV_0,\ \sep(\Pstar,P')\le R\}$, hence in the worst case over $\Pstar\in\cV_0$ at least $\max\{\sep(P,P'):P,P'\in\cV_0,\sep(P,P')\le R\}$; with a binary target its answer strings over the $C$ classes form a prefix-free code, so in the worst case over $\Pstar$ it runs at least $\lceil\log_2 C\rceil$ interventions.
\item (Greedy approximation, non-adaptive) Let $\mathrm{OPT}$ be the minimum total cost of a set of queries in $\cQ_R$ separating every pair in $\cV_0$ of separation cost at most $R$. Greedily adding the query maximizing newly separated pairs per unit cost yields such a set of cost at most $(1+\ln\binom{|\cV_0|}{2})\mathrm{OPT}$, and running it certifies radius $R$.
\end{enumerate}
\end{theorem}
Part (b) bounds the cost of \emph{any} learner in a given environment by the hardest hypothesis it must refute; part (c) bounds \emph{non-adaptive} certification, which adaptive learning can undercut (Appendix~\ref{app:mnist}).

\textbf{The acquisition step is computationally hard.}
Deciding whether a separating query of cost at most $r$ exists is NP-complete, even for Boolean SCMs with in-degree $2$, unit intervention costs and a single target. Bounded intervention size makes it tractable, since all candidates can then be enumerated in polynomial time; so when every variable is manipulable, deciding \emph{whether} two hypotheses can be separated is easy, while finding the \emph{cheapest} separating intervention can still be hard. Algorithm~\ref{alg:iss} handles this exactly by cost-ordered enumeration.

\textbf{The full version space, implicitly.}
The version space has $|\cV_0|=2^{m_N}$ members, exponentially many in the number of unobserved cells, and a certificate over a \emph{sample} of it is not a certificate. Under Proposition~\ref{prop:vspace} the answer of $P$ to $q$ is a Boolean function $f_q$ of the $m:=m_N$ unobserved cells, and propagating $q$ while branching on every unobserved cell met on the way writes $f_q$ as a DNF whose cubes are the branches. An answer $y$ to $q$ thus adds the constraint $f_q=y$, which can be written exactly as linear inequalities over binary variables, with one indicator variable per cube that answers $y$. After $t$ answers the version space is the set of completions consistent with all $t$ constraints, and each new answer appends its inequalities to the existing system.

\begin{proposition}[Exact acquisition over the implicit version space]
\label{prop:implicit}
(i) $f_q$ has at most $2^{n'}$ cubes, $n'$ the number of non-root variables, one per trajectory of the propagation. (ii) A query separates two members of $\cV_t$ iff two of its cubes with different answers are each consistent with the constraints---one feasibility check per cube, a 0-1 integer program with the cube's literals fixed. (iii) Scanning $\cQ_R$ in cost order therefore yields $\kappa(\cV_t)$, a minimum-cost separating query and the certificate of Theorem~\ref{thm:sound}(c), exact over the full $\cV_0$, in $O(|\cQ_R|\max_q|f_q|)$ feasibility checks per step. (iv) A uniform sample of $\cV_0$ conditioned on the answers of \emph{fixed} queries is a uniform sample of $\cV_t$, but in general not when the queries were chosen as a function of the sample; so the sample serves only acquisition (tie-breaking, split scores), and every risk estimate is computed on a fresh, independent uniform sample of $\cV_t$ drawn by enumeration or rejection.
\end{proposition}
\textbf{Noisy mechanisms: a soft certificate.}
When $p>0$ the maximum-likelihood fit can be wrong on an observed cell, $\Pstar\notin\cV_0$, and Theorem~\ref{thm:sound} is void. The remedy is to keep every hypothesis whose likelihood is within a calibrated margin of the fit. For a fitted cell $z$ of variable $j$ with $n_z$ samples of which $k_z$ are ones, let $d_z:=|2k_z-n_z|$; flipping the fitted value of $z$ costs $d_z\,\ell$ in log-likelihood, $\ell:=\log\frac{1-p}{p}$, so the truth's deficit on variable $j$ is $D_j:=\ell\sum_{z\in j:\ \mathrm{fit}_z\ne g^{\star}_j(z)} d_z$.
\begin{theorem}[Soft certificate]
\label{thm:soft}
Under Assumption~\ref{ass:regime} with known $p\in(0,\tfrac12)$, conditional on the cell counts $(n_z)_{z\in j}$ of variable $j$, $D_j/\ell$ is distributed as $\sum_{z\in j}\max(n_z-2K_z,0)$ with independent $K_z\sim\mathrm{Bin}(n_z,1-p)$, whatever the noise at other variables. Let $T_j$ be the smallest integer with $\Pr[D_j/\ell>T_j\mid (n_z)_{z\in j}]\le\delta/n'$, with $n'$ the number of non-root variables, and let $\cV_0^{\delta}$ be the set of hypotheses that agree with the fit except on unobserved cells and on fitted cells of each variable $j$ of total margin at most $T_j$. Then $\Pr[\Pstar\in\cV_0^{\delta}]\ge1-\delta$, and \iss{} run on $\cV_0^{\delta}$ certifies correctly---every survivor has zero risk within the certified radius---with probability at least $1-\delta$.
\end{theorem}
The quantile is a convolution of capped binomials computed from the observed counts, the soft version space is again a product structure amenable to Proposition~\ref{prop:implicit} (flip variables with an allowed-set constraint), and the price of the guarantee is the budget needed to eliminate the flipped hypotheses.

\textbf{Continuous variables.}
Nothing in Algorithm~\ref{alg:iss} needs finite domains. Let variables take real values, let $c$ be any departure from $\xobs$ (for example $\sum_s w_s|a_s-\xobs_s|$), and let answers be real vectors. For a tolerance $\varepsilon>0$, a query $q$ \emph{$\varepsilon$-separates} $P$ and $P'$ when $\|\Ans{P}(q)-\Ans{P'}(q)\|_\infty>\varepsilon$, and $\kappa^{\varepsilon}(\cV)$ is the least cost of such a query over pairs in $\cV$. Run Algorithm~\ref{alg:iss} with $\varepsilon$-separation, and after an answer $y$ keep the hypotheses whose answer lies within $\eta$ of $y$, where $\eta$ bounds the error of the observed answers.
\begin{proposition}[Continuous certificate]
\label{prop:continuous}
If $\Pstar\in\cV_0$ and every observed answer lies within $\eta$ of $\Pstar$'s, then $\Pstar$ is never eliminated, and if the loop halts with $\kappa^{\varepsilon}(\cV)>R$, every survivor is within $\varepsilon$ of $\Pstar$ on every $q\in\cQ_R$. If $2\eta<\varepsilon$ and $\cQ_R$ is finite, the loop halts after at most $|\cQ_R|$ interventions.
\end{proposition}
Theorem~\ref{thm:floor} carries over with the risk $\mathbb{E}_{q\sim\pi_r}\|\Ans{P}(q)-\Ans{P'}(q)\|_\infty$, again a pseudo-metric. When answers are linear in a hypothesis's parameters, up to decisions on the sign of a linear function, the version space is a finite union of polytopes and $\varepsilon$-separation is decided exactly by mixed-integer linear programs, which extend the 0-1 programs of Proposition~\ref{prop:implicit} to real-valued parameters (Section~\ref{sec:experiments}).

\textbf{Misspecification and validation.} Every certificate above is conditional on the truth lying in the maintained class. When it does not---the fit is wrong beyond the soft margin, or no candidate abstraction of a network is faithful---the loop still halts once the survivors agree, and its certificate is void \emph{silently}. A \emph{validation phase} compares the survivors' shared predictions on unexecuted queries with the environment on a separate budget of random admissible queries and refutes the class at the first contradiction; it certifies nothing, but it alone can falsify the class.

\section{Exogenous Grounding: Supervision at Zero Intervention Cost}
\label{sec:grounding}

A hypothesis can fit the observed state by \emph{choosing} its context---attributing an observed fact to an exogenous surprise rather than to its mechanism.
\begin{proposition}[Abductive freedom]
\label{prop:freedom}
If $f_j(x_{\Pa(j)},u_j)=g_j(x_{\Pa(j)})\oplus u_j$ for every $j$, then for every structure $g$ and every $\xobs$ exactly one context, $u_j=\xobs_j\oplus g_j(\xobs_{\Pa(j)})$, makes $(M_g,u)$ history-equivalent at $\xobs$; history-equivalence constrains structure not at all.
\end{proposition}
An \emph{exogenous evidence set} $E\subseteq[n]$ with verified values $\ustar_E$ (an audit confirming that a component operated nominally) defines the grounded version space $\cV_0^{E}:=\{P\in\cV_0: u_{P,E}=\ustar_E\}$, whose complement is eliminated at cost $0$; $E\subseteq E'$ implies $\cV_0^{E'}\subseteq\cV_0^{E}$ and $\kappa(\cV_0^{E'})\ge\kappa(\cV_0^{E})$ (Proposition~\ref{prop:mono}). The two are complementary: evidence is blind to structural disagreement, interventions see context disagreement only where it is unmasked, and for every $r$ there are structurally identical hypotheses with separation cost $r$ under the full family and $\infty$ below intervention size $r$ that one verified fact tells apart (Theorems~\ref{thm:decomp} and~\ref{thm:cheaper}, Appendix~\ref{app:extra}): \emph{data constrains mechanisms on the visited configurations, evidence constrains contexts, interventions constrain mechanisms off them}.

\section{Experiments: Continuous Causal Abstraction on MNIST}
\label{sec:experiments}

We test our theory on vision networks with continuous causal variables: ink hue, internal percepts, and output probability. Observational data cannot reveal what the network has learned, but the certificates can still be checked exactly. A Boolean version with red/green ink appears in Appendix~\ref{app:mnist}, and the full continuous protocol in Appendix~\ref{app:cont}.

\begin{figure}[t]
\centering
\includegraphics[width=\linewidth]{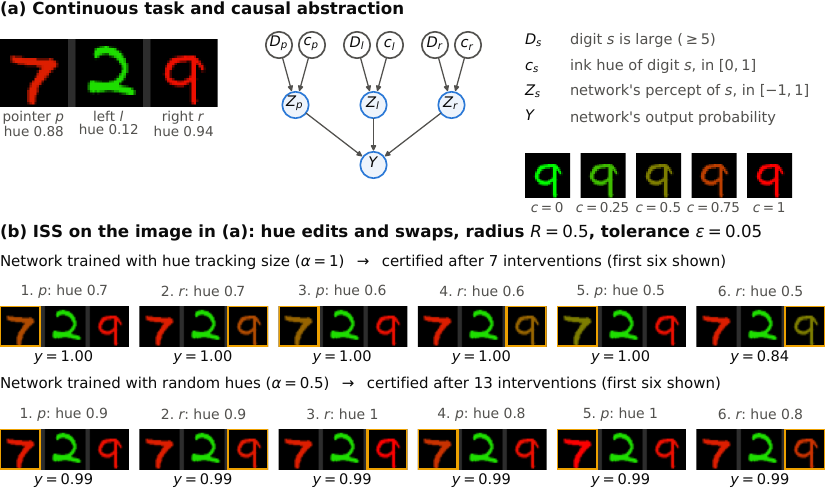}
\caption{\iss{} with continuous variables. (a) An observed image, the high-level causal model and a digit drawn at five hues. (b) \iss{} on this image for two networks, both perfect on observed images. Each step shows the intervention \iss{} chose (edited digits framed; a hue edit sets the digit's hue to the value shown) and the network's output probability $y$. Both certificates hold within $\varepsilon$ on every admissible intervention of cost at most $0.5$.}
\label{fig:cont}
\end{figure}

\textbf{Task.}
Each image contains three MNIST digits \citep{lecun1998gradient}: a pointer $p$, a left digit $l$, and a right digit $r$ (Figure~\ref{fig:cont}a). Each digit has a hue $c\in[0,1]$, from green ($c=0$) to red ($c=1$). A digit is \emph{large} if it is at least $5$ and \emph{small} otherwise; we call this its \emph{size}.
 A small pointer selects the left digit, while a large pointer selects the right. The label is $1$ if the selected digit is large. In Figure~\ref{fig:cont}a, the pointer $7$ selects the $9$, so the label is $1$. This is a version of MNIST pointer-value retrieval \citep{zhang2021pointer}, a benchmark for causal abstraction \citep{geiger2022inducing}.

\textbf{Causal abstraction.}
We describe the network with a small causal model aligned with its internal components \citep{geiger2021causal}. For each digit $s\in\{p,l,r\}$, the model has its size $D_s$, hue $c_s$, and a percept
$Z_s=g_s(D_s,c_s)\in[-1,1]$. The output is
$Y=\sigma(\gamma Z_{\mathrm{sel}}+\beta)$, where $Z_{\mathrm{sel}}=Z_r$ if $Z_p>0$ and $Z_l$ otherwise.

This alignment is exact in our networks: each digit is processed by its own CNN ending in a single $\tanh$ unit representing $Z_s$, and a router passes the selected percept to the output \citep{koh2020concept}. The unknown part is the percept function $g_s$. It may depend on digit shape, hue, or a mixture of both.

\textbf{Why observation cannot decide.}
The auditor only sees images where hue follows digit size: large digits are close to red and small digits are close to green, as in a continuous Colored MNIST shortcut \citep{arjovsky2019invariant}. From $3{,}000$ such images, it can estimate $\gamma$, $\beta$, and constrain the percept curves on observed hues. Outside these hues, the curves remain unconstrained, so the version space $\cV_0$ is infinite. Each hypothesis specifies $72$ percept values, of which $69$ remain free or only bounded by the log.

\textbf{Interventions and certificates.}
A \emph{visual} intervention recolors a digit, with cost equal to the hue change, or swaps it with a digit of the other size, at cost $1$. A \emph{white-box} intervention also allows setting the pointer percept to one of $-1,-0.5,0,0.5,1$ \citep{geiger2021causal}. The network's output probability is the answer.

With tolerance $\varepsilon=0.05$, \iss{} stops when no intervention of cost at most $R$ can make two surviving hypotheses disagree by more than $\varepsilon$. If the truth lies in $\cV_0$, Proposition~\ref{prop:continuous} guarantees that every survivor is within $0.05$ of the network on all such interventions. We compute separation exactly with small mixed-integer linear programs and verify each certificate by testing every admissible intervention.

\textbf{Networks.}
We audit $18$ networks on $10$ observed images each:
\begin{itemize}[nosep,leftmargin=*]
\item \emph{trained} ($9$): hue matches size with probability $\alpha\in\{1,0.9,0.5\}$, with three seeds per setting;
\item \emph{planted} ($6$): each percept uses a known random mix of shape and hue;
\item \emph{bypass} ($3$): the router uses a separate pointer encoder, so none of our candidate abstractions matches the network.
\end{itemize}

\textbf{Result 1: observation cannot reveal the abstraction.}
Held-out accuracy on observed images cannot tell what a network reads. Every network in the class scores at least \mcInclassAccMin{}, and the hue readers ($\alpha=1$) score highest, \mcColorAcc{}. Yet the intended abstraction mispredicts \mcColorVisIntendedOne{} of their interventions of cost at most $1$ by more than $\varepsilon$ (Table~\ref{tab:cont}). By Theorem~\ref{thm:floor}, in its continuous form, whatever an observational auditor reports is off by at least \mcInclassVisFloorOne{} in expected output probability for some network consistent with its log, averaged over observed images, however large the log.

The same limit binds ordinary learners. Call the intended labeling rule the \emph{shape world}, and the same rule applied to hues, with $c\ge0.5$ counting as large, the \emph{hue world}; the two label every observed image identically. Plain CNNs trained on up to $60{,}000$ observed images reach zero held-out error, yet disagree with the shape world on \mcCnnAShape{} of single-digit edits (Figure~\ref{fig:contplots}a): they learned the hue world. When one digit in ten has a random hue ($\alpha=0.9$), mismatched hues occur in the data, and the error falls with $N$ to \mcCnnCShape{}.

\begin{figure}[t]
\centering
\includegraphics[width=\linewidth]{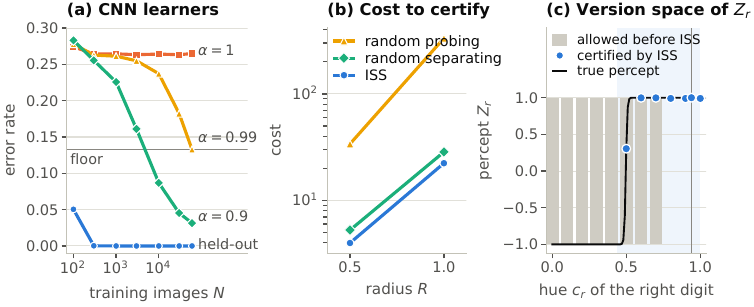}
\caption{(a) Plain CNNs trained on $N$ observed images: held-out error, and error on single-digit edits against the shape world; gray: the floor of Theorem~\ref{thm:floor} for the $\alpha=1$ regime, whose missing cells the $\alpha<1$ learners eventually observe. (b) Cost of certifying radius $R$ with visual interventions, networks in the class. (c) The right digit's percept in the hue reader of Figure~\ref{fig:cont}: what the log allows at each admissible hue, and what \iss{} certified within $R=0.5$ of the observed hue (shaded).}
\label{fig:contplots}
\end{figure}

\textbf{Result 2: certificates are correct and cheap.}
Figure~\ref{fig:cont}b shows \iss{} on one image for two networks, both perfect on observed images. At each step \iss{} runs the cheapest intervention on which survivors disagree by more than $\varepsilon$ and keeps those that predict the answer. For the hue reader it moves the pointer's and the right digit's hues down toward $0.5$, where this network's percepts switch. The shape reader's percepts vary from one handwriting to another, so the log bounds them only loosely even at observed hues, and \iss{} pins this digit's percepts at each hue within reach. Figure~\ref{fig:contplots}c shows the outcome for one percept: every admissible hue within the radius gets a certified value on the true curve, and beyond the radius the version space still allows anything.

The truth lies in $\cV_0$ at \mcInclassInBox{} of observed images. On all of them every run ends with a certificate, and every certificate holds within $\varepsilon$ on every admissible intervention: \mcThmRealRuns{} runs, without exception. At the other images the log's bounds miss some digit's percept, so the class is misspecified, and \mcOutVoid{} of the certificates issued there are void (Appendix~\ref{app:cont}). At $R=0.5$ with visual interventions, \iss{} uses \mcInclassVisIssNHalf{} interventions of total cost \mcInclassVisIssCostHalf{}. Random separating queries cost \mcInclassVisRsCostHalf{}, and random probing, with the same stopping test, \mcInclassVisRpCostHalf{} (Figure~\ref{fig:contplots}b).

\begin{table}[t]
\caption{Auditing $18$ networks at $R=0.5$ (visual interventions unless marked; means over $10$ observed images each). \emph{In class}: images at which the true percepts lie in $\cV_0$. \emph{Intended risk}: admissible interventions on which the intended abstraction is off by more than $\varepsilon$. \emph{\iss{}}: interventions and total cost; \emph{Random}: total cost. \emph{Correct}: certificates within $\varepsilon$ of the network on every admissible intervention, as all are where the truth lies in $\cV_0$.}
\label{tab:cont}
\centering\small
\setlength{\tabcolsep}{4.5pt}
\begin{tabular}{lcccccccc}
\toprule
& Held-out & In & Intended & \multicolumn{2}{c}{\iss{}} & Random & Random & \\
\cmidrule(lr){5-6}
Networks ($n$) & accuracy & class & risk & int. & cost & sep. & probing & Correct\\
\midrule
Trained, $\alpha=1$ (\mcColorNets) & \mcColorAcc & \mcColorInBox & \mcColorVisIntendedHalf & \mcColorVisIssNHalf & \mcColorVisIssCostHalf & \mcColorVisRsCostHalf & \mcColorVisRpCostHalf & \mcColorVisIssCorrectHalf\\
Trained, $\alpha=0.9$ (\mcShapeANets) & \mcShapeAAcc & \mcShapeAInBox & \mcShapeAVisIntendedHalf & \mcShapeAVisIssNHalf & \mcShapeAVisIssCostHalf & \mcShapeAVisRsCostHalf & \mcShapeAVisRpCostHalf & \mcShapeAVisIssCorrectHalf\\
Trained, $\alpha=0.5$ (\mcShapeBNets) & \mcShapeBAcc & \mcShapeBInBox & \mcShapeBVisIntendedHalf & \mcShapeBVisIssNHalf & \mcShapeBVisIssCostHalf & \mcShapeBVisRsCostHalf & \mcShapeBVisRpCostHalf & \mcShapeBVisIssCorrectHalf\\
Planted (\mcPlantedNets) & \mcPlantedAcc & \mcPlantedInBox & \mcPlantedVisIntendedHalf & \mcPlantedVisIssNHalf & \mcPlantedVisIssCostHalf & \mcPlantedVisRsCostHalf & \mcPlantedVisRpCostHalf & \mcPlantedVisIssCorrectHalf\\
\addlinespace[2pt]
Bypass (\mcBypassNets) & \mcBypassAcc & -- & \mcBypassVisIntendedHalf & \mcBypassVisIssNHalf & \mcBypassVisIssCostHalf & \mcBypassVisRsCostHalf & \mcBypassVisRpCostHalf & \mcBypassVisIssCorrectHalf\\
\quad white-box & & & \mcBypassWbIntendedHalf & \mcBypassWbIssNHalf & \mcBypassWbIssCostHalf & \mcBypassWbRsCostHalf & \mcBypassWbRpCostHalf & \mcBypassWbIssCorrectHalf\\
\bottomrule
\end{tabular}
\end{table}

\textbf{Result 3: a wrong class is caught only by validation.}
The bypass networks separate behavior from abstraction. Their pointer's percept and their router both read the pointer's hue and agree on every observed image, so the log fits the class exactly. With visual interventions, \iss{} certifies their behavior correctly: the class can express a router that reads the pointer's hue. With interchange interventions no candidate matches, because setting the pointer's percept changes nothing. Yet at $R=1$, \mcBypassWbIssVoidOne{} of white-box runs end with a void certificate: \iss{} only asks where the candidates disagree, and they all agree about what the pointer's percept does. Twenty random validation interventions refute \mcBypassWbIssValRefOne{} of these certificates, and random probing, which ignores disagreement, refutes the class in \mcBypassWbRpRefutedOne{} of runs.

\section{Discussion and Conclusion}
\label{sec:conclusion}

Choosing among observationally equivalent causal hypotheses is an interventional learning problem: interventional risk is the loss, minimum separation cost the margin, and the budget fixes what can be certified. Intervening buys a \emph{certificate}, not a better estimate: once no admissible intervention within radius $R$ separates the survivors, every survivor agrees with the true system on all such interventions, provided the truth lies in the version space. The guarantee belongs to the stopping condition, not the acquisition heuristic, and it needs the full version space. Under noise it becomes probabilistic; under misspecification only a separate validation phase can falsify the class. 

\textbf{Limitations} exact separation needs finitely many admissible interventions (our hues lie on a grid) and answers linear in the parameters up to sign decisions; the worst-case cost is exponential; parent sets and noise form are given; validation needs $\ln(1/\beta)/\rho$ queries to catch a void class of risk $\rho$ with probability $1-\beta$; and our networks expose their abstraction through one unit per digit, whereas distributed representations would first need a learned alignment \citep{geiger2024finding}.

\subsubsection*{AI use statement}
Generative AI tools were used to draft text, to write and debug the experimental code. The authors take responsibility for the final content of this work, including text, claims, and artifacts produced with the aid of generative AI. Further we used Generative AI as an aid for proofs and mathematical theory. Finally we used AI to polish some of our sentences for readability.



\bibliography{references}
\bibliographystyle{iclr2027_conference}

\fi 

\ifISSappendix

\ifISSbody\else
\title{\ISStitle\\[1.5ex]{\LARGE\sc Appendix}}
\maketitle
This document is the appendix of \emph{\ISStitle}. It is numbered to continue the main text: cross-references without a letter, such as Theorem~\ref{thm:sound}, Algorithm~\ref{alg:iss} or Section~\ref{sec:experiments}, point into the main paper, while Appendices~\ref{app:proofs}--\ref{app:related} and every table and figure cited below are part of this document. Appendix~\ref{app:proofs} proves every claim made in the main text.
\makeatletter
\renewcommand{\addcontentsline}[3]{%
  \addtocontents{#1}{\protect\contentsline{#2}{#3}{\thepage}{\@currentHref}}}
\makeatother
\setcounter{tocdepth}{2}
\tableofcontents
\newpage
\fi

\appendix
\ifISSbody\else
\IfFileExists{paper.cnt}{\input{paper.cnt}}{\GenericWarning{}{main.tex Warning: paper.cnt not found, so the appendix numbering will not match the main text; run `make split'}}
\fi
\section{Proofs}
\label{app:proofs}

\subsection{Proof of Proposition~\ref{prop:margin}}
By definition $\Risk_r(P,P')=|\cW_{PP'}\cap\cQ_r|/|\cQ_r|$. If $r<\sep(P,P')$ then every $q\in\cW_{PP'}$ has $c(q)\ge\sep(P,P')>r$, so $\cW_{PP'}\cap\cQ_r=\emptyset$ and $\Risk_r=0$. If $r\ge\sep(P,P')$ then a minimizing $q^{\star}\in\cW_{PP'}$ has $0<w_{\min}\le c(q^{\star})=\sep(P,P')\le r$ by Proposition~\ref{prop:basic}(c), which applies because $P$ and $P'$ are history-equivalent, so $q^{\star}\in\cW_{PP'}\cap\cQ_r$ and $\Risk_r>0$. The three stated equivalences follow, the last one because $P\equiv^{r}_{\cQ,c}P'$ was defined as agreement on all of $\cQ_r$. History-equivalence is needed: otherwise a cost-zero query can separate $P$ and $P'$, giving $\sep(P,P')=0\le r$, while $\cQ_r$, which excludes cost-zero queries, may contain no separating query. \qed

\subsection{Proof of Proposition~\ref{prop:vspace}}
With $p=0$ every non-root noise bit is $0$, so each sample $x$ satisfies $x_j=g^{\star}_j(x_{\Pa(j)})$ for every non-root $j$; a hypothesis $g$ has zero training error exactly when $g_j(x_{\Pa(j)})=x_j$ for every sample and every $j$, that is, when $g_j(z)=g^{\star}_j(z)$ for every cell $(j,z)$ realized by some sample. Cells outside $\mathrm{supp}(S_N)$ appear in no constraint, so $\cV_0(S_N)$ is the product of $\{g^{\star}_j(z)\}$ over observed cells with $\{0,1\}$ over the $m_N$ unobserved ones, of size $2^{m_N}$. Every member has zero empirical error and therefore the same likelihood. Since $\mathrm{supp}(S_N)\subseteq\mathrm{supp}(S_{N+1})$ and both are contained in the set of cells of positive probability under the regime, $m_N$ is non-increasing and converges to the number $m_\infty$ of cells of probability zero.

For $p>0$ the log-likelihood of a hypothesis factorizes over cells: a cell $(j,z)$ observed $n_z\ge1$ times with $k_z$ ones contributes $k_z\log\rho+(n_z-k_z)\log(1-\rho)$ with $\rho=p$ if $g_j(z)=0$ and $\rho=1-p$ if $g_j(z)=1$. Since $p<1/2$ this is maximized by the majority value, independently across cells, and cells with $n_z=0$ contribute nothing. \qed

\subsection{Proof of Theorem~\ref{thm:floor}}
Fix $S_N$ and abbreviate $\cV_0:=\cV_0(S_N)$. For each $r$, $\Risk_r(\cdot,\cdot)$ is a normalized Hamming distance between the answer vectors $(\Ans{P}(q))_{q\in\cQ_r}$, hence a pseudo-metric; in particular it satisfies the triangle inequality. Let $P,P'\in\cV_0$ be arbitrary and put $\delta:=\Risk_r(P,P')$.

A learner using only observational data sees the same input $S_N$ whether the environment is $P$ or $P'$, because both have zero training error on $S_N$ and are therefore consistent with it; its output distribution is the same in the two cases. Writing $\hat P=L(S_N)$ for a draw from that distribution,
\[
\mathbb{E}\,\Risk_r(\hat P,P)+\mathbb{E}\,\Risk_r(\hat P,P')\;=\;\mathbb{E}\big[\Risk_r(\hat P,P)+\Risk_r(\hat P,P')\big]\;\ge\;\Risk_r(P,P')=\delta ,
\]
so at least one of the two expectations is at least $\delta/2$. Since both $P$ and $P'$ lie in $\cV_0$, the supremum over $\Pstar\in\cV_0$ of $\mathbb{E}\,\Risk_r(L(S_N);\Pstar)$ is at least $\delta/2$. Taking the maximum over $P,P'\in\cV_0$ gives \eqref{eq:floor}, and the second inequality is the definition of $\Risk_r$ for the particular pair.

By Proposition~\ref{prop:vspace}, $\cV_0$ is determined by $\mathrm{supp}(S_N)$ alone, so the right-hand side is a function of the observational support. Any two completions $P,P'$ that agree on all observed cells and are separated by some $q\in\cQ_r$ give $\Risk_r(P,P')\ge 1/|\cQ_r|>0$. If such a pair exists among the completions of the $m_\infty$ cells that the regime reaches with probability zero, then it remains in $\cV_0(S_N)$ almost surely for every $N$, and the bound does not vanish. \qed

\subsection{Proof of Theorem~\ref{thm:sound}}
(a) At each iteration the algorithm keeps exactly those $P$ with $\Ans{P}(q)=y=\Ans{\Pstar}(q)$, a condition $\Pstar$ satisfies; and $\Pstar\in\cV_0$ by assumption, the initial $\cV$ being $\cV_0$ intersected with the evidence constraint, which $\Pstar$ also satisfies because $\ustar_E$ are its own exogenous values. In particular $\cV\ne\emptyset$ throughout, so the refutation branch of Algorithm~\ref{alg:iss} is reached only if $\Pstar\notin\cV_0$.

(b) The chosen $q$ separates two members $P,P'\in\cV$, so $\Ans{P}(q)\ne\Ans{P'}(q)$ and at least one of them differs from $y$ and is removed; $\Pstar$ is never removed by (a), so $1\le|\cV_{t+1}|<|\cV_t|$ and the loop runs at most $|\cV_0|-1$ times.

(c) Halting with $\kappa(\cV)>R$ means no pair in $\cV$ has separation cost at most $R$, i.e.\ no query in $\cQ_R$ separates any two members. By (a) $\Pstar\in\cV$, so for every $\hat P\in\cV$ and every $q\in\cQ_R$ we have $\Ans{\hat P}(q)=\Ans{\Pstar}(q)$, which is the definition of $\hat P\equiv^{R}_{\cQ,c}\Pstar$. Proposition~\ref{prop:margin} then gives $\Risk_r(\hat P)=0$ for all $r\le R$, since $\cQ_r\subseteq\cQ_R$. If instead it halts on the budget, let $r:=\kappa(\cV)-w_{\min}<\kappa(\cV)$. Every $\hat P\in\cV$ equals $\Pstar$ or satisfies $\sep(\hat P,\Pstar)\ge\kappa(\cV)>r$, since $\Pstar\in\cV$, so $\hat P\equiv^{r}_{\cQ,c}\Pstar$ by Proposition~\ref{prop:margin}, and the same argument applies with $R$ replaced by $r$.

(d) $\kappa(\cV)$ is computed from the answers of the members of $\cV$ to admissible queries, which the learner can evaluate; no property of $\Pstar$ beyond the observed answers $y$ is used. \qed

\subsection{Proof of Theorem~\ref{thm:cost}}
(a) At step $t$ the algorithm selects a minimum-cost query separating two members of $\cV_t$, whose cost is $\min_{P\ne P'\in\cV_t}\min_{q\in\cW_{PP'}}c(q)=\kappa(\cV_t)$. Since $\cV_{t+1}\subseteq\cV_t$, the minimum over the smaller pair set is no smaller, so $\kappa(\cV_t)$ is non-decreasing in $t$ and $\sum_{t<T}\kappa(\cV_t)\le T\kappa(\cV_{T-1})$. The loop condition enforces $\kappa(\cV_t)\le R$, and $T\le|\cV_0|-1$ by Theorem~\ref{thm:sound}(b).

(b) \emph{Truth-specific cost bound.} Fix the environment $\Pstar$ and any $P'\in\cV_0$ with $\sep(\Pstar,P')\le R$. Run the learner in environment $\Pstar$; it executes queries $q_1,\dots,q_T$ (chosen adaptively) and halts with a certificate. Suppose no $q_i$ separates $\Pstar$ from $P'$. Then in environment $P'$ the learner receives the same answers $\Ans{P'}(q_i)=\Ans{\Pstar}(q_i)$, hence executes the same queries and halts with the same output and the same certificate. Both $\Pstar$ and $P'$ are consistent with the observational data and with all observed answers, but they are not $R$-locally equivalent, so a certificate asserting that every consistent hypothesis is $R$-locally equivalent to the output is false in at least one of the two environments, contradicting correctness. Hence some $q_i$ separates $\Pstar$ from $P'$ and, by the definition of $\sep$, costs at least $\sep(\Pstar,P')$. Taking the maximum over $P'$ gives the bound $k^{\star}(\Pstar)$. For a randomized learner, fix its random seed: the learner is then deterministic, and the argument applies to every run that halts with a correct certificate. The bound therefore holds run by run, and so does the query count below, seed by seed. For the worst case over $\Pstar\in\cV_0$, let $(P,P')$ attain $\max\{\sep(P,P'):\sep(P,P')\le R\}$ and take $\Pstar=P$.

\emph{Query count.} With a binary target, the run in environment $\Pstar$ produces an answer string $a(\Pstar)\in\{0,1\}^{T(\Pstar)}$. Let $\Pstar,\Pstar{}'$ lie in different $R$-local classes. If $a(\Pstar)$ were a prefix of $a(\Pstar{}')$ (or equal), the learner in environment $\Pstar{}'$ would, after $T(\Pstar)$ answers, have seen exactly the string $a(\Pstar)$, executed the same queries, and halted with the same certificate as in environment $\Pstar$---false for one of the two, as above. Hence the strings $a(\cdot)$ of representatives of the $C$ classes form a prefix-free code with $C$ codewords. By Kraft's inequality $\sum_{i\le C}2^{-T_i}\le1$, so $\max_i T_i\ge\lceil\log_2C\rceil$, and by the source-coding bound the expected length under the uniform distribution over classes is at least $\log_2C$.

(c) Let the ground set be $\Pi:=\{\{P,P'\}\subseteq\cV_0:\sep(P,P')\le R\}$, with $M:=|\Pi|\le\binom{|\cV_0|}{2}$, and for $q\in\cQ_R$ let $\mathrm{cov}(q):=\{\{P,P'\}\in\Pi: q\in\cW_{PP'}\}$ with weight $c(q)>0$. Every pair in $\Pi$ is covered by at least one $q\in\cQ_R$, by the definition of $\sep$. Minimizing $\sum_{q\in\cQ'}c(q)$ over covers $\cQ'$ of $\Pi$ is weighted set cover, and the stated rule is its greedy algorithm, whose output has cost at most $H_{M}\,\mathrm{OPT}\le(1+\ln M)\mathrm{OPT}$ with $H_M=\sum_{i\le M}1/i$. Running all queries of a cover leaves a version space in which no two members are separated by any query of cost at most $R$: if $P,P'$ both survive they agreed with $\Pstar$, hence with each other, on every executed query, and any pair with $\sep(P,P')\le R$ is covered and so was separated. Therefore $\kappa>R$ and the radius is certified by Theorem~\ref{thm:sound}(c). Adaptivity is not used by the cover, and an adaptive learner may spend less than $\mathrm{OPT}$: it skips the queries of the cover that only separate pairs already refuted. \qed

\subsection{Proof of Corollary~\ref{cor:survive}}
Let $q=(\Do(X_S=a_S),T)\in\cQ_R$ and let $x^{\star}$ be the state of $\Pstar$ under $q$; by assumption $x^{\star}$ is realized by a query in $\cQ_R$. We show $x(M^{I},u_P)=x^{\star}$ by induction along a topological order of $M$. For $s\in S$ both equal $a_s$. For $l\notin S$, assume the claim for all predecessors; then $\varphi^{P}_l$ and $\varphi^{\Pstar}_l$ are evaluated at the same state $x^{\star}$, on which they agree by hypothesis, so the $l$-th coordinates agree. Hence $\Ans{P}(q)=\Ans{\Pstar}(q)$ for every $q\in\cQ_R$, so $P$ is never eliminated by any answer, whichever queries the rule chooses. That the condition is not necessary: take $P$ differing from $\Pstar$ at a single cell $(l,z)$ with $l\notin T$ and $l$ not an ancestor of $T$ in either model; the state under a query realizing $z$ differs at $l$ but $x_T$ is unchanged, so $P$ answers every query like $\Pstar$ and survives although a reachable difference exists. The description of the terminal class is Theorem~\ref{thm:sound}(c) read as a set, and $\Risk_r(\hat P)$ for $r>R$ counts the queries on which a member of that class disagrees with $\Pstar$, whichever of the two kinds of difference causes it. \qed

\subsection{Proof of Proposition~\ref{prop:mono}}
If $E\subseteq E'$ and $u_{P,E'}=\ustar_{E'}$ then $u_{P,E}=\ustar_E$, so $\cV_0^{E'}\subseteq\cV_0^{E}$; a minimum over a smaller set of pairs is no smaller, giving $\kappa(\cV_0^{E'})\ge\kappa(\cV_0^{E})$. The bound of Theorem~\ref{thm:cost}(a) is increasing in $|\cV_0|$, which shrinks, and the halting condition $\kappa>R$ is implied for $E'$ whenever it holds for $E$. \qed

\subsection{Proof of Proposition~\ref{prop:implicit}}
(i) Under Proposition~\ref{prop:vspace} all members of $\cV_0$ share the observed cells and the abduced context, so the state of $P$ under $q=(\Do(X_S=a_S),T)$ is obtained by propagating in topological order: intervened variables take their clamped values, roots their context values, and a non-root $j$ the table entry of the cell $(j,z)$ its parents currently realize. If that cell is observed the entry is common to all of $\cV_0$; if it is unobserved the entry is the cell's free bit. Branching on the free bit whenever an unobserved cell is met, and continuing each branch, produces a tree whose leaves are the possible trajectories; a leaf fixes the target's value and is reached by all completions agreeing with the partial assignment read along the way, which is its cube. Each non-root contributes at most one branching per trajectory, so there are at most $2^{n'}$ leaves, and the cubes partition $\cV_0$ (each completion follows exactly one trajectory), so $f_q$ equals the DNF whose cubes with answer $y$ are the leaves labeled $y$.

(ii) $\cV_t=\{b\in\cV_0: f_{q_s}(b)=y_s,\ s\le t\}$. Two members $b,b'$ answer $q$ differently iff they lie in cubes of $f_q$ with different labels, i.e.\ iff two cubes with different labels each contain a member of $\cV_t$. A cube $\kappa$ contains a member of $V_t$ iff the constraints have a 0-1 solution that agrees with the literals of $\kappa$, which is one feasibility check with those literals fixed as variable bounds. Each constraint $\{f_{q_s}=y_s\}$ is encoded with one binary indicator $e_\lambda$ per cube $\lambda$ of $f_{q_s}$ labeled $y_s$: the indicators sum to at least one, and each forces the literals of its cube, with $b_i \ge e_\lambda$ if $\lambda$ sets cell $i$ to 1 and $b_i \le 1-e_\lambda$ if it sets it to 0. Constraints are added as answers arrive.

(iii) $\kappa(\cV_t)$ is the cost of the first separating query in cost order, and the absence of a separating query in $\cQ_R$ is exactly the halting condition $\kappa(\cV_t)>R$ of Theorem~\ref{thm:sound}(c); both are decided by the tests of (ii), at most $|f_q|$ per query, hence $O(|\cQ_R|\max_q|f_q|)$ feasibility checks per step, each on a program whose size is linear in the number of cubes seen so far.

(iv) If $b$ is uniform on $\cV_0$ and $q_1,\dots,q_t$ are fixed, then conditional on $f_{q_s}(b)=y_s$ for $s\le t$, $b$ is uniform on $\cV_t$, so filtering a uniform sample by the answers to fixed queries yields a uniform sample of $\cV_t$. This fails when $q_s$ is chosen as a function of the sample: conditional on the event that a particular query maximized the sample's split, the sample's composition is no longer exchangeable with the rest of $\cV_t$ (elements that make that query the maximizer are over-represented), so statistics averaged over the surviving sample are biased in general. We therefore use the acquisition sample only to choose queries, and evaluate risk on a fresh uniform sample of $\cV_t$, drawn independently: by enumerating the assignments of the variables the constraints mention (the others are free and uniform) when there are at most $22$, otherwise by rejection from uniform draws of $\cV_0$, each draw accepted iff it satisfies every constraint. Since acceptance depends only on the draw, the accepted draws are i.i.d.\ uniform on $\cV_t$. \qed

\subsection{Proof of Proposition~\ref{prop:continuous}}
Each elimination keeps the hypotheses whose answer lies within $\eta$ of the observed one, and $\Pstar$'s answer does, so $\Pstar$ survives every step. Halting with $\kappa^{\varepsilon}(\cV)>R$ means that no $q\in\cQ_R$ $\varepsilon$-separates two members of $\cV$; since $\Pstar\in\cV$, every survivor is within $\varepsilon$ of $\Pstar$ on every $q\in\cQ_R$. For termination, the loop runs only $\varepsilon$-separating queries. Once it has run $q$ and observed $y$, every survivor is within $\eta$ of $y$ at $q$, so any two survivors are within $2\eta<\varepsilon$ there and $q$ never $\varepsilon$-separates again; each query of the finite family $\cQ_R$ is therefore run at most once. The proof of Theorem~\ref{thm:floor} uses only the triangle inequality of the risk and the fact that members of $\cV_0$ cannot be told apart by observation; both hold for $\mathbb{E}_{q\sim\pi_r}\|\Ans{P}(q)-\Ans{P'}(q)\|_\infty$. \qed

\subsection{Proof of Theorem~\ref{thm:soft}}
Order the variables topologically. The cell counts $(n_z)_{z\in j}$ of variable $j$ are determined by the roots and by the noise bits of the ancestors of $j$; the noise bits $u^{(i)}_j$ of the $N$ samples at variable $j$ are i.i.d.\ $\mathrm{Bern}(p)$ and independent of everything at other variables (Assumption~\ref{ass:regime}). Conditional on the counts of $j$---indeed on all roots and all noise at other variables---the samples in cell $z$ have $x^{(i)}_j=g^{\star}_j(z)\oplus u^{(i)}_j$ with i.i.d.\ noise, so the number $K_z$ of samples agreeing with the true value $g^{\star}_j(z)$ is $\mathrm{Bin}(n_z,1-p)$, independently across the (disjoint) cells of $j$. The majority fit is wrong at $z$ iff $K_z<n_z/2$, and then its margin is $n_z-2K_z$; flipping it back to the truth changes the log-likelihood by $(n_z-2K_z)\log\frac{1-p}{p}$, so $D_j/\ell=\sum_{z\in j}\max(n_z-2K_z,0)$ with the stated distribution. (Conditioning on the counts of \emph{other} variables would break this independence, because downstream counts depend on $x_j$; this is why the budgets are per variable.)

Here $j$ ranges over the $n'$ non-root variables: a root satisfies $X_j=U_j$, has no fitted cells and so needs no budget. For each such $j$, $T_j$ is a function of $(n_z)_{z\in j}$ and $\Pr[D_j/\ell>T_j\mid(n_z)_{z\in j}]\le\delta/n'$ by construction; by the tower property $\Pr[D_j/\ell>T_j]\le\delta/n'$ unconditionally, and by the union bound over these $n'$ variables $\Pr[\exists j: D_j/\ell>T_j]\le n'\cdot\delta/n'=\delta$. On the complementary event the truth differs from the fit, on each variable $j$, only on fitted cells of total margin $\sum_z d_z\le T_j$ (and arbitrarily on unobserved cells), i.e.\ $\Pstar\in\cV_0^{\delta}$. On that event Theorem~\ref{thm:sound} applies verbatim to the initial version space $\cV_0^{\delta}$: the truth is never eliminated, and at halting every survivor is $R$-locally equivalent to it. \qed

\subsection{Proof of Proposition~\ref{prop:basic}}
(a) Symmetry holds because $\cW_{ij}=\cW_{ji}$, and $\cW_{ii}=\emptyset$ gives $\sep(P_i,P_i)=\min\emptyset=\infty$.

(b) If $\cQ\subseteq\cQ'$ then the separating set with respect to $\cQ$ is contained in the one with respect to $\cQ'$, so the minimum over the larger set is no larger. If $c\le c'$ pointwise, the minimum of $c$ over the (unchanged) separating set is at most the minimum of $c'$.

(c) Let $q=(I,T)$ with $I=\Do(X_S=a_S)$ and $c(q)=0$; since all $w_s>0$, this means $a_s=\xobs_s$ for every $s\in S$. By history-equivalence, $\xobs$ satisfies every structural equation of $M_i$ under $u_i$. It also satisfies the replaced equations $X_s=a_s=\xobs_s$. Hence $\xobs$ is a solution of $M_i^{I}$ under $u_i$, and since $M_i^{I}$ is acyclic (removing incoming edges preserves acyclicity) its solution is unique, so $x(M_i^{I},u_i)=\xobs$ and $\Ans{P_i}(q)=\xobs_T$. The same argument applies to $P_j$. Therefore no zero-cost query belongs to $\cW_{ij}$, and every $q\in\cW_{ij}$ has $c(q)\ge w_{\min}$.

(d) Let $q\in\cW_{il}$, so $\Ans{P_i}(q)\ne\Ans{P_l}(q)$. Then $\Ans{P_j}(q)$ differs from at least one of $\Ans{P_i}(q)$ and $\Ans{P_l}(q)$, i.e. $q\in\cW_{ij}\cup\cW_{jl}$. Hence $\cW_{il}\subseteq\cW_{ij}\cup\cW_{jl}$ and
\[
\sep(P_i,P_l)=\min_{q\in\cW_{il}}c(q)\ \ge\ \min_{q\in\cW_{ij}\cup\cW_{jl}}c(q)=\min\{\sep(P_i,P_j),\sep(P_j,P_l)\}.
\]
Reflexivity and symmetry of $\equiv^{r}_{\cQ,c}$ are immediate; transitivity follows because $\sep(P_i,P_j)>r$ and $\sep(P_j,P_l)>r$ imply $\sep(P_i,P_l)>r$. If $r\le r'$ then $\equiv^{r'}$ implies $\equiv^{r}$, so the classes of $\equiv^{r'}$ refine those of $\equiv^{r}$. Finally $d=1/\sep$ is symmetric, non-negative, satisfies $d(P,P)=0$, and $d(P_i,P_l)\le\max\{d(P_i,P_j),d(P_j,P_l)\}$ is the inequality just proved. \qed

\subsection{Proof of Lemma~\ref{lem:divergence}}
Let $D:=\{l:x^{i}_l\ne x^{j}_l\}\neq\emptyset$. For $s\in S$ both states equal $a_s$, so $D\cap S=\emptyset$. Choose $l\in D$ that is minimal with respect to a topological order of $M_i$. Every parent $p\in\Pa_{M_i}(l)$ precedes $l$ in that order, hence $p\notin D$ and $x^{i}_p=x^{j}_p$. Since $l\notin S$,
\[
x^{i}_l=\varphi^{P_i}_l(x^{i})=f^{i}_l\big(x^{i}_{\Pa_{M_i}(l)},u_{i,l}\big)=f^{i}_l\big(x^{j}_{\Pa_{M_i}(l)},u_{i,l}\big)=\varphi^{P_i}_l(x^{j}),
\]
while $x^{j}_l=\varphi^{P_j}_l(x^{j})$ because $l\notin S$. As $x^{i}_l\ne x^{j}_l$, we conclude $\varphi^{P_i}_l(x^{j})\ne\varphi^{P_j}_l(x^{j})$. \qed

\subsection{Proof of Theorem~\ref{thm:finite}}
\emph{Infinite separation when induced mechanisms agree.} If $\varphi^{P_i}=\varphi^{P_j}$, Lemma~\ref{lem:divergence} shows that $x(M_i^{I},u_i)=x(M_j^{I},u_j)$ for every intervention $I$, so $\Ans{P_i}(q)=\Ans{P_j}(q)$ for every query and $\cW_{ij}=\emptyset$ for every $\cQ$. Hence $\sep(P_i,P_j)=\infty$ for all $\cQ,c$, and in particular $\Delta=\emptyset$ implies $k_{\Qfull,c}=\infty$.

\emph{Finite separation when they differ.} Let $l\in\Delta$ and $z\in\{0,1\}^n$ with $\varphi^{P_i}_l(z)\ne\varphi^{P_j}_l(z)$. Consider $q=(\Do(X_{R_l}=z_{R_l}),\{l\})\in\Qfull$, where $R_l=\Pa_{M_i}(l)\cup\Pa_{M_j}(l)$; acyclicity gives $l\notin R_l$. In $M_i^{I}$ the variables in $R_l$ take the values $z_{R_l}$, and $l\notin R_l$ takes the value $f^{i}_l(z_{\Pa_{M_i}(l)},u_{i,l})=\varphi^{P_i}_l(z)$; likewise $\Ans{P_j}(q)=\varphi^{P_j}_l(z)$. So $q\in\cW_{ij}$ with $c(q)=\sum_{s\in R_l}w_s\ind{z_s\ne\xobs_s}$. Minimizing over $l\in\Delta$ and admissible $z$ gives the middle bound of \eqref{eq:bound}; the right-hand bound follows from $\ind{\cdot}\le1$ and $w_s\le w_{\max}$, and the left-hand bound is Proposition~\ref{prop:basic}(c).

\emph{Complexity of the witness.} $\varphi^{P_i}_l$ depends only on the coordinates in $\Pa_{M_i}(l)$, so $\varphi^{P_i}_l$ and $\varphi^{P_j}_l$ can be compared on the $2^{|R_l|}\le 2^{2d}$ configurations of $R_l$ by two table look-ups each; doing so for every $l$ costs $O(n2^{2d})$ and yields both $\Delta$ and a minimizing $(l,z)$. \qed

\subsection{Proof of Theorem~\ref{thm:complexity}}
\begin{sloppypar}
(i) A certificate is a query $q=(\Do(X_S=a_S),T)$ with $S\subseteq A$; its cost and both answers are computable in polynomial time by evaluating the two acyclic Boolean networks.

(ii) \emph{$A$ the roots, unit weights.} We reduce from \textsc{Sat}. Let $\phi=C_1\wedge\dots\wedge C_m$ be a CNF over $z_1,\dots,z_p$. If $\phi(0,\dots,0)=1$ output a fixed yes-instance. Otherwise build an SCM with root variables $X_1,\dots,X_p$ ($X_l=U_l$), a NOT gate for each negated literal, a chain of binary OR gates computing each clause, and a chain of binary AND gates computing $\phi$, with output variable $X_{\mathrm{out}}$; all gate mechanisms ignore their exogenous variable. The size is polynomial and every in-degree is at most $2$. Perspective $P_i$ uses these mechanisms; perspective $P_j$ is identical except that $f_{\mathrm{out}}\equiv0$. Take $u_i=u_j=0$, so $\xobs$ has all roots $0$ and $X_{\mathrm{out}}=\phi(0)=0$ under both perspectives: they are history-equivalent, and they agree on every gate under every intervention, so only queries with $\mathrm{out}\in T$ can separate them. Let $A$ be the set of roots and $w\equiv1$. Under $\Do(X_S=a_S)$ with $S\subseteq A$ the roots take a value $z$ with $z_s=a_s$ on $S$ and $z_s=0$ elsewhere, $\Ans{P_i}=\phi(z)$ at $X_{\mathrm{out}}$, $\Ans{P_j}=0$, and the cost is the number of $s\in S$ with $a_s=1$, i.e.\ the Hamming weight of $z$. Hence $k_{\cQ_A,c}\le p$ if and only if $\phi$ is satisfiable.

\emph{$A=[n]$, general weights.} When gates are manipulable, a query may clamp a gate at its observed value at zero cost, which can shortcut the circuit (for instance a satisfied clause gate can be frozen at $1$ while the root that satisfied it is flipped), so a different reduction is needed. We reduce from \textsc{Vertex Cover}. Given a graph $G=(V,E)$ with $|V|=p$ and a budget $r\le p$, let $\phi=\bigwedge_{\{u,v\}\in E}(z_u\vee z_v)$, a \emph{monotone} CNF, and build the circuit above without NOT gates. At $z=0$ every clause gate and every AND gate equals $0$, so $\xobs$ is the all-zero state and the perspectives ($P_j$ with $f_{\mathrm{out}}\equiv0$) are history-equivalent. Set $w_s=1$ for roots and $w_s=p+1$ for gates. If $C\subseteq V$ is a vertex cover of size at most $r$, the query $\Do(X_C=1)$ with target $\mathrm{out}$ costs $|C|\le r$ and separates, since $\phi(\mathbb{1}_C)=1$. Conversely, let $q$ separate at cost at most $r\le p$. No gate can be set to $1$ (cost $p+1$), so every intervened gate is clamped at its observed value $0$. The circuit is monotone, and clamping a gate at $0$ can only decrease the values of downstream gates relative to the unclamped circuit evaluated at the same root values; hence $\Ans{P_i}(q)=1$ at $X_{\mathrm{out}}$ requires $\phi(z)=1$ for the root values $z$ realized by $q$, i.e.\ $\{v:z_v=1\}$ is a vertex cover, and its size equals the cost of $q$, which is at most $r$. Thus $k_{\cQ_{[n]},c}\le r$ if and only if $G$ has a vertex cover of size at most $r$.

(iii) In the \textsc{Sat} construction with $A$ the roots, $k<\infty$ iff $\phi$ is satisfiable. With $A=[n]$, $\cQ_A=\Qfull$ and Theorem~\ref{thm:finite} decides finiteness by comparing induced mechanisms in polynomial time.

(iv) Enumerate all $q\in\cQ$, evaluate both answers, and take the minimum cost over the separating ones. \qed

\end{sloppypar}
\subsection{Proof of Proposition~\ref{prop:freedom}}
Under $f_j(x_{\Pa(j)},u_j)=g_j(x_{\Pa(j)})\oplus u_j$, the equation $\xobs_j=g_j(\xobs_{\Pa(j)})\oplus u_j$ has the unique solution $u_j=\xobs_j\oplus g_j(\xobs_{\Pa(j)})$ for each $j$. With this context, $\xobs$ satisfies all equations, and by uniqueness of solutions $x(M_g,u(g,\xobs))=\xobs$. Since the argument applies to every $g$, every structure is history-equivalent at $\xobs$ for its abduced context. \qed

\subsection{Proof of Theorem~\ref{thm:decomp}}
(a) If $u_i=u_j$ then $u_{i,E}=\ustar_E$ iff $u_{j,E}=\ustar_E$ for every $E$. The characterization of finiteness is Theorem~\ref{thm:finite}.

(b) If $f^{i}_l=f^{j}_l$ and $u_{i,l}=u_{j,l}$ then $\varphi^{P_i}_l=\varphi^{P_j}_l$. Hence when $\Delta_F=\emptyset$, $\varphi^{P_i}_l\ne\varphi^{P_j}_l$ forces $l\in\Delta_u$, and the variable produced by Lemma~\ref{lem:divergence} lies in $\Delta_u$. If $u_i=\ustar$ and $E\cap\Delta_u\ni l$, then $u_{j,l}\ne u_{i,l}=\ustar_l$, so $P_j\notin\Pi_E$.

(c) For $P_i,P_j\in\Pi_{[n]}$ we have $u_i=u_j=\ustar$, so $\varphi^{P_i}_l=f^{i}_l(\cdot,\ustar_l)$ and $\varphi^{P_j}_l=f^{j}_l(\cdot,\ustar_l)$; Theorem~\ref{thm:finite} gives the equivalence. Every answer $\Ans{P}(q)$ of a perspective with context $\ustar$ depends on $f_l$ only through $f_l(\cdot,\ustar_l)$, so differences at other values of $u_l$ affect no observational state and no answer to any query. \qed

\subsection{Proof of Theorem~\ref{thm:cheaper}}
Let $X_l=U_l$ for $l\le r$ and $X_{r+1}=(X_1\vee\dots\vee X_r)\vee U_{r+1}$, all variables Boolean, and $\xobs=(1,\dots,1)$. Let $u_i=(1,\dots,1,0)$ and $u_j=(1,\dots,1,1)$. Both contexts produce $\xobs$, so the perspectives are history-equivalent, share all structural functions, and differ exactly in coordinate $l_0=r+1$.

(i) Since $\Delta_F=\emptyset$, by Theorem~\ref{thm:decomp}(b) any divergence occurs at $X_{r+1}$, so a separating query must have $r+1\in T$ and $r+1\notin S$. Under any intervention, $X_{r+1}=1$ in both perspectives unless $X_1=\dots=X_r=0$. The roots have no parents, so $X_l=0$ for $l\le r$ requires $l\in S$ with $a_l=0\ne\xobs_l$, at cost $1$ each. Hence every separating query costs at least $r$, and $q=(\Do(X_1=\dots=X_r=0),\{r+1\})$ separates ($\Ans{P_i}=0$, $\Ans{P_j}=1$) at cost exactly $r$. Thus $\sep(P_i,P_j)=r$.

(ii) A separating query needs all $r$ roots in its domain, which a family with domain size at most $r-1$ does not admit; so $\cW_{ij}=\emptyset$ and $\sep=\infty$.

(iii) With $E=\{r+1\}$ and $\ustar=u_i$, $u_{j,r+1}=1\ne0=\ustar_{r+1}$, so $P_j\notin\Pi_E$. \qed

\section{Structural Results Used in the Main Text}
\label{app:extra}

\begin{proposition}[Basic properties]
\label{prop:basic}
(a) $\sep$ is symmetric and $\sep(P,P)=\infty$; (b) $\cQ\subseteq\cQ'$ implies $k_{\cQ',c}\le k_{\cQ,c}$, and $c\le c'$ pointwise implies $k_{\cQ,c}\le k_{\cQ,c'}$; (c) for history-equivalent $P,P'$ and $c(q)=0$, $\Ans{P}(q)=\Ans{P'}(q)=\xobs_T$, hence $\sep(P,P')\ge w_{\min}$; (d) $\sep(P,P'')\ge\min\{\sep(P,P'),\sep(P',P'')\}$, so $\equiv^{r}_{\cQ,c}$ is an equivalence relation whose classes refine as $r$ grows.
\end{proposition}

\textbf{Induced mechanisms.} The interventional behavior of $P=(M,u)$ is determined by its induced mechanisms $\varphi^{P}_j(x):=f_j(x_{\Pa_M(j)},u_j)$: under $I=\Do(X_S=a_S)$, $x(M^{I},u)$ is the unique solution of $x_s=a_s$ ($s\in S$) and $x_j=\varphi^{P}_j(x)$ ($j\notin S$).

\begin{lemma}[First divergence]
\label{lem:divergence}
Let $x^{P},x^{P'}$ be the states of $P,P'$ under an intervention with domain $S$. If $x^{P}\neq x^{P'}$, there is $l\notin S$ with $x^{P}_l\ne x^{P'}_l$ and $\varphi^{P}_l(x^{P'})\neq\varphi^{P'}_l(x^{P'})$.
\end{lemma}

\begin{corollary}[A sufficient condition for survival]
\label{cor:survive}
If $\varphi^{P}$ and $\varphi^{\Pstar}$ agree on every state that some query in $\cQ_R$ realizes, then $P$ survives every run of Algorithm~\ref{alg:iss} to radius $R$, whatever the query rule. The condition is sufficient, not necessary: $P$ may differ from $\Pstar$ on a cell that a query in $\cQ_R$ does realize and still survive, because the difference can be \emph{masked}---it changes a variable that does not, for that query, propagate to the target set $T$ (Lemma~\ref{lem:divergence} locates a divergence at some variable, not necessarily in $T$). The terminal class is therefore $\{P:\Ans{P}(q)=\Ans{\Pstar}(q)\ \forall q\in\cQ_R\}$, which contains the hypotheses differing only on cells that no query of cost at most $R$ realizes together with those whose reachable differences are masked at $T$; the risk of $\hat P$ beyond radius $R$ is carried by both.
\end{corollary}

\begin{theorem}[Finite separation under the full family]
\label{thm:finite}
Let $P,P'$ be history-equivalent at $\xobs$, $\Delta:=\{l:\varphi^{P}_l\ne\varphi^{P'}_l\}$ and $R_l:=\Pa_{M}(l)\cup\Pa_{M'}(l)$. Then $k_{\Qfull,c}(P,P')<\infty$ iff $\Delta\neq\emptyset$, and then
\begin{equation}
w_{\min}\;\le\;k_{\Qfull,c}(P,P')\;\le\;\min_{l\in\Delta}\;\min_{z:\,\varphi^{P}_l(z)\ne\varphi^{P'}_l(z)}\;\sum_{s\in R_l}w_s\ind{z_s\ne\xobs_s}\;\le\;w_{\max}\min_{l\in\Delta}|R_l|.
\label{eq:bound}
\end{equation}
With truth tables of in-degree at most $d$, $\Delta$ and a witness attaining the middle bound are computable in $O(n2^{2d})$ time. If $\varphi^{P}=\varphi^{P'}$ then $\sep(P,P')=\infty$ for \emph{every} $\cQ$ and $c$.
\end{theorem}
The middle bound clamps the parents of a differing mechanism to a configuration on which the mechanisms differ; it can be loose, because an upstream intervention may steer several parents at once more cheaply. For restricted families $\cQ\subsetneq\Qfull$ separation can be infinite even when $\Delta\neq\emptyset$; that is when the terminal class of Corollary~\ref{cor:survive} is non-trivial, which is why the experiments compare a visual and a white-box intervention family.

\begin{theorem}[Complexity]
\label{thm:complexity}
Let $\cQ_A$ allow any intervention on a manipulable set $A\subseteq[n]$ with any target. Deciding $k_{\cQ_A,c}(P,P')\le r$ for Boolean SCMs of in-degree at most $2$ is NP-complete: it is in NP; it is NP-hard even with unit weights, a single target and $A$ the roots, and remains NP-hard for $A=[n]$ under general weights. Deciding $k_{\cQ_A,c}<\infty$ is NP-hard for $A$ the roots but polynomial for $A=[n]$ by Theorem~\ref{thm:finite}. If $|\cQ|$ is polynomial in $n$ (interventions of size at most a constant $s$) the problem is in P by enumeration.
\end{theorem}
The contrast is informative: when everything is manipulable, \emph{whether} two hypotheses can be separated is easy but \emph{how cheaply} is hard, because a cheap witness may have to steer several mechanisms through few upstream facts. Cost-ordered enumeration with a size bound $s$ costs $O((2n)^{s}n)$ mechanism evaluations, vectorizable over a whole sample of hypotheses in one pass.

\textbf{Grounding.} For history-equivalent $P,P'$ let $\Delta_F:=\{l:f^{P}_l\ne f^{P'}_l\}$ and $\Delta_u:=\{l:u_{P,l}\ne u_{P',l}\}$.
\begin{proposition}[Monotonicity]
\label{prop:mono}
If $E\subseteq E'$ then $\cV_0^{E'}\subseteq\cV_0^{E}$ and $\kappa(\cV_0^{E'})\ge\kappa(\cV_0^{E})$; the cost bound of Theorem~\ref{thm:cost}(a) and the certified-radius condition can only improve.
\end{proposition}
\begin{theorem}[Complementarity]
\label{thm:decomp}
(a) If $\Delta_u=\emptyset$ then $P\in\cV_0^{E}\iff P'\in\cV_0^{E}$ for every $E$. (b) If $\Delta_F=\emptyset$ then every divergence in Lemma~\ref{lem:divergence} occurs at some $l\in\Delta_u$, and if $P$ is fully grounded any $E$ meeting $\Delta_u$ eliminates $P'$ at zero cost. (c) For fully grounded $P,P'$, $k_{\Qfull,c}<\infty$ iff $f^{P}_l(\cdot,\ustar_l)\ne f^{P'}_l(\cdot,\ustar_l)$ for some $l$: differences at other context values are neither observable nor testable.
\end{theorem}
\begin{theorem}[Evidence can be arbitrarily cheaper than intervention]
\label{thm:cheaper}
For every $r\ge1$ there are an SCM on $r+1$ variables, a history $\xobs$, and history-equivalent $P,P'$ with identical structural functions such that (i) $\sep(P,P')=r$ under unit weights and $\Qfull$; (ii) $\sep(P,P')=\infty$ for any $\cQ$ whose interventions have domain size at most $r-1$; (iii) $u_P,u_{P'}$ differ in one coordinate $l_0$, so the single verified fact $E=\{l_0\}$ eliminates $P'$ at zero cost.
\end{theorem}
Roots $X_1,\dots,X_r$ equal their exogenous variables, $X_{r+1}=(X_1\vee\dots\vee X_r)\vee U_{r+1}$, the history is all ones, and the two hypotheses differ only in $U_{r+1}$, which the observed causes mask until all $r$ of them are clamped to $0$.

\section{Details of the Continuous MNIST Study}
\label{app:cont}

This appendix completes Section~\ref{sec:experiments}. The code is \texttt{code/mnist/cont.py} (networks), \texttt{audit\_cont.py} (the auditor) and \texttt{floor\_cnn.py cont} (the plain CNNs). Every number is generated from the result files into \texttt{macros/} by \texttt{analyze\_cont.py}.

\subsection{Networks}
\label{app:cont-nets}

\textbf{Rendering.} A digit of hue $c\in[0,1]$ puts its gray levels, scaled by $c$, in the red channel and, scaled by $1-c$, in the green channel of a black RGB image; $c=1$ is pure red, $c=0$ pure green, and $c=0.5$ an olive mix (Figure~\ref{fig:cont}a). In observed images a large digit has a hue drawn uniformly from $[0.75,1]$ and a small digit from $[0,0.25]$.

\textbf{Architecture.} Each slot has the encoder of Appendix~\ref{app:mnist-setup}, whose output unit $u_s$ gives the percept $Z_s=\tanh(u_s)$. The head is a top-1 router: it outputs $\gamma Z_r+\beta$ if $Z_p>0$ and $\gamma Z_l+\beta$ otherwise, with learned $\gamma,\beta$. The routing is hard in the forward pass and uses the gradient of $\sigma(4Z_p)$ in the backward pass. In a bypass network the router reads the sign of a second encoder of the pointer image instead of $Z_p$.

\textbf{Training.} Adam with learning rate $10^{-3}$ and a cosine schedule, $4{,}000$ steps of $256$ fresh images; the label follows the shape world.
\begin{itemize}[nosep,leftmargin=*]
\item \emph{Trained} ($\alpha$): each digit's hue is drawn from its size's observed range with probability $\alpha$ and uniformly otherwise. The loss is the task loss plus a concept loss training each $u_s$ to report its digit's size.
\item \emph{Planted}: each slot $s$ gets an angle $\phi_s$ drawn uniformly from $[0,\pi/2]$, and its percept is trained by squared error towards $\tanh\big(3(\cos\phi_s(2D_s-1)+\sin\phi_s(2c_s-1))\big)$, which reads shape at $\phi_s=0$, hue at $\phi_s=\pi/2$, and a continuous mix in between. Hues are uniform, and the label is routed by the planted percepts.
\item \emph{Bypass}: value percepts as in the $\alpha=1$ networks. The pointer's percept is trained on uniform hues to report whether the pointer's hue is at least $0.5$, and the router's encoder is trained on the task with observed hues. Both read the pointer's hue and agree on every observed image, so the log fits the class exactly, yet the router never reads the percept.
\end{itemize}

\subsection{The auditor}
\label{app:cont-audit}

\textbf{Log and version space.} For each network the auditor records $3{,}000$ observed images with their percepts and outputs. It fits $\gamma$ and $\beta$ by least squares on the logit, which reproduces every logged output of every network, the bypass networks included, to within \mcAllHeadResid{} in the logit. For each slot, size and hue within the observed range, it bounds the percept by the $0.5\%$ and $99.5\%$ quantiles of the logged percepts in the hue bin of width $0.05$ that contains that hue, widened by $0.02$. The version space $\cV_0$ is the box that these bounds cut out of $[-1,1]^{72}$, with the observed digit's percept at its observed hue fixed to its observed value. A hypothesis is a point of this box. An observed image is used only if its own percepts lie within their bounds.

\textbf{When the truth is outside the class.} The bounds are estimated from other digits, so a particular digit's percept can fall outside them, for instance that of the held-out digit a swap puts in, which the log never saw. Up to a floating-point error of $10^{-5}$, the true percepts lie in the box at \mcInclassInBox{} of the \mcInclassImages{} observed images of the in-class networks. At the others the class is misspecified and Proposition~\ref{prop:continuous} does not apply; \mcOutVoid{} of the certificates issued there are void.

\textbf{Queries and answers.} A visual query sets each slot's hue to one of the eleven grid hues or keeps it, and may swap the digit for one fixed held-out digit of the other size; its cost is the total change in hue plus one per swap. The white-box family may also set the pointer's percept to $-1$, $-0.5$, $0$, $0.5$ or $1$, at a cost of half the change. The answer is the network's output probability, computed exactly.

\textbf{Exact separation.} Write $\theta$ for the hypothesis. A query's selected percept is a coordinate of $\theta$ or a patched constant, chosen by the sign of the pointer's percept. An answer $y$ constrains the selected percept to the interval that $y$ allows, with $\eta=10^{-4}$ in probability. When the pointer's sign is already determined by the current bounds, this is a bound on one coordinate. Otherwise it is a disjunction encoded with one binary variable and big-$M$ constraints. Since the network routes right exactly when $Z_p>0$, routing right requires $Z_p\ge\mu$ with $\mu=10^{-3}$, far above the solver's feasibility tolerance; the true pointer percepts of the in-class networks never fall in $(0,\mu)$. The largest and smallest output a query can have over the version space are two mixed-integer linear programs, solved exactly with HiGHS. A cheap interval bound, computed from exact per-coordinate projections, skips queries it already proves narrower than $\varepsilon=0.05$, and a query once proven narrow is never tested again, since the version space only shrinks.

\textbf{Rules and checks.} The rules are those of Appendix~\ref{app:mnist-audit}, restricted to minimum cost, random separating and random probing. A certificate is \emph{correct} when every survivor is within $\varepsilon$ of the network on every admissible query within the radius; all of these queries are run for the check. Validation draws up to $20$ unexecuted admissible queries within the radius and refutes at the first answer farther than $\varepsilon$ from some survivor.

\subsection{Results}
\label{app:cont-results}

Table~\ref{tab:cont-full} reports every network family, intervention family and radius. Wherever the truth lies in the version space, no certificate fails: \mcThmWrong{} errors in \mcThmRealRuns{} runs. Table~\ref{tab:cont-cnn} gives the plain CNNs.

\begin{table}[t]
\caption{The continuous study, all networks, families and radii (means over observed images). $|\cQ_R|$ is the number of admissible interventions; \iss{} gives the number of interventions and their total cost; the next two columns give the cost of random separating queries and of random probing; then the fraction of minimum-cost certificates that are correct and void, and the fraction of random-probing runs that refute the class.}
\label{tab:cont-full}
\centering\footnotesize
\setlength{\tabcolsep}{3.5pt}
\begin{tabular}{llccccccccc}
\toprule
& & & & \multicolumn{2}{c}{\iss{}} & random & random & & & probing\\
Networks & Family & $R$ & $|\cQ_R|$ & int. & cost & sep. & probing & correct & void & refutes\\
\midrule
Trained, $\alpha=1$ & visual & 0.5 & 210 & 7.6 & 2.47 & 3.27 & 34.4 & 100\% & 0\% & 0\%\\
 &  & 1.0 & 752 & 24.1 & 14.87 & 18.52 & 332.8 & 100\% & 0\% & 0\%\\
\addlinespace[1pt]
 & white-box & 0.5 & 486 & 7.6 & 2.47 & 3.14 & 80.3 & 100\% & 0\% & 0\%\\
 &  & 1.0 & 2224 & 24.2 & 14.92 & 17.82 & 839.8 & 100\% & 0\% & 0\%\\
\addlinespace[1pt]
Trained, $\alpha=0.9$ & visual & 0.5 & 221 & 13.4 & 4.00 & 5.34 & 43.1 & 100\% & 0\% & 0\%\\
 &  & 1.0 & 785 & 38.9 & 23.63 & 31.14 & 375.4 & 100\% & 0\% & 0\%\\
\addlinespace[1pt]
 & white-box & 0.5 & 510 & 13.4 & 4.00 & 5.50 & 82.4 & 100\% & 0\% & 0\%\\
 &  & 1.0 & 2318 & 36.9 & 21.70 & 25.57 & 1060.3 & 100\% & 0\% & 0\%\\
\addlinespace[1pt]
Trained, $\alpha=0.5$ & visual & 0.5 & 206 & 21.1 & 6.10 & 8.33 & 38.0 & 100\% & 0\% & 0\%\\
 &  & 1.0 & 742 & 58.8 & 35.16 & 46.95 & 337.3 & 100\% & 0\% & 0\%\\
\addlinespace[1pt]
 & white-box & 0.5 & 478 & 21.1 & 6.12 & 8.05 & 78.6 & 100\% & 0\% & 0\%\\
 &  & 1.0 & 2185 & 54.3 & 31.18 & 33.98 & 1058.0 & 100\% & 0\% & 0\%\\
\addlinespace[1pt]
Planted & visual & 0.5 & 210 & 13.0 & 3.68 & 4.70 & 27.1 & 98\% & 2\% & 3\%\\
 &  & 1.0 & 753 & 33.2 & 19.06 & 22.94 & 310.1 & 98\% & 2\% & 3\%\\
\addlinespace[1pt]
 & white-box & 0.5 & 487 & 13.2 & 3.76 & 5.13 & 60.8 & 98\% & 2\% & 3\%\\
 &  & 1.0 & 2280 & 29.8 & 16.16 & 21.46 & 942.9 & 98\% & 2\% & 3\%\\
\addlinespace[1pt]
Bypass & visual & 0.5 & 210 & 7.9 & 2.72 & 3.33 & 35.4 & 100\% & 0\% & 0\%\\
 &  & 1.0 & 753 & 25.2 & 15.83 & 18.74 & 287.1 & 100\% & 0\% & 0\%\\
\addlinespace[1pt]
 & white-box & 0.5 & 483 & 7.9 & 2.72 & 3.42 & 52.6 & 63\% & 37\% & 23\%\\
 &  & 1.0 & 2208 & 25.3 & 15.88 & 18.33 & 47.4 & 3\% & 97\% & 97\%\\
\addlinespace[1pt]
All in class & visual & 0.5 & 212 & 13.6 & 3.99 & 5.27 & 33.9 & 99\% & 1\% & 1\%\\
 &  & 1.0 & 757 & 37.6 & 22.35 & 28.50 & 333.1 & 99\% & 1\% & 1\%\\
\addlinespace[1pt]
 & white-box & 0.5 & 490 & 13.7 & 4.02 & 5.39 & 72.6 & 99\% & 1\% & 1\%\\
 &  & 1.0 & 2258 & 35.0 & 20.02 & 24.05 & 968.8 & 99\% & 1\% & 1\%\\
\bottomrule
\end{tabular}
\end{table}

\begin{table}[t]
\caption{Plain CNNs trained on $N$ images with continuous hue (mean of three seeds): held-out error on observed images, and error against the shape world on every single-slot edit of cost at most $1$ (a recolor to any hue of the grid, or a swap). The shape and color worlds disagree on 26.6\% of these edits.}
\label{tab:cont-cnn}
\centering\small
\begin{tabular}{rcccccc}
\toprule
& \multicolumn{2}{c}{$\alpha=1$} & \multicolumn{2}{c}{$\alpha=0.99$} & \multicolumn{2}{c}{$\alpha=0.9$}\\
\cmidrule(lr){2-3}\cmidrule(lr){4-5}\cmidrule(lr){6-7}
$N$ & held-out & edits & held-out & edits & held-out & edits\\
\midrule
100 & 5.0\% & 27.4\% & 6.3\% & 27.8\% & 14.3\% & 28.3\%\\
300 & 0.0\% & 26.4\% & 0.3\% & 26.3\% & 3.9\% & 25.5\%\\
1{,}000 & 0.0\% & 26.4\% & 0.1\% & 26.1\% & 1.4\% & 22.6\%\\
3{,}000 & 0.0\% & 26.3\% & 0.0\% & 25.5\% & 1.1\% & 16.1\%\\
10{,}000 & 0.0\% & 26.4\% & 0.1\% & 23.7\% & 0.9\% & 8.7\%\\
30{,}000 & 0.0\% & 26.3\% & 0.1\% & 18.2\% & 0.6\% & 4.5\%\\
60{,}000 & 0.0\% & 26.5\% & 0.1\% & 13.3\% & 0.4\% & 3.2\%\\
\bottomrule
\end{tabular}
\end{table}

\textbf{Compute.} On one CPU core a minimum-cost run takes \mcTimeIss{}\,s on average (at most \mcTimeIssMax{}\,s) and solves \mcMilpsIss{} MILPs. Random probing executes far more queries and piles up disjunctions: it averages \mcTimeRp{}\,s and takes up to \mcTimeRpMax{}\,s. Networks are audited in parallel, one process each; the whole continuous study, including training, takes under an hour on one RTX 4080 Super GPU and $15$ CPU cores.

\clearpage

\section{The Boolean MNIST Study}
\label{app:mnist}

This appendix reports the Boolean special case of Section~\ref{sec:experiments}. The ink is pure red or pure green, so each digit's color is a Boolean variable $C_s$ (the ink is red), and each percept is a single binary unit. The version space is then finite, every certificate is exact with no tolerance, and every claim of Theorems~\ref{thm:sound} and~\ref{thm:cost} can be checked on every run. The code is in \texttt{code/mnist}, and every number is generated from its result files into \texttt{macros/} by \texttt{make mnist-tables}.

\subsection{Setting and results}
\label{app:mnist-main}

\begin{figure}[t]
\centering
\includegraphics[width=\linewidth]{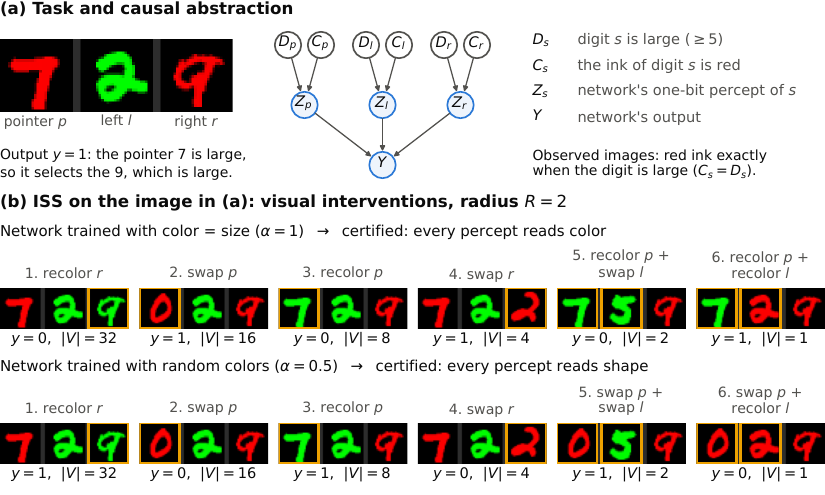}
\caption{\iss{} on MNIST. (a) An observed image and the high-level causal model; image edits set $D_s$ and $C_s$, and the percepts $Z_s$ are the network's one-bit units. (b) Two networks whose logs fit the same cells, audited on this image: the intervention \iss{} chose at each step (edited digits framed), the network's answer $y$ and the number $|V|$ of surviving candidates. Both certificates hold on all $21$ interventions of cost at most $2$, checked on the networks.}
\label{fig:mnist}
\end{figure}

\textbf{Task.}
Each image shows three handwritten MNIST digits \citep{lecun1998gradient} side by side, each drawn in red or green ink (Figure~\ref{fig:mnist}a): a pointer $p$, a left digit $l$ and a right digit $r$. Call a digit \emph{large} if it is at least $5$. A small pointer selects the left digit and a large pointer the right one, and the label says whether the selected digit is large. In Figure~\ref{fig:mnist}a the pointer $7$ is large, so it selects the $9$, which is large: the label is $1$. The task is a Boolean version of MNIST pointer-value retrieval \citep{zhang2021pointer}, a benchmark for causal abstraction \citep{geiger2022inducing}.

\textbf{Causal abstraction.}
A causal abstraction explains a network by a small causal model whose variables are aligned with parts of the network \citep{geiger2021causal}. Ours is drawn in Figure~\ref{fig:mnist}a. For each digit $s\in\{p,l,r\}$ it has two input variables, $D_s$ (the digit is large) and $C_s$ (its ink is red), and a percept $Z_s=g_s(D_s,C_s)$, which is what the network believes about the digit. The output is $Y=h(Z_p,Z_l,Z_r)$. The audited networks make the alignment exact. Each digit is read by its own small CNN that ends in a single binary unit, which plays the role of $Z_s$, and an MLP computes the output from the three bits: a concept-bottleneck network \citep{koh2020concept}. What remains open is the mechanisms. The intended abstraction reads shape, $g_s(D,C)=D$, and routes with the pointer ($h$ is a multiplexer). But a percept could as well read the ink, $g_s(D,C)=C$, or need both (\emph{and}) or either (\emph{or}).

\textbf{Why observation cannot decide.}
The auditor observes the network only on images in which the ink is red exactly when the digit is large, the shortcut of Colored MNIST \citep{arjovsky2019invariant}. On such images shape and ink always agree, so a percept that reads shape and one that reads ink produce the same bits and the same outputs. A log of $5{,}000$ observed images, with the network's bits and outputs, fixes $h$ and the two cells of each $g_s$ that occur: small green digits and large red ones. The other two cells, a small red digit and a large green one, never occur, however large the log. They are the permanently unobserved cells of Proposition~\ref{prop:vspace}, $m_\infty=6$ in all, so the version space $\cV_0$ holds $2^6=64$ candidate abstractions: one of the four tables for each digit (Table~\ref{tab:mnist-tables}).

\textbf{Interventions and certificates.}
A \emph{visual} intervention edits the image: it swaps a digit for a held-out digit of the other size, $\Do(D_s)$, or recolors it, $\Do(C_s)$. An \emph{interchange} intervention \citep{geiger2021causal} acts inside the network: it overwrites a percept bit with the value it takes on another input, $\Do(Z_s)$. The visual family allows image edits only; the white-box family allows both kinds. Each changed variable costs $1$, so the radius $R$ is the number of variables an intervention may change. For example, $R=2$ covers recoloring the pointer while swapping the left digit. The answer is the network's output, by majority over five renderings of each swap with different held-out digits. \iss{} stops at radius $R$ once all surviving candidates predict the same output for every intervention of cost at most $R$. By Theorem~\ref{thm:sound}, the survivors then predict the network's output on all of these interventions, provided its abstraction is among the $64$ candidates. The binary bottleneck lets us check this claim directly, by running every admissible intervention on the network.

\textbf{Networks.}
We audit $50$ networks at $50$ observed images each:
\begin{itemize}[nosep,leftmargin=*]
\item \emph{trained} ($20$): trained on the task with ink agreeing with size with probability $\alpha\in\{1,0.99,0.9,0.5\}$, five seeds each; at $\alpha=1$ the training data contain the shortcut, and at $\alpha=0.5$ the ink is random;
\item \emph{planted} ($20$): each percept is trained to one of the four tables at random, so the true abstraction is known in advance;
\item \emph{bypass} ($10$): the pointer's bit is trained to report the pointer's size, but the output ignores that bit and reads the pointer through a second, separate encoder, so that no candidate describes the network.
\end{itemize}

\textbf{Result 1: observation cannot reveal the abstraction.}
Held-out accuracy on the observed kind of image cannot tell what a network reads. Every network in the class scores at least \mnInclassAccMin{}, and the color readers ($\alpha=1$) score highest, \mnColorAcc{}, because the ink never misleads them. Yet the intended abstraction mispredicts \mnColorVisIntendedPctTwo{} of their interventions of cost at most $2$ (Table~\ref{tab:mnist}). Theorem~\ref{thm:floor} makes this a limit of observation itself: whatever abstraction an observational auditor reports, it mispredicts at least \mnInclassVisFloorPctTwo{} of these interventions for some network consistent with its log, averaged over observed images, however large the log.

The same limit binds ordinary learners. Call the intended labeling rule, which reads digit sizes, the \emph{shape world}, and the same rule applied to ink colors the \emph{color world}. The two worlds label every observed image identically. Plain CNNs trained on up to $60{,}000$ observed images reach zero held-out error, yet their outputs on single-variable interventions disagree with the shape world \mnCnnASixtyThouShapePct{} of the time (Figure~\ref{fig:mnistplots}a): they learned the color world. When one digit in ten carries the other ink ($\alpha=0.9$), the mismatched cells do occur in the data, and the error falls with $N$ to \mnCnnCSixtyThouShapePct{}. These are the finite-sample cells of Proposition~\ref{prop:vspace}, which more data does reveal.

\begin{figure}[t]
\centering
\includegraphics[width=\linewidth]{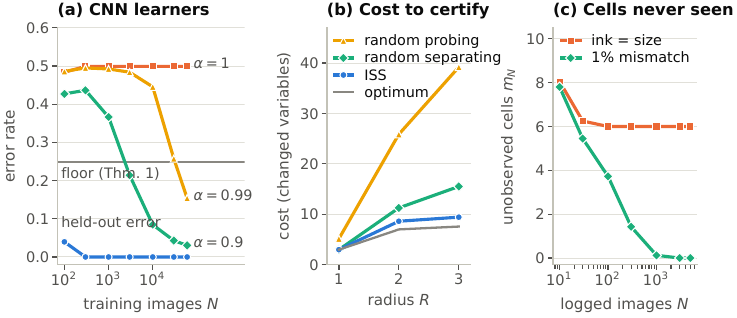}
\caption{(a) Plain CNNs trained on $N$ observed images: held-out error, and error on single-variable interventions against the shape world, for three values of $\alpha$; the gray line is the floor of Theorem~\ref{thm:floor} for the $\alpha=1$ regime; under $\alpha<1$ the missing cells occur in the data and the floor vanishes with $N$. (b) Variables changed to certify radius $R$ with visual interventions, over the $40$ networks in the class. (c) Mechanism cells never seen in the auditor's log after $N$ images.}
\label{fig:mnistplots}
\end{figure}

\textbf{Result 2: certificates are correct and cheap.}
Figure~\ref{fig:mnist}b shows \iss{} on one image for two networks whose logs fit the same cells. At each step \iss{} runs the cheapest intervention on which the surviving candidates disagree, and keeps the candidates that predicted the network's answer. The two networks answer each of the first four interventions oppositely. From then on \iss{} steers them differently: to reach the hidden left digit, it recolors the pointer of the color reader and swaps the pointer of the shape reader, because by then it knows how each network reads the pointer. After six interventions a single candidate survives for each network, and it is the right one.

The network's abstraction lies in the class at every observed image, except at \mnMixedVisNotReal{} of those of the $\alpha=0.99$ networks, whose percepts of mismatched digits vary from one handwriting to another. Wherever it lies in the class, every run of every acquisition rule ends with a certificate, and every certificate agrees with the network on every intervention within its radius: \mnThmRealRuns{} runs, without exception (Table~\ref{tab:mnist-checks}).

Certification is also cheap. At $R=2$ with visual interventions, \iss{} never needs more than \mnInclassVisIssNMaxTwo{} interventions, which is $\lceil\log_2 C\rceil$, the fewest any learner can guarantee (Theorem~\ref{thm:cost}(b)). Together they change \mnInclassVisIssCostTwo{} variables on average. A learner told the true abstraction still needs \mnInclassVisOptTwo{}, the non-adaptive greedy cover of Theorem~\ref{thm:cost}(c) needs \mnInclassVisCoverCostTwo{}, and random probing with the same stopping test needs \mnInclassVisRpCostTwo{} (Figure~\ref{fig:mnistplots}b).

\begin{table}[t]
\caption{Auditing $50$ networks at $R=2$ (visual interventions unless marked; means over $50$ observed images). \emph{In class}: images at which one of the $64$ candidates matches the network. \emph{Intended risk}: fraction of interventions of cost at most $2$ that the intended abstraction mispredicts. \emph{Cost to certify}: variables changed by \iss{}, by a learner told the truth (optimum), and by random probing. \emph{Correct}: certificates that hold on every intervention within the radius.}
\label{tab:mnist}
\centering\small
\setlength{\tabcolsep}{4pt}
\begin{tabular}{lcccccccc}
\toprule
& Held-out & & In & Intended & \multicolumn{3}{c}{Cost to certify} & \\
\cmidrule(lr){6-8}
Networks ($n$) & accuracy & Reads & class & risk & \iss{} & optimum & random & Correct\\
\midrule
Trained, $\alpha=1$ (\mnColorNets) & \mnColorAcc & \mnColorReads & \mnColorVisReal & \mnColorVisIntendedTwo & \mnColorVisIssCostTwo & \mnColorVisOptTwo & \mnColorVisRpCostTwo & \mnColorVisIssCorrectTwo\\
Trained, $\alpha=0.99$ (\mnMixedNets) & \mnMixedAcc & \mnMixedReads & \mnMixedVisReal & \mnMixedVisIntendedTwo & \mnMixedVisIssCostTwo & \mnMixedVisOptTwo & \mnMixedVisRpCostTwo & \mnMixedVisIssCorrectTwo\\
Trained, $\alpha\le0.9$ (\mnShapeNets) & \mnShapeAcc & \mnShapeReads & \mnShapeVisReal & \mnShapeVisIntendedTwo & \mnShapeVisIssCostTwo & \mnShapeVisOptTwo & \mnShapeVisRpCostTwo & \mnShapeVisIssCorrectTwo\\
Planted (\mnPlantedNets) & \mnPlantedAcc & \mnPlantedReads & \mnPlantedVisReal & \mnPlantedVisIntendedTwo & \mnPlantedVisIssCostTwo & \mnPlantedVisOptTwo & \mnPlantedVisRpCostTwo & \mnPlantedVisIssCorrectTwo\\
\addlinespace[2pt]
Bypass (\mnBypassNets) & \mnBypassAcc & \mnBypassReads & \mnBypassVisReal & \mnBypassVisIntendedTwo & \mnBypassVisIssCostTwo & \mnBypassVisOptTwo & \mnBypassVisRpCostTwo & \mnBypassVisIssCorrectTwo\\
\quad white-box & & & \mnBypassWbReal & \mnBypassWbIntendedTwo & \mnBypassWbIssCostTwo & -- & \mnBypassWbRpCostTwo & \mnBypassWbIssCorrectTwo\\
\bottomrule
\end{tabular}
\end{table}

\textbf{Result 3: a wrong class is caught only by validation.}
The bypass networks separate behavior from abstraction. With visual interventions, \iss{} certifies that the pointer is read by color. As a claim about behavior this is correct: the network answers every image edit within the radius as predicted. As an abstraction it is wrong. The pointer's bit reads shape (it reports the size of \mnBypassPointerShape{} of mismatched digits correctly), so a probe trained on that bit would endorse the intended abstraction \citep{elazar2021amnesic}; but the output never uses the bit.

Interchange interventions expose the difference: flipping the pointer's bit changes nothing, while every candidate predicts that it reroutes the output. No candidate matches these networks, yet \iss{} still halts with agreement in every white-box run at $R=2$, and every such certificate is void. The reason is structural. \iss{} only asks about interventions on which the candidates disagree, and they all agree about the pointer's bit, so it never asks the question that would expose the bypass (Figure~\ref{fig:mnist-bypass}). A validation phase of twenty random interventions refutes \mnBypassWbIssValRefTwo{} of these certificates, and random probing, which does not look for disagreement, refutes the class during learning in \mnBypassWbRpRefutedTwo{} of runs (Table~\ref{tab:mnist-bypass}). The $\alpha=0.99$ networks are a milder case: \mnMixedVisIssVoidTwo{} of their certificates at $R=2$ are void, and validation refutes \mnMixedVisIssValRefTwo{} of those.

\subsection{Task, data and networks}
\label{app:mnist-setup}

\textbf{Data.} Networks are trained on the $60{,}000$ MNIST training digits. The $10{,}000$ test digits are split at random into two disjoint pools of $5{,}000$: the \emph{observational pool} supplies the auditor's log and the observed images, and the \emph{exemplar pool} supplies the digits that realize interventions. A digit is drawn in red or green by placing its gray levels in the red or the green channel of a black RGB image. An image has three slots, pointer $p$, left $l$ and right $r$.

\textbf{Architecture.} Each slot has its own encoder: two $5\times5$ convolutions with $16$ and $32$ channels, each followed by ReLU and $2\times2$ max pooling, a $64$-unit ReLU layer and a linear unit whose sign is the percept bit $Z_s$. The forward pass uses the hard bit, and the backward pass uses the gradient of the sigmoid, as in the straight-through estimator \citep{bengio2013estimating}. The head is an MLP with two hidden layers of $64$ ReLU units that reads the three bits as $\pm1$. In a bypass network the pointer has a second encoder of the same shape, whose $64$ features pass through a $4$-unit $\tanh$ layer. The head reads these four numbers and the two value bits, and never the pointer bit.

\textbf{Training.} All networks use Adam with learning rate $10^{-3}$ and a cosine schedule, for $4{,}000$ steps of $256$ freshly drawn images. The label follows the shape world, $Y=\mathrm{MUX}(D_p;D_l,D_r)$, except for planted networks.
\begin{itemize}[nosep,leftmargin=*]
\item \emph{Trained} networks ($\alpha$): ink agrees with size independently in each slot with probability $\alpha$. The loss is the task loss plus a concept loss that trains each bit to report its digit's size, as in joint concept-bottleneck training \citep{koh2020concept}. With $\alpha=1$ size and ink coincide on every training image, so this loss cannot favor shape over ink.
\item \emph{Planted} networks: each slot is assigned one of the four tables of Table~\ref{tab:mnist-tables} at random ($20$ distinct assignments), the images have random ink, each bit is trained to its planted table and the label is the multiplexer of the planted percepts.
\item \emph{Bypass} networks: the value bits are trained as in the $\alpha=1$ networks. The pointer bit is trained to report the pointer's size on a separate batch with random ink, so it reads shape. The head is trained on the task with ink matching size.
\end{itemize}

\begin{table}[t]
\caption{The four completions of a percept's table. The log fixes the two diagonal cells; the two off-diagonal cells are never observed.}
\label{tab:mnist-tables}
\centering\small
\begin{tabular}{lcccc}
\toprule
& small, green & small, red & large, green & large, red\\
Percept reads & $(D,C)=(0,0)$ & $(0,1)$ & $(1,0)$ & $(1,1)$\\
\midrule
shape & 0 & 0 & 1 & 1\\
color & 0 & 1 & 0 & 1\\
and (both) & 0 & 0 & 0 & 1\\
or (either) & 0 & 1 & 1 & 1\\
\bottomrule
\end{tabular}
\end{table}

\begin{table}[t]
\caption{The audited networks. Held-out accuracy is on the observed regime (ink matches size) against the shape world's label, as a mean with the minimum in parentheses; percept accuracy is the agreement of the bits with the digits' sizes there; $\hat p$ is the fraction of logged percepts that disagree with the fitted cells. \emph{Reads} summarises the truth's tables over all observed images, and the last two columns give the fraction of observed images at which some member of $\cV_0$ matches the network on every query of the visual and of the white-box family.}
\label{tab:mnist-nets}
\centering\footnotesize
\setlength{\tabcolsep}{3.5pt}
\begin{tabular}{lccccccc}
\toprule
& & Held-out & Percept & & & \multicolumn{2}{c}{Truth in $\cV_0$}\\
Networks & $n$ & accuracy & accuracy & $\hat p$ & Reads & visual & white-box\\
\midrule
Trained, $\alpha=1$ & 5 & 100.0\% (100.0\%) & 100.0\% & 0.00\% & color & 100\% & 100\%\\
Trained, $\alpha=0.99$ & 5 & 99.7\% (99.7\%) & 99.8\% & 0.18\% & shape (84\%) & 92\% & 92\%\\
Trained, $\alpha\le0.9$ & 10 & 98.6\% (97.6\%) & 99.1\% & 0.82\% & shape (98\%) & 100\% & 100\%\\
Planted & 20 & 99.0\% (98.0\%) & 99.3\% & 0.65\% & as planted (98\%) & 100\% & 100\%\\
Bypass & 10 & 100.0\% (100.0\%) & 99.6\% & 0.38\% & color & 100\% & 0\%\\
\bottomrule
\end{tabular}
\end{table}

\subsection{The audit}
\label{app:mnist-audit}

\textbf{Log and version space.} For each network the auditor draws $5{,}000$ images from the observational pool with ink matching size, records the concepts, the three bits and the output, and fits every cell that occurs by majority (Proposition~\ref{prop:vspace} with $p>0$). The fitted head is the multiplexer for every network. The fraction of logged percepts that disagree with the fit, the perception noise $\hat p$, averages \mnInclassMis{}. The soft certificate of Theorem~\ref{thm:soft} costs nothing here. At $\delta=0.01$ and $\hat p$, the probability that any fitted cell of a variable is wrong is below $\delta/n'$ already at a budget of $T_j=0$, so no fitted cell may be flipped and $\cV_0^{\delta}=\cV_0$. The six off-diagonal cells are never realized, and $\cV_0$ has $64$ members for every network. Figure~\ref{fig:mnistplots}(c) shows the unobserved cells as $N$ grows. When $1\%$ of the ink is mismatched instead, every cell is realized within $3{,}000$ images in every log.

\textbf{Observed images.} For each network we draw observed images from the observational pool, with ink matching size, and keep the first $50$ whose percepts agree with the fitted cells, so that the context of the observed state has no perception noise. About \mnInclassSkipped{} of the drawn images are skipped.

\textbf{Queries and answers.} A query sets a subset of the roots to values other than the observed ones and, in the white-box family, sets bits: a bit may be flipped (cost $1$) or clamped at its observed value (cost $0$, allowed only in a slot whose digit is edited, since elsewhere it changes nothing). The visual family has $6$, $15$ and $20$ queries of cost $1$, $2$ and $3$; the white-box family has $15$, $87$ and $248$. For each observed image we draw a bank of five exemplar digits of each size for each slot. Rendering $i$ of a query replaces every swapped digit by the $i$-th exemplar of the required size and ink, sets the patched bits, and runs the network. The answer is the majority of the five outputs. The answer to every query is thus a fixed function of the query, and the truth is the member of $\cV_0$ that matches the network on every query of the family, when there is one.

\textbf{Acquisition rules.} All rules share Algorithm~\ref{alg:iss}'s elimination and stopping test and differ only in the query they pick among the separating queries within the radius:
\begin{itemize}[nosep,leftmargin=*]
\item \emph{minimum cost} (the default): a cheapest separating query; ties go to the most balanced split of the survivors, then at random;
\item \emph{information per cost}: the query maximizing the entropy of the survivors' split per unit cost, the acquisition principle of Bayesian experimental design \citep{chaloner1995bayesian};
\item \emph{random separating}: a separating query uniformly at random;
\item \emph{random probing}: an admissible query within the radius uniformly at random, separating or not, without replacement, until the stopping test holds.
\end{itemize}

\textbf{Checks.} A certificate is \emph{correct} when every survivor answers every admissible query within the radius as the network does; all of these queries are run for the check, not only those the learner executed. The optimum of a learner told the truth is the cheapest set of queries that separates the truth from every hypothesis within separation cost $R$ of it, a weighted set cover solved exactly as a 0-1 integer program with HiGHS; any correct learner pays at least this much (proof of Theorem~\ref{thm:cost}(b)). The greedy cover of Theorem~\ref{thm:cost}(c) and its optimum are computed in the same way. Validation draws up to $20$ admissible queries within the radius that the run did not execute and refutes the class at the first answer that contradicts the survivors.

\subsection{Additional results}
\label{app:mnist-results}

\textbf{Rules and radii.} Table~\ref{tab:mnist-full} reports every rule at every radius. The two rules that look for disagreement, minimum cost and information per cost, pay the same. Random separating queries cost more, and random probing far more, most of all in the white-box family, where most admissible queries separate nothing. Correctness does not depend on the rule: wherever the truth lies in the class, every rule's certificate is correct (Table~\ref{tab:mnist-checks}).

\textbf{Planted abstractions.} The certificates recover the planted tables at \mnPlantedMatch{} of observed images, and at \mnPlantedMatchSlots{} of slots. At the other images the network departs from its planted table on some exemplar digit of the bank, and the certificate follows the network. This is the intended behavior: \iss{} certifies the network, not its training target.

\textbf{Theorem checks.} No claim of Theorems~\ref{thm:sound} and~\ref{thm:cost} fails on any run whose truth lies in the class (Table~\ref{tab:mnist-checks}). Adaptivity usually pays: the non-adaptive greedy cover costs more than the adaptive minimum-cost run in \mnThmCoverWorseThanIss{} of \mnThmCoverWorseN{} instances, and less in \mnThmCoverBetterThanIss{}.

\textbf{Bypass networks.} Table~\ref{tab:mnist-bypass} gives the bypass results at every radius. At $R=1$ about half of the white-box certificates are still correct. These are the observed images whose two value digits have the same size, where rerouting the pointer changes nothing within one edit. Figure~\ref{fig:mnist-bypass} shows a white-box run on the image of Figure~\ref{fig:mnist}. To reach the left digit, \iss{} flips the pointer's bit instead of editing the pointer. The network ignores the flip, so \iss{} reads its unchanged answers as facts about the left digit and certifies a wrong table for it. Both validation interventions contradict every survivor.

\begin{figure}[t]
\centering
\includegraphics[width=\linewidth]{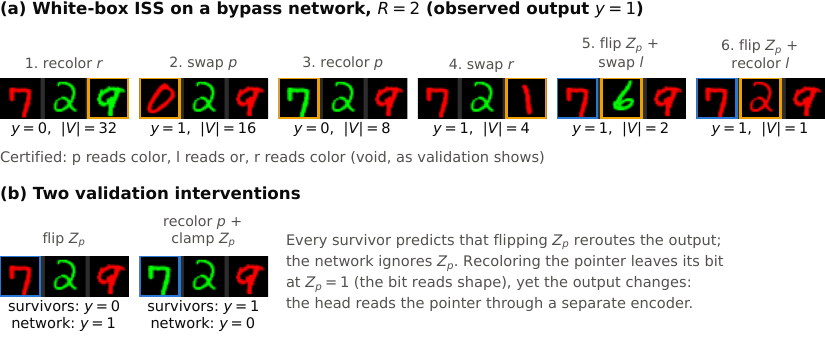}
\caption{A bypass network audited with visual and interchange interventions on the image of Figure~\ref{fig:mnist}. Yellow frames mark edited digits and blue frames patched bits. (a) Steps 1--4 are those of the color reader in Figure~\ref{fig:mnist}; at steps 5 and 6 the flipped pointer bit changes nothing, and the answers are misread as facts about the left digit. (b) Two admissible interventions within the radius on which every survivor is wrong.}
\label{fig:mnist-bypass}
\end{figure}

\textbf{A larger version space.} In a four-way pointer, two pointer digits select one of four value digits. The version space then has $2^{12}=4{,}096$ members, and a value digit is hidden unless both pointer percepts route to it, so separating queries need up to three edits. We audit \mnKtwoNets{} networks (five trained with $\alpha=1$, five with $\alpha=0.5$ and ten planted) at $25$ observed images each (Table~\ref{tab:mnist-k2}). The truth lies in the class at \mnKtwoReal{} of observed images, and no certificate fails in \mnKtwoRealruns{} runs. The gap to random probing widens with the masking: at $R=3$ it pays more than an order of magnitude more than \iss{}.

\begin{table}[t]
\caption{Certification at radius $R$, averaged over observed images. \emph{ISS} is the minimum-cost rule (number of interventions and total cost); the next three columns give the total cost of the other rules; \emph{opt.} is the optimum of a learner told the truth, $k^\star$ the truth-specific bound of Theorem~\ref{thm:cost}(b) and $\lceil\log_2 C\rceil$ the query bound (both averaged; blank where the truth is not in $\cV_0$). The last two columns give the fraction of correct certificates for the minimum-cost rule and for the worst of the four rules.}
\label{tab:mnist-full}
\centering\footnotesize
\setlength{\tabcolsep}{4pt}
\begin{tabular}{lcccccccccc}
\toprule
& & \multicolumn{2}{c}{ISS} & info & random & random & & & & correct\\
Networks & $R$ & int. & cost & /cost & sep. & probing & opt. & $k^\star$ & $\lceil\log_2 C\rceil$ & ISS / worst\\
\midrule
\multicolumn{11}{l}{\emph{Visual interventions}}\\
Trained, $\alpha=1$ & 1 & 2.8 & 2.8 & 2.8 & 2.8 & 5.0 & 2.8 & 1.0 & 2.8 & 100\%\,/\,100\%\\
 & 2 & 6.0 & 9.2 & 9.2 & 11.4 & 26.2 & 7.0 & 2.0 & 6.0 & 100\%\,/\,100\%\\
 & 3 & 6.0 & 9.2 & 9.2 & 15.5 & 35.5 & 7.0 & 2.0 & 6.0 & 100\%\,/\,100\%\\
\addlinespace[2pt]
Trained, $\alpha=0.99$ & 1 & 2.9 & 2.9 & 2.9 & 2.9 & 5.0 & 2.9 & 1.0 & 2.9 & 100\%\,/\,100\%\\
 & 2 & 6.0 & 9.0 & 9.0 & 11.3 & 26.5 & 7.1 & 2.0 & 6.0 & 99\%\,/\,99\%\\
 & 3 & 6.0 & 9.1 & 9.1 & 15.3 & 36.2 & 7.1 & 2.0 & 6.0 & 92\%\,/\,92\%\\
\addlinespace[2pt]
Trained, $\alpha\le0.9$ & 1 & 3.0 & 3.0 & 3.0 & 3.0 & 5.2 & 3.0 & 1.0 & 3.0 & 100\%\,/\,100\%\\
 & 2 & 6.0 & 9.0 & 9.0 & 11.4 & 26.4 & 7.0 & 2.0 & 6.0 & 100\%\,/\,100\%\\
 & 3 & 6.0 & 9.0 & 9.0 & 14.8 & 35.3 & 7.0 & 2.0 & 6.0 & 100\%\,/\,100\%\\
\addlinespace[2pt]
Planted & 1 & 3.0 & 3.0 & 3.0 & 3.0 & 5.1 & 3.0 & 1.0 & 3.0 & 100\%\,/\,100\%\\
 & 2 & 5.6 & 8.2 & 8.2 & 11.1 & 25.3 & 6.9 & 1.9 & 6.0 & 100\%\,/\,100\%\\
 & 3 & 6.1 & 9.8 & 9.7 & 15.9 & 42.9 & 8.1 & 2.2 & 6.0 & 100\%\,/\,100\%\\
\addlinespace[2pt]
Bypass & 1 & 3.0 & 3.0 & 3.0 & 3.0 & 5.2 & 3.0 & 1.0 & 3.0 & 100\%\,/\,100\%\\
 & 2 & 6.0 & 9.0 & 9.0 & 11.4 & 26.4 & 7.0 & 2.0 & 6.0 & 100\%\,/\,100\%\\
 & 3 & 6.0 & 9.0 & 9.0 & 15.0 & 35.1 & 7.0 & 2.0 & 6.0 & 100\%\,/\,100\%\\
\addlinespace[2pt]
All in class & 1 & 3.0 & 3.0 & 3.0 & 3.0 & 5.1 & 3.0 & 1.0 & 3.0 & 100\%\,/\,100\%\\
 & 2 & 5.8 & 8.6 & 8.6 & 11.3 & 25.9 & 7.0 & 2.0 & 6.0 & 100\%\,/\,100\%\\
 & 3 & 6.1 & 9.4 & 9.4 & 15.5 & 39.2 & 7.6 & 2.1 & 6.0 & 99\%\,/\,99\%\\
\midrule
\multicolumn{11}{l}{\emph{Visual and interchange interventions (white-box)}}\\
Trained, $\alpha=1$ & 1 & 2.8 & 2.8 & 2.8 & 2.8 & 11.3 & 2.8 & 1.0 & 2.8 & 100\%\,/\,100\%\\
 & 2 & 6.0 & 9.2 & 9.2 & 12.1 & 97.7 & 7.0 & 2.0 & 6.0 & 100\%\,/\,100\%\\
 & 3 & 6.0 & 9.2 & 9.2 & 16.9 & 107.7 & 7.0 & 2.0 & 6.0 & 100\%\,/\,100\%\\
\addlinespace[2pt]
Trained, $\alpha=0.99$ & 1 & 2.9 & 2.9 & 2.9 & 2.9 & 11.8 & 2.9 & 1.0 & 2.9 & 100\%\,/\,100\%\\
 & 2 & 6.0 & 9.1 & 9.1 & 11.9 & 91.7 & 7.2 & 2.0 & 6.0 & 95\%\,/\,95\%\\
 & 3 & 6.0 & 9.1 & 9.1 & 16.9 & 107.0 & 7.1 & 2.0 & 6.0 & 92\%\,/\,92\%\\
\addlinespace[2pt]
Trained, $\alpha\le0.9$ & 1 & 3.0 & 3.0 & 3.0 & 3.0 & 11.7 & 3.0 & 1.0 & 3.0 & 100\%\,/\,100\%\\
 & 2 & 6.0 & 9.0 & 9.0 & 12.0 & 94.0 & 7.0 & 2.0 & 6.0 & 100\%\,/\,100\%\\
 & 3 & 6.0 & 9.0 & 9.0 & 17.0 & 103.8 & 7.0 & 2.0 & 6.0 & 100\%\,/\,100\%\\
\addlinespace[2pt]
Planted & 1 & 3.0 & 3.0 & 3.0 & 3.0 & 11.8 & 3.0 & 1.0 & 3.0 & 100\%\,/\,100\%\\
 & 2 & 6.0 & 9.0 & 9.0 & 12.1 & 96.4 & 7.7 & 2.0 & 6.0 & 100\%\,/\,100\%\\
 & 3 & 6.0 & 9.0 & 9.0 & 17.1 & 114.4 & 7.5 & 2.0 & 6.0 & 100\%\,/\,100\%\\
\addlinespace[2pt]
Bypass & 1 & 3.0 & 3.0 & 3.0 & 3.0 & 8.0 & -- & -- & 3.0 & 49\%\,/\,49\%\\
 & 2 & 6.0 & 9.0 & 9.0 & 12.0 & 19.1 & -- & -- & 6.0 & 0\%\,/\,0\%\\
 & 3 & 6.0 & 9.0 & 9.0 & 17.0 & 20.2 & -- & -- & 6.0 & 0\%\,/\,0\%\\
\addlinespace[2pt]
All in class & 1 & 3.0 & 3.0 & 3.0 & 3.0 & 11.7 & 3.0 & 1.0 & 3.0 & 100\%\,/\,100\%\\
 & 2 & 6.0 & 9.0 & 9.0 & 12.0 & 95.3 & 7.4 & 2.0 & 6.0 & 99\%\,/\,99\%\\
 & 3 & 6.0 & 9.0 & 9.0 & 17.0 & 110.0 & 7.3 & 2.0 & 6.0 & 99\%\,/\,99\%\\
\bottomrule
\end{tabular}
\end{table}

\begin{table}[t]
\caption{The claims of Theorems~\ref{thm:sound} and~\ref{thm:cost} checked on every run whose truth lies in $\cV_0$ (all rules, radii, families and networks).}
\label{tab:mnist-checks}
\centering\small
\begin{tabular}{lr}
\toprule
Runs (all rules, radii, families, networks) & 60{,}000\\
Runs whose truth lies in $\cV_0$ & 53{,}496\\
\quad truth eliminated (Theorem~\ref{thm:sound}(a)) & 0\\
\quad halted without a certificate & 0\\
\quad certificate contradicted within the radius (Theorem~\ref{thm:sound}(c)) & 0\\
\quad cost above $(|\cV_0|-1)R$ (Theorem~\ref{thm:cost}(a)) & 0\\
\quad cost below the optimum of a learner told the truth & 0\\
Greedy covers within $(1+\ln M)$ of the optimal cover (Theorem~\ref{thm:cost}(c)) & 15{,}000/15{,}000\\
Greedy covers costlier than the adaptive minimum-cost run & 4{,}887/15{,}000\\
Greedy covers cheaper than the adaptive minimum-cost run & 15/15{,}000\\
\bottomrule
\end{tabular}
\end{table}

\begin{table}[t]
\caption{Plain CNNs as observational learners (mean of three seeds). \emph{Held-out} is the error on the observed regime; $r=1,2$ give the interventional risk against the shape world over all edits of cost at most $r$ of $1{,}000$ observed images. The floor of Theorem~\ref{thm:floor} is 0.25 at $r=1$ and 0.36 at $r=2$; the shape and color worlds disagree on 0.50 and 0.52 of these edits.}
\label{tab:mnist-cnn}
\centering\small
\begin{tabular}{rccccccccc}
\toprule
& \multicolumn{3}{c}{$\alpha=1$} & \multicolumn{3}{c}{$\alpha=0.99$} & \multicolumn{3}{c}{$\alpha=0.9$}\\
\cmidrule(lr){2-4}\cmidrule(lr){5-7}\cmidrule(lr){8-10}
$N$ & held-out & $r=1$ & $r=2$ & held-out & $r=1$ & $r=2$ & held-out & $r=1$ & $r=2$\\
\midrule
100 & 4.0\% & 0.48 & 0.52 & 2.1\% & 0.49 & 0.52 & 15.5\% & 0.43 & 0.47\\
300 & 0.0\% & 0.50 & 0.52 & 0.3\% & 0.50 & 0.52 & 7.6\% & 0.44 & 0.47\\
1{,}000 & 0.0\% & 0.50 & 0.52 & 0.3\% & 0.49 & 0.52 & 4.3\% & 0.37 & 0.38\\
3{,}000 & 0.0\% & 0.50 & 0.52 & 0.2\% & 0.48 & 0.51 & 2.6\% & 0.21 & 0.25\\
10{,}000 & 0.0\% & 0.50 & 0.52 & 0.2\% & 0.45 & 0.46 & 1.6\% & 0.08 & 0.11\\
30{,}000 & 0.0\% & 0.50 & 0.52 & 0.2\% & 0.26 & 0.28 & 1.1\% & 0.04 & 0.06\\
60{,}000 & 0.0\% & 0.50 & 0.52 & 0.2\% & 0.15 & 0.19 & 0.8\% & 0.03 & 0.04\\
\bottomrule
\end{tabular}
\end{table}

\begin{table}[t]
\caption{Bypass networks. \emph{Truth in $\cV_0$}: some member of the class matches the network on every query of the family. A certificate is \emph{void} when it halts with agreement but a survivor contradicts the network within the radius; \emph{refuted} runs eliminate every hypothesis during learning; \emph{validation} is the fraction of void certificates refuted by $20$ random admissible queries within the radius.}
\label{tab:mnist-bypass}
\centering\small
\begin{tabular}{lccccccccc}
\toprule
& & Truth & \multicolumn{5}{c}{minimum-cost rule} & \multicolumn{2}{c}{random separating}\\
\cmidrule(lr){4-8}\cmidrule(lr){9-10}
Family & $R$ & in $\cV_0$ & certified & correct & void & refuted & validation & void & validation\\
\midrule
visual & 1 & 100\% & 100\% & 100\% & 0\% & 0\% & -- & 0\% & --\\
 & 2 & 100\% & 100\% & 100\% & 0\% & 0\% & -- & 0\% & --\\
 & 3 & 100\% & 100\% & 100\% & 0\% & 0\% & -- & 0\% & --\\
white-box & 1 & 0\% & 100\% & 49\% & 51\% & 0\% & 100\% & 51\% & 100\%\\
 & 2 & 0\% & 100\% & 0\% & 100\% & 0\% & 93\% & 100\% & 93\%\\
 & 3 & 0\% & 100\% & 0\% & 100\% & 0\% & 96\% & 100\% & 99\%\\
\bottomrule
\end{tabular}
\end{table}

\begin{table}[t]
\caption{The four-way pointer: two pointer digits select one of four value digits, so $\cV_0$ has $2^{12}=4{,}096$ members and the visual and white-box families have $298$ and $3{,}535$ queries of cost at most $3$. 20 networks, 500 observed images; columns as in Table~\ref{tab:mnist-full}.}
\label{tab:mnist-k2}
\centering\footnotesize
\setlength{\tabcolsep}{4pt}
\begin{tabular}{llcccccccc}
\toprule
& & & \multicolumn{2}{c}{ISS} & & random & random & & \\
Networks & Family & $R$ & int. & cost & opt. & sep. & probing & $\lceil\log_2 C\rceil$ & correct\\
\midrule
Trained, $\alpha=1$ & visual & 1 & 3.9 & 3.9 & 3.9 & 3.9 & 10.3 & 3.9 & 100\%\\
 &  & 2 & 10.0 & 16.1 & 12.3 & 19.7 & 118.4 & 10.0 & 100\%\\
 &  & 3 & 12.0 & 22.1 & 18.0 & 35.5 & 549.5 & 12.0 & 100\%\\
\addlinespace[1pt]
 & white-box & 1 & 3.9 & 3.9 & 3.9 & 3.9 & 24.3 & 3.9 & 100\%\\
 &  & 2 & 10.0 & 16.1 & 12.3 & 20.1 & 511.2 & 10.0 & 100\%\\
 &  & 3 & 12.0 & 22.1 & 18.0 & 36.9 & 2864.6 & 12.0 & 100\%\\
\addlinespace[1pt]
Trained, $\alpha=0.5$ & visual & 1 & 3.9 & 3.9 & 3.9 & 3.9 & 10.0 & 3.9 & 100\%\\
 &  & 2 & 10.0 & 16.1 & 12.3 & 19.9 & 117.1 & 10.0 & 100\%\\
 &  & 3 & 12.0 & 22.1 & 18.0 & 35.6 & 552.8 & 12.0 & 100\%\\
\addlinespace[1pt]
 & white-box & 1 & 3.9 & 3.9 & 3.9 & 3.9 & 24.3 & 3.9 & 100\%\\
 &  & 2 & 10.0 & 16.1 & 12.3 & 20.3 & 507.0 & 10.0 & 100\%\\
 &  & 3 & 12.0 & 22.1 & 18.0 & 36.2 & 2939.9 & 12.0 & 100\%\\
\addlinespace[1pt]
Planted & visual & 1 & 3.9 & 3.9 & 3.9 & 3.9 & 10.1 & 3.9 & 100\%\\
 &  & 2 & 9.2 & 14.4 & 12.0 & 19.0 & 107.8 & 10.0 & 100\%\\
 &  & 3 & 11.4 & 21.2 & 17.3 & 35.1 & 485.5 & 12.0 & 100\%\\
\addlinespace[1pt]
 & white-box & 1 & 3.9 & 3.9 & 3.9 & 3.9 & 24.1 & 3.9 & 100\%\\
 &  & 2 & 10.0 & 16.1 & 13.5 & 20.6 & 529.1 & 10.0 & 100\%\\
 &  & 3 & 12.0 & 22.1 & 18.4 & 36.4 & 3426.7 & 12.0 & 100\%\\
\bottomrule
\end{tabular}
\end{table}

\textbf{Compute.} Training one network takes under a minute on one GPU. The audit takes about $5$\,s per network for the three-digit task and about $80$\,s for the four-way pointer, most of it in the exhaustive checks and the exact set-cover optima. The whole experiment, including the floor study, runs in about one and a half hours on a single consumer GPU.

\clearpage

\section{Extended Related Work}
\label{app:related}

This appendix expands the overview in Section~\ref{sec:related}.

\subsection{Structural causal models, interventions, and the causal hierarchy}
We use structural causal models \citep{pearl2009causality,peters2017elements,spirtes2000causation}: endogenous variables determined by assignments from parents and exogenous variables, interventions that replace assignments, and counterfactuals evaluated by abduction, action, and prediction. \citet{bareinboim2022hierarchy} organize these queries into the Pearl Causal Hierarchy and prove that, for almost all SCMs, lower-layer observational data underdetermine interventional and counterfactual quantities. History-equivalent but interventionally distinct perspectives are a finite, deterministic, individual-level instance of this underdetermination; we quantify \emph{how far} one must move from the observed layer to resolve it. The interventionist account of causation \citep{woodward2003making} views causal claims as claims about what happens under interventions. \citet{eberhardt2007interventions} and \citet{korb2004varieties} analyze the structural versus parametric and hard versus soft varieties on which such claims depend; our admissible family $\cQ$ makes this dependence explicit. \citet{halpern2005causes,halpern2005explanations} and \citet{halpern2016actual} define actual causes and explanations relative to an SCM and a context, and explanations relative to an \emph{epistemic state}, a set of contexts the agent considers possible; our exogenous evidence set plays this role. \citet{ibeling2023comparing} compare structural, potential-outcome, graphical, and abstraction-based frameworks; we work in the structural one. \citet{shmueli2010explain} articulates the gap between explanatory and predictive modeling that motivates our setting.

\subsection{Observational and interventional equivalence}
The classical notion is Markov equivalence: DAGs entailing the same conditional independences cannot be distinguished from observational distributions \citep{verma1990equivalence,andersson1997characterization,spirtes2000causation}. \citet{hauser2012characterization} characterize interventional Markov equivalence classes for a fixed family of hard interventions, and \citet{yang2018characterizing} extend this to general, including soft, interventions. These results concern population distributions over many contexts and answer a binary class-membership question. Our notion is individual-level (one observed history), graded (a cost threshold), and relative to an admissible family and cost of departure. Distributional equivalence is neither necessary nor sufficient for history-equivalence, but we share the guiding idea that models should be compared through the interventions that distinguish them.

\subsection{Confounding, identification, and exogenous grounding}
\label{app:confounding}

\textbf{Confounding and observational ambiguity.}
Confounding can arise when treatment and outcome share causes, creating an open back-door path such as $A\leftarrow C\rightarrow Y$; consequently, $P(Y\mid A=a)$ can differ from $P(Y\mid\Do(A=a))$ even with unlimited observational data \citep{pearl1995causal,hernan2020causal}. A variable is not a confounder merely because it is exogenous or correlated with treatment: its causal role and the adjustment set matter. This is one source of observational ambiguity, whereas the paper's history-equivalence only requires agreement at a single realized state. Such agreement can also arise from different mechanisms away from that state, even without a shared cause.

\textbf{Identification versus separation.}
Back-door adjustment identifies population effects when suitable observed covariates block the relevant paths and the required support conditions hold \citep{pearl1995causal,hernan2020causal}. More generally, \citet{shpitser2008complete} characterize identification of interventional and counterfactual distributions, including when latent variables prevent identification from lower-layer information. These results ask whether all models compatible with the available distribution and causal assumptions agree on a specified query. We start with specified candidates and search for a query witnessing disagreement at minimum departure cost. Thus, a separating intervention is a proposed discriminating experiment; it does not itself identify a population effect from observational data or establish which candidate is correct.

\textbf{A common-cause illustration.}
Consider Boolean models with $C=U_C$ and $A=C$, but with $Y=A$ in $M_1$ and $Y=C$ in $M_2$; unused exogenous inputs are ignored. At $U_C=1$, both perspectives explain $(C,A,Y)=(1,1,1)$. In $M_2$, $C$ is a common cause of $A$ and $Y$. The query $(\Do(A=0),Y)$ gives answers $0$ and $1$, respectively, and hence separation cost $1$ under unit costs when this query is admissible. Verifying $U_C=1$ leaves both perspectives grounded because their disagreement is structural. This illustrates why evidence about a common cause and an intervention on treatment supply different information.

\textbf{Scope of exogenous grounding.}
Grounding constrains a realized exogenous coordinate only when its meaning is shared across perspectives and its value is independently verified. Abducting a residual from a fitted model does not verify that value. These constraints neither establish conditional exchangeability nor guarantee that all common causes have been measured; even fully grounded candidates can disagree structurally (Theorem~\ref{thm:decomp}). The fixed-context theory specifies no probability law over $U$, so it makes no population-level independence or identification claim. A probabilistic extension could represent latent common causes through dependence among exogenous disturbances and combine grounding with identification or sensitivity analysis. The present benchmark does not evaluate robustness to latent confounding.

\subsection{Experimental design for causal discovery and model discrimination}
Two traditions ask which experiments distinguish causal hypotheses. For structure learning, \citet{eberhardt2005number} show that $\log_2 N+1$ multi-variable experiments suffice, and are worst-case necessary, to identify all causal relations among $N$ variables, while \citet{eberhardt2008almost} gives almost optimal intervention sets. Later work optimizes intervention size \citep{shanmugam2015learning}, cost \citep{kocaoglu2017cost,lindgren2018experimental}, fixed budgets \citep{ghassami2018budgeted}, or targeted subgraphs \citep{agrawal2019abcd}; \citet{hauser2014two} and \citet{squires2020active} give optimal and near-optimal active strategies, and Bayesian active structure learning dates to \citet{tong2001active} and \citet{murphy2001active}. Closest to us, \citet{choo2022verification} study \emph{verification}, the minimum interventions needed to confirm a purported DAG within its equivalence class, characterized through covered edges, with subset versions in \citet{choo2023subset}. These works identify one structure among all DAGs; we compare two \emph{given} models, possibly learned and possibly wrong, and seek the cheapest intervention on which their answers differ. In statistics and systems biology, \citet{hunter1965designs} and \citet{box1967discrimination} design sequential experiments to discriminate between rival mechanistic models, \citet{atkinson1975design} introduce $T$-optimal designs maximizing rival lack of fit, and \citet{chaloner1995bayesian} review Bayesian designs based on the expected information of \citet{lindley1956measure}. For Boolean regulatory networks, \citet{ideker2000discovery} choose perturbations discriminating among hypotheses consistent with the data, \citet{akutsu2003identification} bound the disruption and overexpression experiments needed to identify a network, and \citet{apgar2008stimulus} and \citet{kreutz2009systems} design stimuli for model selection in signaling networks. Separation cost is a worst-case, budget-indexed relative: rather than maximizing expected divergence under a prior, it finds the minimum departure from the observed state at which rivals are \emph{certain} to disagree, treating departure cost as primary. The Boolean-network methods of \citet{ideker2000discovery} and \citet{akutsu2003identification} are the closest in spirit and differ on three points: they keep the candidate networks as an explicit list and score an experiment by the candidates it eliminates or the divergence it induces, whereas our version space is implicit and separation is decided exactly over all $2^m$ completions by integer programs; their target is identification of the true network, whereas ours is a certified radius of interventional agreement that is meaningful even when many networks remain; and their guarantee is identification under the class assumption, whereas the certificate is exact, cost-bounded above and below, and explicitly conditional on realizability, with a validation phase for the misspecified case. In the terms of \citet{popper1959logic}, this is the cheapest potential falsifier of at least one perspective.

\subsection{Amortized and cost-constrained active interventional design}
\label{app:design}
The strongest recent methods for choosing interventions are amortized or explicitly cost-constrained, and it is worth saying precisely where \iss{} stands relative to them. \citet{annadani2024amortized} train CAASL, a transformer policy, by reinforcement learning on a simulator of the design environment, formalized as a hidden-parameter MDP whose hidden parameters are the graph and the mechanism parameters; the reward is the improvement in the number of correct entries of the adjacency matrix predicted by the amortized posterior of \citet{lorch2022amortized}, which the authors relate to an approximation of multi-step expected information gain, and the design horizon is a fixed number of interventions. \citet{zhang2023bayesian} assume a ladder of experimental oracles---higher fidelity is more precise and more expensive---and choose which variable to intervene on \emph{and at which fidelity} by a mutual-information criterion, with a cascading model tying the fidelities together and an $\varepsilon$-submodularity argument restoring the greedy guarantee that ordinary submodularity fails to provide here. \citet{guo2026constrained} give the most explicitly cost-bounded of the three: COPEx pre-trains an amortized posterior and design policy offline and then plans online by multi-step lookahead over a scenario tree of hypothetical outcomes, so that a design sequence can respect a budget, varying costs and design-dependent feasibility at test time. Earlier steps in the same direction include the scalable interventional design of \citet{tigas2022interventions}.

Five differences separate this line of work from ours; none of them is a claim that one target is better than the other.

\textbf{What is unknown.} All three carry uncertainty over the causal \emph{structure} and score a design by how much of it the design resolves; CAASL rewards recovery of the adjacency matrix and explicitly does not reward learning the mechanism parameters. In our setting the parent sets are given and the structure is not in doubt: what is undetermined is the mechanism on the parent configurations the observational regime never realizes, together with the exogenous context at a fixed observed state. CAASL's reward therefore has nothing to gain here: the quantity it optimizes is handed to the learner at the outset. The two targets are sequential rather than rival---structure discovery upstream, certification of the surviving mechanisms downstream---which is why Section~\ref{sec:conclusion} lists given parent sets as a limitation and not as an assumption we defend.

\textbf{What cost means.} CAASL has no cost model at all: its budget is a count of interventions, and two designs of the same count are equally affordable however far they are from the observed state. \citet{zhang2023bayesian} price the \emph{oracle} rather than the intervention, so cost indexes measurement quality, not how radical an experiment is. COPEx does price designs and plan under a budget, but cost there constrains the design sequence; it does not index a guarantee. In \iss{} the cost of an intervention is its distance from the observed state, and the same number indexes what is certified: the radius $R$ is both the budget and the scope of the guarantee (Equation~\eqref{eq:cost}, Theorem~\ref{thm:sound}).

\textbf{What is returned, and when to stop.} These methods return an estimate---a posterior, or a design sequence---after a fixed horizon, and their guarantees are about expected information or empirical recovery; none of them stops on a property of what survives. The difference is not that an information criterion \emph{cannot} express our stopping condition. Answers here are deterministic and the version space is finite, so a query's expected information gain is zero exactly when it separates nobody, and $\max_{q\in\cQ_R}\mathrm{EIG}(q)=0$ is precisely $\kappa(\cV)>R$: the exhaustion test and the information criterion agree. The difference is that the equivalence holds only when the expectation is taken over the \emph{whole} version space. Under an amortized posterior, or a sample of one, a vanishing maximum is a statement about the approximation, not about the version space. \iss{}'s test is exact by construction (Proposition~\ref{prop:implicit}) and is a statement about the surviving hypotheses alone, hence checkable without knowing the truth (Theorem~\ref{thm:sound}(d)). That is the substantive difference, and it is why the certificate survives replacing our acquisition rule with theirs.

\textbf{Expectation against worst case.} An expected-information criterion averages over a prior. Separation cost is a worst-case, budget-indexed quantity: the minimum departure from the observed state at which two rivals are \emph{certain} to disagree. The two often coincide in practice---in the Boolean MNIST study the information-per-cost rule pays almost exactly what the minimum-cost rule pays (Table~\ref{tab:mnist-full})---and differ entirely in what they license at the end of a run.

\subsection{Underspecification, multiplicity, and disagreement between predictors}
That many models fit the same data equally well is the Rashomon effect \citep{breiman2001statistical}; \citet{fisher2019all} and \citet{semenova2022existence} study Rashomon sets, while \citet{marx2020predictive}quantify ambiguity and discrepancy among near-equal-accuracy models and \citet{black2022model} survey the opportunities and concerns such multiplicity raises. \citet{damour2022underspecification} show that underspecified pipelines yield predictors with equivalent held-out performance but divergent stress-test behavior, arguing that such tests rather than held-out accuracy must decide credibility; we formalize the \emph{minimal} stress test deciding between two candidates. Shortcut learning \citep{geirhos2020shortcut,beery2018recognition} and invariance \citep{peters2016causal,arjovsky2019invariant,sagawa2020distributionally,veitch2021counterfactual} explain why observationally agreeing predictors differ causally and use environments or counterfactual edits to identify causal features. \citet{dehaan2019causal} show that imitation policies can fit demonstrations while conditioning on the wrong causes, then resolve this confusion through targeted interventions; \citet{richens2024robust} prove that agents robust to distribution shift must learn a causal world model, motivating small separation cost as fragility. Disagreement already guides query-by-committee, which shrinks a version space using queries on which hypotheses disagree \citep{mitchell1982generalization,seung1992query,freund1997selective}; Bayesian active learning scores expected disagreement \citep{houlsby2011bayesian}, ensembles use disagreement as uncertainty \citep{lakshminarayanan2017simple}, and disagreement across training runs predicts test error \citep{jiang2022assessing}. We replace the input space by interventions and the disagreement \emph{rate} by the disagreement \emph{threshold}: the cheapest intervention around the observed history on which two candidates disagree.

\subsection{Counterfactual explanations, recourse, and minimal perturbations}
Counterfactual explanations return the closest input changing a prediction \citep{wachter2017counterfactual,mothilal2020explaining}, and actionable recourse restricts changes to those an individual can make \citep{ustun2019actionable}. \citet{karimi2021algorithmic} cast recourse as a minimum-cost SCM intervention accounting for downstream effects; \citet{karimi2022survey} survey the field. Our departure cost is that of causal recourse, but the objective differs: recourse flips one model's output whereas separation finds the cheapest intervention on which two outputs differ. Adversarial examples minimally perturb an input to flip one model's decision \citep{szegedy2014intriguing,moosavi2016deepfool}; separation instead targets \emph{disagreement between models}, with $\Do$-semantics rather than pixel perturbations. Behavioral testing manually constructs near-history edits through counterfactually augmented data \citep{kaushik2020learning}, contrast sets \citep{gardner2020evaluating}, CheckList \citep{ribeiro2020beyond}, and the CEBaB benchmark of real-world concept interventions \citep{abraham2022cebab}. These are empirical low-cost interventions; our theory identifies which are guaranteed informative and how cheap the cheapest can be.

\subsection{Differential testing, distinguishing sequences, and query learning}
Differential testing searches for inputs on which implementations disagree \citep{mckeeman1998differential}; DeepXplore does so for deep networks in a whitebox manner \citep{pei2017deepxplore}, and DiffChaser searches for disagreements between network versions \citep{xie2019diffchaser}. Distinguishing and checking sequences separate finite-state machines \citep{lee1996principles}, while in query learning an equivalence query returns a counterexample separating hypothesis from target \citep{angluin1988queries}. Separation cost is the cheapest counterexample to the claim of interventional equivalence, measured by departure from an observed state rather than sequence length or input norm, with interventions on a causal model rather than inputs to a function.

\subsection{Causal abstraction and interventional interpretability}
Causal abstraction asks when one SCM faithfully coarse-grains another \citep{rubenstein2017causal,beckers2019abstracting,beckers2019approximate}. \citet{geiger2021causal} test whether a network implements a hypothesized causal model through interchange interventions on internal representations, \citet{geiger2024finding} learn alignments between causal variables and distributed representations, and \citet{geiger2023causal} give the theoretical foundations. Causal mediation \citep{vig2020investigating}, tracing and editing \citep{meng2022locating}, circuit discovery \citep{wang2023interpretability,conmy2023towards}, and neural-causal models \citep{xia2021causal} likewise probe models interventionally. These methods quantify agreement between a network and a hypothesis \emph{averaged} over interventions; we seek the \emph{minimum-cost} environment intervention distinguishing two hypotheses, a complementary criterion applicable to the abstracted models they produce. Causal representation learning and independent causal mechanisms \citep{scholkopf2021toward,parascandolo2018learning} motivate our benchmark's sparse-mechanism-difference regime, while causal data fusion \citep{bareinboim2016causal} formalizes the combination of observational and interventional evidence, which is what the \iss{} loop does adaptively and at a cost. Evaluations of causal reasoning in language models \citep{jin2023cladder,kiciman2023causal,zecevic2023causal} and causal criteria in fairness \citep{kusner2017counterfactual,zhang2018fairness} are natural application targets.

\subsection{Exogenous variables, abduction, and computation}
Counterfactual evaluation begins by abducting the exogenous context \citep{balke1994counterfactual,pearl2009causality}; probabilities of causation require bounds precisely because contexts are underdetermined \citep{tian2000probabilities}, while Gumbel-max SCMs \citep{oberst2019counterfactual} and counterfactual data augmentation \citep{lu2020sample} infer posteriors over exogenous noise from trajectories. Exogenous grounding takes the opposite stance: verified exogenous facts are constraints that eliminate hypotheses from the version space, not latent variables to marginalize. Computationally, \citet{eiter2002complexity} and \citet{aleksandrowicz2017computational} settle the complexity of structure-based actual causation, \citet{ibrahim2020checking} solve actual-causality queries in Boolean models with integer linear programs, and \citet{ibeling2020probabilistic} and \citet{mosse2024causal} show that satisfiability for causal and counterfactual languages is no harder than for their probabilistic counterparts; Boolean causal models also inherit the Boolean-network tradition \citep{kauffman1969metabolic}. Our NP-completeness result sits alongside these, and our separation checks follow the integer-programming approach of \citet{ibrahim2020checking}.

\ifISSbody\else
\bibliography{references}

@book{pearl2009causality,
  author    = {Pearl, Judea},
  title     = {Causality: Models, Reasoning, and Inference},
  edition   = {2},
  publisher = {Cambridge University Press},
  address   = {Cambridge, UK},
  year      = {2009}
}

@book{peters2017elements,
  title={Elements of causal inference: foundations and learning algorithms},
  author={Peters, Jonas and Janzing, Dominik and Sch{\"o}lkopf, Bernhard},
  year={2017},
  publisher={The MIT press}
}

@incollection{bareinboim2022hierarchy,
  title={On {P}earl's hierarchy and the foundations of causal inference},
  author={Bareinboim, Elias and Correa, Juan D and Ibeling, Duligur and Icard, Thomas},
  booktitle={Probabilistic and Causal Inference: The Works of Judea Pearl},
  publisher={Association for Computing Machinery},
  pages={507--556},
  year={2022}
}

@article{halpern2005causes,
  title={Causes and explanations: A structural-model approach. Part I: Causes},
  author={Halpern, Joseph Y and Pearl, Judea},
  journal={The British journal for the philosophy of science},
  year={2005},
  publisher={The University of Chicago Press}
}

@article{halpern2005explanations,
  title={Causes and explanations: A structural-model approach. Part II: Explanations},
  author={Halpern, Joseph Y and Pearl, Judea},
  journal={The British journal for the philosophy of science},
  year={2005},
  publisher={The University of Chicago Press}
}

@book{halpern2016actual,
  title={Actual causality},
  author={Halpern, Joseph Y},
  volume={3},
  year={2016},
  publisher={MiT Press Cambridge}
}

@inproceedings{verma1990equivalence,
  title={Equivalence and synthesis of causal models},
  author={Verma, Thomas S and Pearl, Judea},
  booktitle={Proceedings of the 6th Conference on Uncertainty in Artificial Intelligence},
  year={1990}
}

@article{andersson1997characterization,
  title={A characterization of {Markov} equivalence classes for acyclic digraphs},
  author={Andersson, Steen A and Madigan, David and Perlman, Michael D},
  journal={The Annals of Statistics},
  volume={25},
  number={2},
  pages={505--541},
  year={1997},
  publisher={Institute of Mathematical Statistics}
}

@article{hauser2012characterization,
  title={Characterization and greedy learning of interventional {Markov} equivalence classes of directed acyclic graphs},
  author={Hauser, Alain and B{\"u}hlmann, Peter},
  journal={The Journal of Machine Learning Research},
  volume={13},
  number={1},
  pages={2409--2464},
  year={2012},
  publisher={JMLR. org}
}

@inproceedings{yang2018characterizing,
  title={Characterizing and learning equivalence classes of causal dags under interventions},
  author={Yang, Karren and Katcoff, Abigail and Uhler, Caroline},
  booktitle={International Conference on Machine Learning},
  pages={5541--5550},
  year={2018},
  organization={PMLR}
}

@book{spirtes2000causation,
   author    = {Spirtes, Peter and Glymour, Clark and Scheines, Richard},
  title     = {Causation, Prediction, and Search},
  edition   = {2},
  publisher = {MIT Press},
  address   = {Cambridge, MA},
  year      = {2000}
}

@article{shmueli2010explain,
  title={To explain or to predict?},
  author={Shmueli, Galit},
  journal={Statistical science},
  pages={289--310},
  year={2010},
  publisher={JSTOR}
}

@inproceedings{rubenstein2017causal,
  title={Causal consistency of structural equation models},
  author={Rubenstein, Paul K and Weichwald, Sebastian and Bongers, Stephan and Mooij, Joris M and Janzing, Dominik and Grosse-Wentrup, Moritz and Sch{\"o}lkopf, Bernhard},
  booktitle={Proceedings of the 33rd Conference on Uncertainty in Artificial Intelligence},
  year={2017}
}

@inproceedings{beckers2019abstracting,
  title={Abstracting causal models},
  author={Beckers, Sander and Halpern, Joseph Y},
  booktitle={Proceedings of the aaai conference on artificial intelligence},
  volume={33},
  pages={2678--2685},
  year={2019}
}

@inproceedings{beckers2019approximate,
  title={Approximate causal abstractions},
  author={Beckers, Sander and Eberhardt, Frederick and Halpern, Joseph Y},
  booktitle={Uncertainty in artificial intelligence},
  pages={606--615},
  year={2020},
  organization={PMLR}
}

@article{tian2000probabilities,
  title={Probabilities of causation: Bounds and identification},
  author={Tian, Jin and Pearl, Judea},
  journal={Annals of Mathematics and Artificial Intelligence},
  volume={28},
  number={1},
  pages={287--313},
  year={2000},
  publisher={Springer}
}

@inproceedings{ibeling2023comparing,
  title={Comparing causal frameworks: Potential outcomes, structural models, graphs, and abstractions},
  author={Ibeling, Duligur and Icard, Thomas},
  booktitle={Advances in Neural Information Processing Systems},
  volume={36},
  pages={80130--80141},
  year={2023}
}

@inproceedings{balke1994counterfactual,
  title={Counterfactual probabilities: Computational methods, bounds and applications},
  author={Balke, Alexander and Pearl, Judea},
  booktitle={Uncertainty in artificial intelligence},
  pages={46--54},
  year={1994},
  organization={Elsevier}
}

@book{woodward2003making,
  author    = {Woodward, James},
  title     = {Making Things Happen: A Theory of Causal Explanation},
  publisher = {Oxford University Press},
  address   = {Oxford},
  year      = {2003}
}

@article{eberhardt2007interventions,
  title={Interventions and causal inference},
  author={Eberhardt, Frederick and Scheines, Richard},
  journal={Philosophy of science},
  volume={74},
  number={5},
  pages={981--995},
  year={2007},
  publisher={Cambridge University Press}
}

@inproceedings{eberhardt2005number,
  title={On the number of experiments sufficient and in the worst case necessary to identify all causal relations among {$N$} variables},
  author={Eberhardt, Frederick and Glymour, Clark and Scheines, Richard},
  booktitle={Proceedings of the 21st Conference on Uncertainty in Artificial Intelligence},
  pages={178--184},
  year={2005}
}

@inproceedings{eberhardt2008almost,
  title={Almost optimal intervention sets for causal discovery},
  author={Eberhardt, Frederick},
  booktitle={Proceedings of the 24th Conference on Uncertainty in Artificial Intelligence},
  pages={161--168},
  year={2008}
}

@article{hauser2014two,
  title={Two optimal strategies for active learning of causal models from interventional data},
  author={Hauser, Alain and B{\"u}hlmann, Peter},
  journal={International Journal of Approximate Reasoning},
  volume={55},
  number={4},
  pages={926--939},
  year={2014},
  publisher={Elsevier}
}

@inproceedings{shanmugam2015learning,
  title={Learning causal graphs with small interventions},
  author={Shanmugam, Karthikeyan and Kocaoglu, Murat and Dimakis, Alexandros G and Vishwanath, Sriram},
  booktitle={Advances in Neural Information Processing Systems},
  volume={28},
  year={2015}
}

@inproceedings{kocaoglu2017cost,
  title={Cost-optimal learning of causal graphs},
  author={Kocaoglu, Murat and Dimakis, Alex and Vishwanath, Sriram},
  booktitle={International Conference on Machine Learning},
  pages={1875--1884},
  year={2017},
  organization={PMLR}
}

@inproceedings{lindgren2018experimental,
  title={Experimental design for cost-aware learning of causal graphs},
  author={Lindgren, Erik and Kocaoglu, Murat and Dimakis, Alexandros G and Vishwanath, Sriram},
  booktitle={Advances in Neural Information Processing Systems},
  volume={31},
  year={2018}
}

@inproceedings{ghassami2018budgeted,
  title={Budgeted experiment design for causal structure learning},
  author={Ghassami, AmirEmad and Salehkaleybar, Saber and Kiyavash, Negar and Bareinboim, Elias},
  booktitle={International Conference on Machine Learning},
  pages={1724--1733},
  year={2018},
  organization={PMLR}
}

@inproceedings{agrawal2019abcd,
  title={Abcd-strategy: Budgeted experimental design for targeted causal structure discovery},
  author={Agrawal, Raj and Squires, Chandler and Yang, Karren and Shanmugam, Karthikeyan and Uhler, Caroline},
  booktitle={The 22nd International Conference on Artificial Intelligence and Statistics},
  pages={3400--3409},
  year={2019},
  organization={PMLR}
}

@inproceedings{tong2001active,
  title={Active learning for structure in {Bayesian} networks},
  author={Tong, Simon and Koller, Daphne},
  booktitle={Proceedings of the 17th International Joint Conference on Artificial Intelligence},
  year={2001},
  publisher={Morgan Kaufmann}
}

@techreport{murphy2001active,
  title={Active learning of causal {B}ayes net structure},
  author={Murphy, Kevin P},
  institution={University of California, Berkeley},
  year={2001}
}

@inproceedings{squires2020active,
  title={Active structure learning of causal {DAGs} via directed clique trees},
  author={Squires, Chandler and Magliacane, Sara and Greenewald, Kristjan and Katz, Dmitriy and Kocaoglu, Murat and Shanmugam, Karthikeyan},
  booktitle={Advances in Neural Information Processing Systems},
  volume={33},
  pages={21500--21511},
  year={2020}
}

@inproceedings{choo2022verification,
  title={Verification and search algorithms for causal {DAGs}},
  author={Choo, Davin and Shiragur, Kirankumar and Bhattacharyya, Arnab},
  booktitle={Advances in Neural Information Processing Systems},
  volume={35},
  pages={12787--12799},
  year={2022}
}

@inproceedings{choo2023subset,
  title={Subset verification and search algorithms for causal {DAGs}},
  author={Choo, Davin and Shiragur, Kirankumar},
  booktitle={International Conference on Artificial Intelligence and Statistics},
  pages={4409--4442},
  year={2023},
  organization={PMLR}
}

@article{box1967discrimination,
  title={Discrimination among mechanistic models},
  author={Box, George EP and Hill, WILLIAM J},
  journal={Technometrics},
  volume={9},
  number={1},
  pages={57--71},
  year={1967},
  publisher={Taylor \& Francis}
}

@article{atkinson1975design,
  author  = {Atkinson, Anthony C. and Fedorov, Valerii V.},
  title   = {The design of experiments for discriminating between two rival models},
  journal = {Biometrika},
  volume  = {62},
  number  = {1},
  pages   = {57--70},
  year    = {1975},
  doi     = {10.1093/biomet/62.1.57}
}

@article{chaloner1995bayesian,
  author = {Chaloner, Kathryn and Verdinelli, Isabella},
  title = {{Bayesian} Experimental Design: A Review},
  journal = {Statistical Science},
  volume = {10},
  number = {3},
  pages = {273--304},
  year = {1995},
  doi = {10.1214/SS/1177009939},
  publisher = {Institute of Mathematical Statistics},
}

@article{lindley1956measure,
  title={On a measure of the information provided by an experiment},
  author={Lindley, Dennis V},
  journal={The Annals of Mathematical Statistics},
  volume={27},
  number={4},
  pages={986--1005},
  year={1956},
  publisher={Institute of Mathematical Statistics}
}

@inproceedings{ideker2000discovery,
  author    = {Ideker, Trey E. and Thorsson, Vesteinn and Karp, Richard M.},
  title     = {Discovery of regulatory interactions through perturbation: Inference and experimental design},
  booktitle = {Pacific Symposium on Biocomputing 2000},
  pages     = {305--316},
  publisher = {World Scientific},
  year      = {2000}
}

@article{akutsu2003identification,
  title={Identification of genetic networks by strategic gene disruptions and gene overexpressions under a boolean model},
  author={Akutsu, Tatsuya and Kuhara, Satoru and Maruyama, Osamu and Miyano, Satoru},
  journal={Theoretical Computer Science},
  volume={298},
  number={1},
  pages={235--251},
  year={2003},
  publisher={Elsevier}
}

@article{apgar2008stimulus,
  title={Stimulus design for model selection and validation in cell signaling},
  author={Apgar, Joshua F and Toettcher, Jared E and Endy, Drew and White, Forest M and Tidor, Bruce},
  journal={PLoS computational biology},
  volume={4},
  number={2},
  pages={e30},
  year={2008},
  publisher={Public Library of Science San Francisco, USA}
}

@article{hunter1965designs,
  title={Designs for discriminating between two rival models},
  author={Hunter, William G and Reiner, Albey M},
  journal={Technometrics},
  volume={7},
  number={3},
  pages={307--323},
  year={1965},
  publisher={Taylor \& Francis}
}

@article{damour2022underspecification,
  title={Underspecification presents challenges for credibility in modern machine learning},
  author={D'Amour, Alexander and Heller, Katherine and Moldovan, Dan and Adlam, Ben and Alipanahi, Babak and Beutel, Alex and Chen, Christina and Deaton, Jonathan and Eisenstein, Jacob and Hoffman, Matthew D and others},
  journal={Journal of Machine Learning Research},
  volume={23},
  number={226},
  pages={1--61},
  year={2022}
}

@article{breiman2001statistical,
  title={Statistical modeling: The two cultures (with comments and a rejoinder by the author)},
  author={Breiman, Leo},
  journal={Statistical Science},
  volume={16},
  number={3},
  pages={199--231},
  year={2001},
  publisher={Institute of Mathematical Statistics}
}

@inproceedings{marx2020predictive,
  title={Predictive multiplicity in classification},
  author={Marx, Charles and Calmon, Flavio and Ustun, Berk},
  booktitle={International conference on machine learning},
  pages={6765--6774},
  year={2020},
  organization={PMLR}
}

@inproceedings{black2022model,
  title={Model multiplicity: Opportunities, concerns, and solutions},
  author={Black, Emily and Raghavan, Manish and Barocas, Solon},
  booktitle={Proceedings of the 2022 ACM conference on fairness, accountability, and transparency},
  pages={850--863},
  year={2022}
}

@article{fisher2019all,
  title={All models are wrong, but many are useful: Learning a variable's importance by studying an entire class of prediction models simultaneously},
  author={Fisher, Aaron and Rudin, Cynthia and Dominici, Francesca},
  journal={Journal of machine learning research},
  volume={20},
  number={177},
  pages={1--81},
  year={2019}
}

@inproceedings{semenova2022existence,
  title={On the existence of simpler machine learning models},
  author={Semenova, Lesia and Rudin, Cynthia and Parr, Ronald},
  booktitle={Proceedings of the 2022 ACM Conference on Fairness, Accountability, and Transparency},
  pages={1827--1858},
  year={2022}
}

@article{geirhos2020shortcut,
   title={Shortcut learning in deep neural networks},
  author={Geirhos, Robert and Jacobsen, J{\"o}rn-Henrik and Michaelis, Claudio and Zemel, Richard and Brendel, Wieland and Bethge, Matthias and Wichmann, Felix A},
  journal={Nature Machine Intelligence},
  volume={2},
  number={11},
  pages={665--673},
  year={2020},
  publisher={Nature Publishing Group UK London}
}

@inproceedings{beery2018recognition,
  title={Recognition in terra incognita},
  author={Beery, Sara and Van Horn, Grant and Perona, Pietro},
  booktitle={European conference on computer vision},
  pages={472--489},
  year={2018},
  organization={Springer}
}

@misc{arjovsky2019invariant,
  title={Invariant risk minimization},
  author={Arjovsky, Martin and Bottou, L{\'e}on and Gulrajani, Ishaan and Lopez-Paz, David},
  journal={arXiv preprint arXiv:1907.02893},
  year={2019}
}

@article{peters2016causal,
  title={Causal inference by using invariant prediction: identification and confidence intervals},
  author={Peters, Jonas and B{\"u}hlmann, Peter and Meinshausen, Nicolai},
  journal={Journal of the Royal Statistical Society Series B: Statistical Methodology},
  volume={78},
  number={5},
  pages={947--1012},
  year={2016},
  publisher={Oxford University Press}
}

@inproceedings{jiang2022assessing,
  title={Assessing generalization of {SGD} via disagreement},
  author={Jiang, Yiding and Nagarajan, Vaishnavh and Baek, Christina and Kolter, J Zico},
  booktitle={International Conference on Learning Representations},
  year={2022}
}

@inproceedings{seung1992query,
  title={Query by committee},
  author={Seung, H Sebastian and Opper, Manfred and Sompolinsky, Haim},
  booktitle={Proceedings of the fifth annual workshop on Computational learning theory},
  pages={287--294},
  year={1992}
}

@article{freund1997selective,
  title={Selective sampling using the query by committee algorithm},
  author={Freund, Yoav and Seung, H Sebastian and Shamir, Eli and Tishby, Naftali},
  journal={Machine learning},
  volume={28},
  number={2},
  pages={133--168},
  year={1997},
  publisher={Springer}
}

@misc{houlsby2011bayesian,
  title={{Bayesian} active learning for classification and preference learning},
  author={Houlsby, Neil and Husz{\'a}r, Ferenc and Ghahramani, Zoubin and Lengyel, M{\'a}t{\'e}},
  journal={arXiv preprint arXiv:1112.5745},
  year={2011}
}

@inproceedings{dehaan2019causal,
  title={Causal confusion in imitation learning},
  author={De Haan, Pim and Jayaraman, Dinesh and Levine, Sergey},
  booktitle={Advances in Neural Information Processing Systems},
  volume={32},
  year={2019}
}

@inproceedings{veitch2021counterfactual,
  title={Counterfactual invariance to spurious correlations in text classification},
  author={Veitch, Victor and D'Amour, Alexander and Yadlowsky, Steve and Eisenstein, Jacob},
  booktitle={Advances in Neural Information Processing Systems},
  volume={34},
  pages={16196--16208},
  year={2021}
}

@inproceedings{szegedy2014intriguing,
  title={Intriguing properties of neural networks},
  author={Szegedy, Christian and Zaremba, Wojciech and Sutskever, Ilya and Bruna, Joan and Erhan, Dumitru and Goodfellow, Ian and Fergus, Rob},
  booktitle={International Conference on Learning Representations},
  year={2014}
}

@inproceedings{moosavi2016deepfool,
  title={Deepfool: a simple and accurate method to fool deep neural networks},
  author={Moosavi-Dezfooli, Seyed-Mohsen and Fawzi, Alhussein and Frossard, Pascal},
  booktitle={Proceedings of the IEEE conference on computer vision and pattern recognition},
  pages={2574--2582},
  year={2016}
}

@inproceedings{sagawa2020distributionally,
  title={Distributionally robust neural networks for group shifts: On the importance of regularization for worst-case generalization},
  author={Sagawa, Shiori and Koh, Pang Wei and Hashimoto, Tatsunori B and Liang, Percy},
  booktitle={International Conference on Learning Representations},
  year={2020}
}

@inproceedings{richens2024robust,
  title={Robust agents learn causal world models},
  author={Richens, Jonathan and Everitt, Tom},
  booktitle={International Conference on Learning Representations},
  year={2024}
}

@inproceedings{lakshminarayanan2017simple,
  title={Simple and scalable predictive uncertainty estimation using deep ensembles},
  author={Lakshminarayanan, Balaji and Pritzel, Alexander and Blundell, Charles},
  booktitle={Advances in Neural Information Processing Systems},
  volume={30},
  year={2017}
}

@article{wachter2017counterfactual,
  title={Counterfactual explanations without opening the black box: Automated decisions and the {GDPR}},
  author={Wachter, Sandra and Mittelstadt, Brent and Russell, Chris},
  journal={Harvard Journal of Law \& Technology},
  volume={31},
  number={2},
  pages={841--887},
  year={2018}
}

@inproceedings{ustun2019actionable,
  title={Actionable recourse in linear classification},
  author={Ustun, Berk and Spangher, Alexander and Liu, Yang},
  booktitle={Proceedings of the conference on fairness, accountability, and transparency},
  pages={10--19},
  year={2019}
}

@inproceedings{karimi2021algorithmic,
  title={Algorithmic recourse: from counterfactual explanations to interventions},
  author={Karimi, Amir-Hossein and Sch{\"o}lkopf, Bernhard and Valera, Isabel},
  booktitle={Proceedings of the 2021 ACM conference on fairness, accountability, and transparency},
  pages={353--362},
  year={2021}
}

@article{karimi2022survey,
  title={A survey of algorithmic recourse: contrastive explanations and consequential recommendations},
  author={Karimi, Amir-Hossein and Barthe, Gilles and Sch{\"o}lkopf, Bernhard and Valera, Isabel},
  journal={ACM Computing Surveys},
  volume={55},
  number={5},
  pages={1--29},
  year={2022},
  publisher={ACM New York, NY}
}

@inproceedings{geiger2021causal,
  author = {Geiger, Atticus and Lu, Hanson and Icard, Thomas and Potts, Christopher},
 booktitle = {Advances in Neural Information Processing Systems},
 editor = {M. Ranzato and A. Beygelzimer and Y. Dauphin and P.S. Liang and J. Wortman Vaughan},
 pages = {9574--9586},
 publisher = {Curran Associates, Inc.},
 title = {Causal Abstractions of Neural Networks},
 url = {https://proceedings.neurips.cc/paper_files/paper/2021/file/4f5c422f4d49a5a807eda27434231040-Paper.pdf},
 volume = {34},
 year = {2021}
}

@inproceedings{geiger2024finding,
  title={Finding alignments between interpretable causal variables and distributed neural representations},
  author={Geiger, Atticus and Wu, Zhengxuan and Potts, Christopher and Icard, Thomas and Goodman, Noah},
  booktitle={Causal Learning and Reasoning},
  pages={160--187},
  year={2024},
  organization={PMLR}
}

@article{geiger2023causal,
  title={Causal abstraction: A theoretical foundation for mechanistic interpretability},
  author={Geiger, Atticus and Ibeling, Duligur and Zur, Amir and Chaudhary, Maheep and Chauhan, Sonakshi and Huang, Jing and Arora, Aryaman and Wu, Zhengxuan and Goodman, Noah and Potts, Christopher and others},
  journal={Journal of Machine Learning Research},
  volume={26},
  number={83},
  pages={1--64},
  year={2025}
}

@inproceedings{vig2020investigating,
  title={Investigating gender bias in language models using causal mediation analysis},
  author={Vig, Jesse and Gehrmann, Sebastian and Belinkov, Yonatan and Qian, Sharon and Nevo, Daniel and Singer, Yaron and Shieber, Stuart},
  booktitle={Advances in Neural Information Processing Systems},
  volume={33},
  pages={12388--12401},
  year={2020}
}

@inproceedings{meng2022locating,
  title={Locating and editing factual associations in {GPT}},
  author={Meng, Kevin and Bau, David and Andonian, Alex and Belinkov, Yonatan},
  booktitle={Advances in Neural Information Processing Systems},
  volume={35},
  pages={17359--17372},
  year={2022}
}

@inproceedings{conmy2023towards,
  title={Towards automated circuit discovery for mechanistic interpretability},
  author={Conmy, Arthur and Mavor-Parker, Augustine N and Lynch, Aengus and Heimersheim, Stefan and Garriga-Alonso, Adri{\`a}},
  booktitle={Advances in Neural Information Processing Systems},
  volume={36},
  year={2023}
}

@inproceedings{wang2023interpretability,
  title={Interpretability in the wild: a circuit for indirect object identification in {GPT}-2 small},
  author={Wang, Kevin and Variengien, Alexandre and Conmy, Arthur and Shlegeris, Buck and Steinhardt, Jacob},
  booktitle={International Conference on Learning Representations},
  year={2023}
}

@inproceedings{kaushik2020learning,
  title={Learning the difference that makes a difference with counterfactually-augmented data},
  author={Kaushik, Divyansh and Hovy, Eduard and Lipton, Zachary C},
  booktitle={International Conference on Learning Representations},
  year={2020}
}

@inproceedings{gardner2020evaluating,
  title={Evaluating models' local decision boundaries via contrast sets},
  author={Gardner, Matt and Artzi, Yoav and Basmov, Victoria and Berant, Jonathan and Bogin, Ben and Chen, Sihao and Dasigi, Pradeep and Dua, Dheeru and Elazar, Yanai and Gottumukkala, Ananth and Gupta, Nitish and Hajishirzi, Hannaneh and Ilharco, Gabriel and Khashabi, Daniel and Lin, Kevin and Liu, Jiangming and Liu, Nelson F and Mulcaire, Phoebe and Ning, Qiang and Singh, Sameer and Smith, Noah A and Subramanian, Sanjay and Tsarfaty, Reut and Wallace, Eric and Zhang, Ally and Zhou, Ben},
  booktitle={Findings of the Association for Computational Linguistics: EMNLP 2020},
  pages={1307--1323},
  year={2020}
}

@inproceedings{ribeiro2020beyond,
  title={Beyond accuracy: Behavioral testing of {NLP} models with {CheckList}},
  author={Ribeiro, Marco Tulio and Wu, Tongshuang and Guestrin, Carlos and Singh, Sameer},
  booktitle={Proceedings of the 58th Annual Meeting of the Association for Computational Linguistics},
  pages={4902--4912},
  year={2020}
}

@inproceedings{abraham2022cebab,
  title={{CEBaB}: Estimating the causal effects of real-world concepts on {NLP} model behavior},
  author={Abraham, Eldar David and D'Oosterlinck, Karel and Feder, Amir and Gat, Yair Ori and Geiger, Atticus and Potts, Christopher and Reichart, Roi and Wu, Zhengxuan},
  booktitle={Advances in Neural Information Processing Systems},
  volume={35},
  year={2022}
}

@inproceedings{pei2017deepxplore,
  title={Deepxplore: Automated whitebox testing of deep learning systems},
  author={Pei, Kexin and Cao, Yinzhi and Yang, Junfeng and Jana, Suman},
  booktitle={proceedings of the 26th Symposium on Operating Systems Principles},
  pages={1--18},
  year={2017}
}

@inproceedings{xie2019diffchaser,
  title={{DiffChaser}: Detecting disagreements for deep neural networks},
  author={Xie, Xiaofei and Ma, Lei and Wang, Haijun and Li, Yuekang and Liu, Yang and Li, Xiaohong},
  booktitle={Proceedings of the 28th International Joint Conference on Artificial Intelligence},
  pages={5772--5778},
  year={2019}
}

@article{mckeeman1998differential,
  title={Differential testing for software},
  author={McKeeman, William M},
  journal={Digital Technical Journal},
  volume={10},
  number={1},
  pages={100--107},
  year={1998}
}

@article{angluin1988queries,
  title={Queries and concept learning},
  author={Angluin, Dana},
  journal={Machine learning},
  volume={2},
  number={4},
  pages={319--342},
  year={1988},
  publisher={Springer}
}

@article{lee1996principles,
 title={Principles and methods of testing finite state machines-a survey},
  author={Lee, David and Yannakakis, Mihalis},
  journal={Proceedings of the IEEE},
  volume={84},
  number={8},
  pages={1090--1123},
  year={1996},
  publisher={IEEE}
}

@inproceedings{kusner2017counterfactual,
  title={Counterfactual fairness},
  author={Kusner, Matt J and Loftus, Joshua and Russell, Chris and Silva, Ricardo},
  booktitle={Advances in Neural Information Processing Systems},
  volume={30},
  year={2017}
}

@inproceedings{mothilal2020explaining,
  title={Explaining machine learning classifiers through diverse counterfactual explanations},
  author={Mothilal, Ramaravind K and Sharma, Amit and Tan, Chenhao},
  booktitle={Proceedings of the 2020 conference on fairness, accountability, and transparency},
  pages={607--617},
  year={2020}
}

@article{eiter2002complexity,
  title={Complexity results for structure-based causality},
  author={Eiter, Thomas and Lukasiewicz, Thomas},
  journal={Artificial Intelligence},
  volume={142},
  number={1},
  pages={53--89},
  year={2002},
  publisher={Elsevier}
}

@article{aleksandrowicz2017computational,
  title={The computational complexity of structure-based causality},
  author={Aleksandrowicz, Gadi and Chockler, Hana and Halpern, Joseph Y and Ivrii, Alexander},
  journal={Journal of Artificial Intelligence Research},
  volume={58},
  pages={431--451},
  year={2017}
}

@inproceedings{ibrahim2020checking,
  title={From checking to inference: Actual causality computations as optimization problems},
  author={Ibrahim, Amjad and Pretschner, Alexander},
  booktitle={International Symposium on Automated Technology for Verification and Analysis},
  pages={343--359},
  year={2020},
  organization={Springer}
}

@article{mosse2024causal,
  title={Is causal reasoning harder than probabilistic reasoning?},
  author={Moss{\'e}, Milan and Ibeling, Duligur and Icard, Thomas},
  journal={The Review of Symbolic Logic},
  volume={17},
  number={1},
  pages={106--131},
  year={2024},
  publisher={Cambridge University Press}
}

@inproceedings{ibeling2020probabilistic,
 title={Probabilistic reasoning across the causal hierarchy},
  author={Ibeling, Duligur and Icard, Thomas},
  booktitle={Proceedings of the AAAI Conference on Artificial Intelligence},
  volume={34},
  pages={10170--10177},
  year={2020}
}

@inproceedings{jin2023cladder,
  title={{CLadder}: Assessing causal reasoning in language models},
  author={Jin, Zhijing and Chen, Yuen and Leeb, Felix and Gresele, Luigi and Kamal, Ojasv and Lyu, Zhiheng and Blin, Kevin and Gonzalez Adauto, Fernando and Kleiman-Weiner, Max and Sachan, Mrinmaya and others},
  booktitle={Advances in Neural Information Processing Systems},
  volume={36},
  pages={31038--31065},
  year={2023}
}

@article{kiciman2023causal,
  title={Causal reasoning and large language models: Opening a new frontier for causality},
  author={K{\i}c{\i}man, Emre and Ness, Robert and Sharma, Amit and Tan, Chenhao},
  journal={Transactions on Machine Learning Research},
  year={2024}
}

@article{zecevic2023causal,
  title={Causal parrots: Large language models may talk causality but are not causal},
  author={Ze{\v{c}}evi{\'c}, Matej and Willig, Moritz and Dhami, Devendra Singh and Kersting, Kristian},
  journal={Transactions on Machine Learning Research},
  year={2023}
}

@article{scholkopf2021toward,
  title={Toward causal representation learning},
  author={Sch{\"o}lkopf, Bernhard and Locatello, Francesco and Bauer, Stefan and Ke, Nan Rosemary and Kalchbrenner, Nal and Goyal, Anirudh and Bengio, Yoshua},
  journal={Proceedings of the IEEE},
  volume={109},
  number={5},
  pages={612--634},
  year={2021},
  publisher={IEEE}
}

@inproceedings{parascandolo2018learning,
  title={Learning independent causal mechanisms},
  author={Parascandolo, Giambattista and Kilbertus, Niki and Rojas-Carulla, Mateo and Sch{\"o}lkopf, Bernhard},
  booktitle={International Conference on Machine Learning},
  pages={4036--4044},
  year={2018},
  organization={PMLR}
}

@article{bareinboim2016causal,
  title={Causal inference and the data-fusion problem},
  author={Bareinboim, Elias and Pearl, Judea},
  journal={Proceedings of the National Academy of Sciences},
  volume={113},
  number={27},
  pages={7345--7352},
  year={2016},
  publisher={National Academy of Sciences}
}

@inproceedings{xia2021causal,
  title={The causal-neural connection: Expressiveness, learnability, and inference},
  author={Xia, Kevin and Lee, Kai-Zhan and Bengio, Yoshua and Bareinboim, Elias},
  booktitle={Advances in Neural Information Processing Systems},
  volume={34},
  pages={10823--10836},
  year={2021}
}

@article{kauffman1969metabolic,
  title={Metabolic stability and epigenesis in randomly constructed genetic nets},
  author={Kauffman, Stuart A},
  journal={Journal of theoretical biology},
  volume={22},
  number={3},
  pages={437--467},
  year={1969},
  publisher={Elsevier}
}

@article{mitchell1982generalization,
  title={Generalization as search},
  author={Mitchell, Tom M},
  journal={Artificial intelligence},
  volume={18},
  number={2},
  pages={203--226},
  year={1982},
  publisher={Elsevier}
}

@article{kreutz2009systems,
  title={Systems biology: experimental design},
  author={Kreutz, Clemens and Timmer, Jens},
  journal={The FEBS journal},
  volume={276},
  number={4},
  pages={923--942},
  year={2009},
  publisher={Wiley Online Library}
}

@inproceedings{oberst2019counterfactual,
  title={Counterfactual off-policy evaluation with gumbel-max structural causal models},
  author={Oberst, Michael and Sontag, David},
  booktitle={International Conference on Machine Learning},
  pages={4881--4890},
  year={2019},
  organization={PMLR}
}

@book{popper1959logic,
  author    = {Popper, Karl},
  title     = {The Logic of Scientific Discovery},
  publisher = {Routledge},
  address   = {London},
  year      = {2002},
  note      = {Routledge Classics edition; first English edition Hutchinson, London, 1959}
}

@inproceedings{korb2004varieties,
  title={Varieties of causal intervention},
  author={Korb, Kevin B and Hope, Lucas R and Nicholson, Ann E and Axnick, Karl},
  booktitle={Pacific Rim international conference on artificial intelligence},
  pages={322--331},
  year={2004},
  organization={Springer}
}

@misc{lu2020sample,
  title={Sample-efficient reinforcement learning via counterfactual-based data augmentation},
  author={Lu, Chaochao and Huang, Biwei and Wang, Ke and Hern{\'a}ndez-Lobato, Jos{\'e} Miguel and Zhang, Kun and Sch{\"o}lkopf, Bernhard},
  journal={arXiv preprint arXiv:2012.09092},
  year={2020}
}

@inproceedings{zhang2018fairness,
  title={Fairness in decision-making—the causal explanation formula},
  author={Zhang, Junzhe and Bareinboim, Elias},
  booktitle={Proceedings of the AAAI conference on artificial intelligence},
  volume={32},
  year={2018}
}

@article{pearl1995causal,
  title={Causal diagrams for empirical research},
  author={Pearl, Judea},
  journal={Biometrika},
  volume={82},
  number={4},
  pages={669--688},
  year={1995}
}

@article{shpitser2008complete,
  title={Complete identification methods for the causal hierarchy},
  author={Shpitser, Ilya and Pearl, Judea},
  journal={Journal of Machine Learning Research},
  volume={9},
  pages={1941--1979},
  year={2008}
}

@book{hernan2020causal,
  title={Causal Inference: What If},
  author={Hern{\'a}n, Miguel A. and Robins, James M.},
  publisher={Chapman \& Hall/CRC},
  year={2020},
  url={https://miguelhernan.org/whatifbook}
}

@inproceedings{annadani2024amortized,
  title={Amortized active causal induction with deep reinforcement learning},
  author={Annadani, Yashas and Tigas, Panagiotis and Bauer, Stefan and Foster, Adam},
  booktitle={Advances in Neural Information Processing Systems},
  volume={37},
  year={2024}
}

@inproceedings{lorch2022amortized,
  title={Amortized inference for causal structure learning},
  author={Lorch, Lars and Sussex, Scott and Rothfuss, Jonas and Krause, Andreas and Sch{\"o}lkopf, Bernhard},
  booktitle={Advances in Neural Information Processing Systems},
  volume={35},
  pages={13104--13118},
  year={2022}
}

@inproceedings{zhang2023bayesian,
  title={{Bayesian} active causal discovery with multi-fidelity experiments},
  author={Zhang, Zeyu and Li, Chaozhuo and Chen, Xu and Xie, Xing},
  booktitle={Advances in Neural Information Processing Systems},
  volume={36},
  year={2023}
}

@inproceedings{guo2026constrained,
  title={Constrained {B}ayesian experimental design via online planning},
  author={Guo, Yujia and Huang, Daolang and Zhang, Xinyu and Katt, Sammie and Kaski, Samuel and Bharti, Ayush},
  booktitle={Proceedings of the 43rd International Conference on Machine Learning},
  year={2026}
}

@inproceedings{tigas2022interventions,
  title={Interventions, where and how? {E}xperimental design for causal models at scale},
  author={Tigas, Panagiotis and Annadani, Yashas and Jesson, Andrew and Sch{\"o}lkopf, Bernhard and Gal, Yarin and Bauer, Stefan},
  booktitle={Advances in Neural Information Processing Systems},
  volume={35},
  pages={24130--24143},
  year={2022}
}

@inproceedings{geiger2022inducing,
  title={Inducing causal structure for interpretable neural networks},
  author={Geiger, Atticus and Wu, Zhengxuan and Lu, Hanson and Rozner, Josh and Kreiss, Elisa and Icard, Thomas and Goodman, Noah and Potts, Christopher},
  booktitle={Proceedings of the 39th International Conference on Machine Learning},
  pages={7324--7338},
  year={2022}
}

@article{elazar2021amnesic,
  title={Amnesic probing: Behavioral explanation with amnesic counterfactuals},
  author={Elazar, Yanai and Ravfogel, Shauli and Jacovi, Alon and Goldberg, Yoav},
  journal={Transactions of the Association for Computational Linguistics},
  volume={9},
  pages={160--175},
  year={2021}
}

@misc{zhang2021pointer,
  title={Pointer Value Retrieval: A new benchmark for understanding the limits of neural network generalization},
  author={Zhang, Chiyuan and Raghu, Maithra and Kleinberg, Jon and Bengio, Samy},
  howpublished={arXiv preprint arXiv:2107.12580},
  year={2021}
}

@inproceedings{koh2020concept,
  title={Concept bottleneck models},
  author={Koh, Pang Wei and Nguyen, Thao and Tang, Yew Siang and Mussmann, Stephen and Pierson, Emma and Kim, Been and Liang, Percy},
  booktitle={Proceedings of the 37th International Conference on Machine Learning},
  pages={5338--5348},
  year={2020}
}

@article{lecun1998gradient,
  title={Gradient-based learning applied to document recognition},
  author={LeCun, Yann and Bottou, L{\'e}on and Bengio, Yoshua and Haffner, Patrick},
  journal={Proceedings of the IEEE},
  volume={86},
  number={11},
  pages={2278--2324},
  year={1998}
}

@misc{bengio2013estimating,
  title={Estimating or propagating gradients through stochastic neurons for conditional computation},
  author={Bengio, Yoshua and L{\'e}onard, Nicholas and Courville, Aaron},
  howpublished={arXiv preprint arXiv:1308.3432},
  year={2013}
}
\bibliographystyle{iclr2027_conference}
\fi

\fi 

\end{document}